\documentclass[10pt,twocolumn,letterpaper]{article}

\usepackage{iccv}
\usepackage{times}
\usepackage{epsfig}
\usepackage{graphicx}
\usepackage{amsmath}
\usepackage{amssymb}

\usepackage{xcolor}
\usepackage[normalem]{ulem}
\usepackage{algorithm}
\usepackage{booktabs,multirow,array,caption}
\usepackage{subcaption}

\usepackage{cases}
\usepackage{algpseudocode}
\usepackage{pdflscape}
\usepackage{adjustbox}

\usepackage[pagebackref=true,breaklinks=true,letterpaper=true,colorlinks,bookmarks=false]{hyperref}

\iccvfinalcopy

\ificcvfinal\fi

\begin{document}

%%%%%%%%% TITLE
\title{Non-uniform Blur Kernel Estimation via Adaptive Basis Decomposition}

\author{Guillermo Carbajal\\
Universidad de la República\\
Uruguay

\and
Patricia Vitoria\\
Universitat Pompeu Fabra\\
Spain

\and

Mauricio Delbracio\\
Universidad de la República\\
Uruguay

\and
Pablo Musé\\
Universidad de la República\\
Uruguay

\and
José Lezama\\
Universidad de la República\\
Uruguay

}

\maketitle
\ificcvfinal\thispagestyle{empty}\fi

%%%%%%%%% ABSTRACT
\begin{abstract}
Motion blur estimation remains an important task for scene analysis and image restoration. In recent years, the removal of motion blur in photographs has seen impressive progress in the hands of deep learning-based methods, trained to map directly from blurry to sharp images. Characterization of the motion blur, on the other hand, has received less attention, and progress in model-based methods for deblurring lags behind that of data-driven end-to-end approaches.
% One paragraph abstract is better (MD).
In this work we revisit the problem of characterizing dense, non-uniform motion blur in a single image and propose a general non-parametric model for this task. Given a blurry image, a neural network is trained to estimate a set of image-adaptive basis motion kernels as well as the mixing coefficients at the pixel level, producing a per-pixel motion blur field. 
%This rich but efficient forward degradation model allows to use existing tools for solving inverse problems. 
We show that our approach overcomes the limitations of existing non-uniform motion blur estimation methods and leads to extremely accurate motion blur kernels.
When applied to real motion-blurred images, a variational non-uniform blur removal method fed with the estimated blur kernels produces high-quality restored images. Qualitative and quantitative evaluation shows that these results are competitive or superior to results obtained with existing end-to-end deep learning (DL) based methods, thus bridging the gap between model-based and data-driven approaches.

% Characterizing and removing motion blur caused by camera shake or object motion remains an important task for scene analysis and image restoration. In recent years, the removal of motion blur in photographs has seen impressive progress in the hands of deep learning-based methods, trained to map directly from blurry to sharp images. Characterization of motion blur, on the other hand, has received less attention, and progress in model-based methods for restoration lags behind that of data-driven end-to-end approaches.
% % One paragraph abstract is better (MD).
% In this work we revisit the problem of characterizing dense, non-uniform motion blur in a single image and propose a general non-parametric model for this task. Given a blurry image, we estimate a set of adaptive basis motion kernels as well as the mixing coefficients at the pixel level, producing a per-pixel motion blur field. This rich but efficient forward model of the degradation process allows to use existing tools for solving inverse problems. We show that our approach overcomes the limitations of existing non-uniform motion blur estimation methods. When applied to real motion-blurred images, a variational non-uniform blur removal method fed with the estimated blur kernels produces high-quality restored images. Qualitative and quantitative evaluation shows that these results are competitive or superior to results obtained with existing end-to-end deep learning (DL) based methods, thus bridging the gap between model-based and data-driven approaches.

\end{abstract}

% https://drive.google.com/file/d/1yA6UIW1x5SSGPiaGMHn18n7WosE89zCh/view?usp=sharing

%%%%%%%%%%%%%%%%%%%%%%%%%%%%%%%%
%%%%%%%%% BODY TEXT
\section{Introduction}

%% OLD - PM 16/03
%Motion blur results from the relative motion between the camera and the scene, which is determined by the interaction of three elements: the motion of the camera or egomotion, the three-dimensional geometry of the scene, and the motion of objects in the scene. When the exposure time is large compared to the relative motion, the camera sensor at each point receives and accumulates light coming from different sources, producing different amounts of blur. 

\begin{figure}[h]
    \centering
    \includegraphics[width=0.48\textwidth]{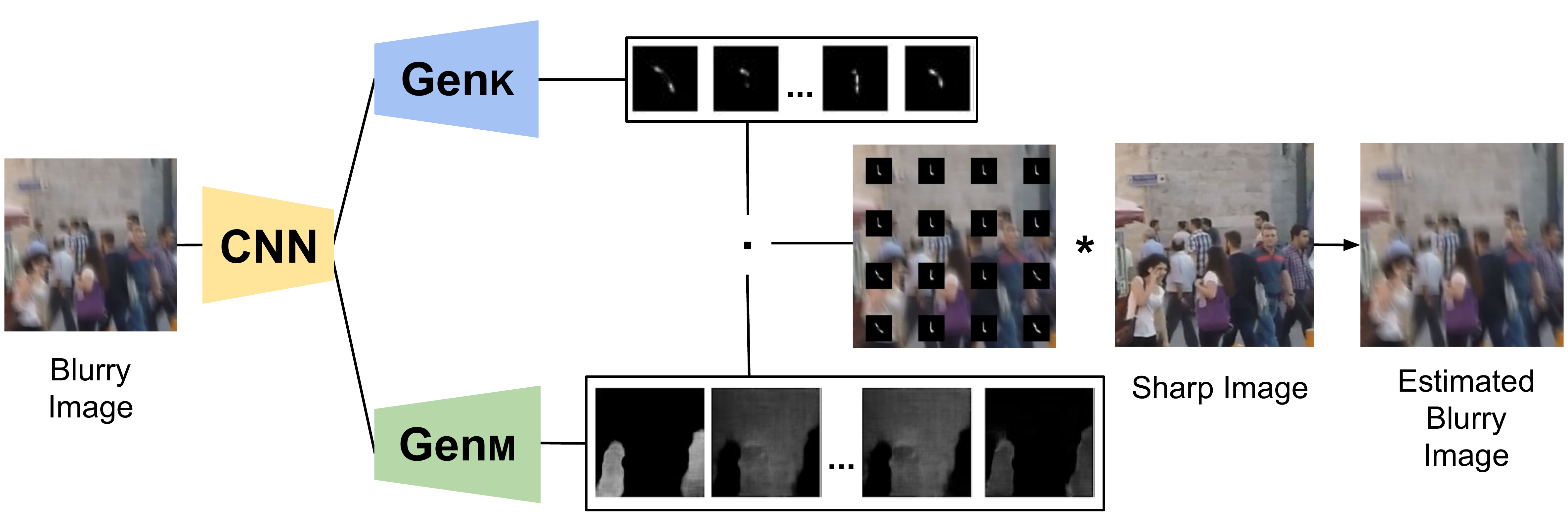}
    \caption{\textbf{Overview of the proposed method.} Given a blurry image, a neural network predicts a set of kernel basis and corresponding pixel-wise mixing coefficients allowing to reblur the corresponding sharp image by first convolving it with each basis kernel and performing a weighted sum using the mixing coefficients.}
    \label{fig:pipeline}
\end{figure}

\begin{figure*}[t!]
\setlength\tabcolsep{1.0pt} % default value: 6pt
\centering
\begin{tabular}{c@{\hspace{1em}}c}
\multirow{2}{*}[0.26cm]{\includegraphics[width=0.07\textwidth]{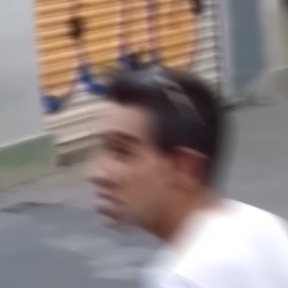}} & \includegraphics[width=0.9\textwidth]{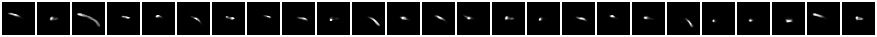}\\
%\addlinespace
& \includegraphics[width=0.896\textwidth]{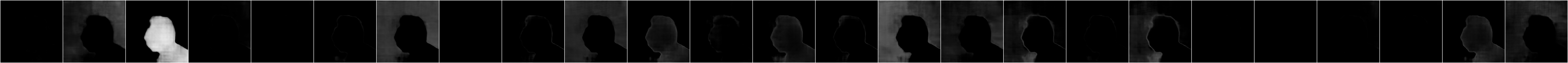}\\
\end{tabular}
\begin{tabular}{c@{\hspace{1em}}c}
\multirow{2}{*}[0.26cm]{\includegraphics[width=0.07\textwidth]{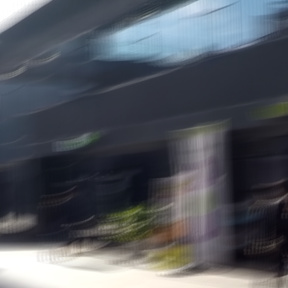}} & \includegraphics[width=0.9\textwidth]{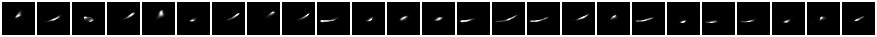}\\
%\addlinespace
& \includegraphics[width=0.896\textwidth]{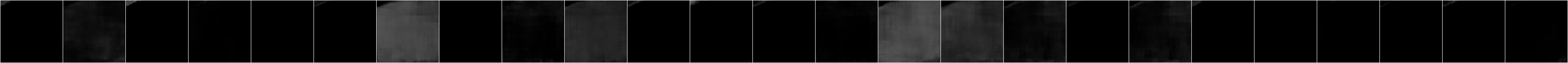}\\
\end{tabular}
\begin{tabular}{c@{\hspace{1em}}c}
\multirow{2}{*}[0.26cm]{\includegraphics[width=0.07\textwidth]{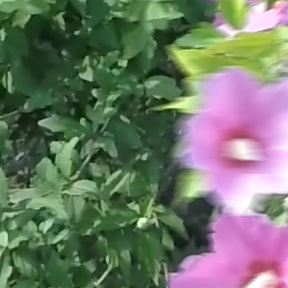}} & \includegraphics[width=0.9\textwidth]{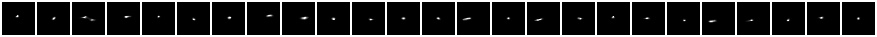}\\
%\addlinespace
& \includegraphics[width=0.896\textwidth]{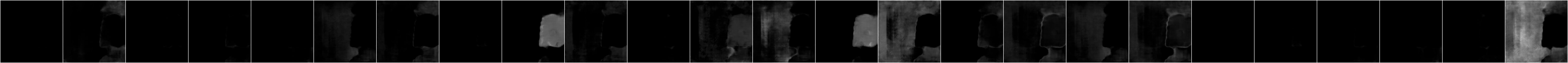}\\
\end{tabular}

\caption{\textbf{Examples of generated kernel basis $\{\mathbf{k}^b\}$  and corresponding mixing coefficients $\{\mathbf{m}^b\}$} predicted from the  blurry images shown on the left. The adaptation to the input is more notorious for the elements  that have significant weights. }
    \label{fig:KernelsAndMasks}
\end{figure*}

Motion blur is a major source of degradation in digital images. This effect, preeminent in low light photography, occurs when the exposure time is large compared to the relative motion speed between the camera and the scene. As a result, the camera sensor at each pixel receives and accumulates light coming from different sources, producing different amounts of blur.

Single image blur kernel estimation is an ill-posed problem, as many pairs of blur kernels and images can generate the same blurry image.  
%{\color{red} JL} Blind image deblurring aims at estimating the blur kernel and the latent image from an input blurry image. This is an ill-posed problem as there are infinitely many pairs of blur kernels and images that could generate the same blurry image.  
In model-based methods, an accurate blur kernel estimation is important for obtaining a high-quality sharp image. Existing non-uniform %motion blur estimation 
methods approximate the blur kernel by assuming a parametric model of the motion field, either by considering a global parametric form induced by camera motion~\cite{gupta2010singleimage,hirsch2011fastremoval,dai2008motion, whyte2010nonuniform}, or  by locally modeling the motion field with linear kernels parameterized by the length and orientation of the kernel support
~\cite{gong2017motion,kim2014segmentation-free,dai2008motion,sun2015learning}. While these methods reduce significantly the complexity and computational cost by solving for a simple approximate motion field, the approximation does not generalize to most real case scenarios, like camera shake from hand tremor~\cite{gavant2011physiological}. 

%%%%%%%%%%%%%%%%%
Motion blur kernel estimation proves useful to infer scene motion information and to solve related tasks such as tracking and motion segmentation. However, its major motivation is certainly blind motion deblurring. Recently, the unprecedented improvements in deblurring obtained by DL methods have contributed to a decreased interest in model-based approaches and motion blur kernel estimation. Indeed, DL deblurring methods skip the kernel estimation step arguing that~\cite{Nah_2017_CVPR}: {\em (i)} blur kernels models are too simplistic and unrealistic to be used in practice; {\em (ii)}  the kernel estimation process is subtle and sensitive to noise and saturation; {\em (iii)} finding a spatially varying kernel for every pixel in a dynamic scene requires a huge amount of memory and computation. 
While these limitations are certainly true, current deblurring DL-based methods also suffer from a major limitation related to the way the training datasets are produced. To synthesize motion-blurred images without relying on a forward kernel-based model, the training dataset is obtained by averaging consecutive frames from a sequence of a dynamic scene captured with a high-speed camera \cite{Nah_2019_CVPR_Workshops_REDS,su2017deep,sun2015learning}. Such datasets are only able to characterize the motion blur to a limited extent,
and their end-to-end training schemes induce a mapping that might be specific to the camera that was used, capturing transformations other than deblurring.
%
%and their generation involve image alignment and averaging resulting in a denoising effect. 
%
As a consequence, and as confirmed by the experiments presented in this work,   DL-based methods sometimes fail to generalize to real blurred images.    

%%%%%%%%%%%%%%%%%%%%
In this work, we propose a novel approach for non-parametric, dense, spatially-varying motion blur estimation based on an efficient low-rank representation of the per-pixel
motion blur kernels. 
%The proposed method does not only produces precise non-uniform kernels but also reduces the complexity of the problem. 
More precisely, for each blurred image, a convolutional neural network (CNN) estimates an image-specific set of kernel basis functions, as well as a set of pixel-wise mixing coefficients, cf. Figures~\ref{fig:pipeline},~\ref{fig:KernelsAndMasks},~\ref{fig:KernelsAndMasks2}. The combination of the two results in a  unique motion blur kernel per pixel.
%In this way, for each pixel, a unique motion blur kernel is assigned, given by the corresponding linear combination of the image-specific kernel basis functions.
Additionally, our model can handle saturation regions by explicitly modeling this phenomenon. When applied to image deblurring, this model leads to high-quality images whether or not saturated regions exist. We show that the proposed method is capable to overcome the above-mentioned limitations of model-based methods, producing precise spatially varying kernels while reducing the complexity of the problem.

%Considering that non-uniform motion blur kernel estimation has received less attention in the last four years for the above-mentioned reasons, w

We design the following strategy to assess the quality of the estimated motion blur kernels: {\em (i)} We compare the estimated kernels obtained by our method with the ones obtained by prior art ~\cite{gong2017motion,sun2015learning}; {\em (ii)} We apply the classical Richardson-Lucy~\cite{Lucy1974,Richardson1972} deblurring method to the non-uniform case,  and compare the results using our estimated motion blur fields with state-of-the-art DL-based end-to-end methods \cite{kupyn2018deblurgan,kupyn2019deblurgan,rim_2020_ECCV,tao2018scale,Zhang_2019_CVPR}. This comparison shows not only that our motion blur kernels are extremely accurate, but that the motion deblurring obtained as a by-side product is competitive and is often able to outperform DL-based methods in standard benchmarks of real blurred images.
Code and pre-trained model weights are publicly available\footnote{ \url{https://github.com/GuillermoCarbajal/NonUniformMotionBlurKernelEstimation}}.

\section{Related Work}

\noindent {\bf Single Image Non-Uniform Motion Blur Estimation.} The literature on motion kernel estimation is extremely vast. We limit the current analysis to spatially varying kernel estimation methods from a single image.  
Early methods attempting to estimate non-uniform motion blur kernels are limited to the case of camera egomotion \cite{gupta2010singleimage,hirsch2011fastremoval,tai2011richardson-lucy,whyte2010nonuniform}. They assume the scenes are planar and therefore are mapped to the image plane by a homography. 
% This leads to the so-called Projective Motion Path Blur Model (PMPB)~\cite{tai2011richardson-lucy}. These methods are computationally intensive since the optimization involves a large number of homographies that must be computed for the intermediate estimated images. Hirsch \etal ~\cite{hirsch2011fastremoval} manage to reduce the computational cost by combining the efficiency of the Efficient Filter Flow (EFF) with the camera motion constraints of PMPB. 
To deal with spatially-varying blur due to depth and moving objects, methods like~\cite{kim2013dynamic,pan2016soft} propose to segment the image in a reduced number of layers at fixed depth.
% according to a metric representing the amount of blur. 
One drawback of the piece-wise constant model is that it can be sensitive to the segmentation of the blurred image.
A different approach, more related to ours, that deals with both scene depth variations and moving objects, consists of predicting motion blur locally.
Sun \etal~\cite{sun2015learning} consider a set of pre-defined linear motion kernels parameterized by their lengths and orientations. Then, they predict a patch-level probabilistic distribution of the kernel parameters using a CNN. The patch-level distribution is then converted to a dense motion field using a Markov random field enforcing motion smoothness.

Given a blurry image, Gong \etal~\cite{gong2017motion} directly estimate a dense linear motion flow parameterized by their horizontal and vertical components using a Fully Convolutional Network (FCN). To train the FCN, they simulate motion flows to generate synthetic blurred image/motion flow pairs. %As in~\cite{sun2015learning}, the predicted motion blur kernels are linear and parameterized by their horizontal and vertical components, each of them defined on a discrete set. 
Both methods propose to deblur images using a non-blind deblurring method with the estimated kernels, combining an $L^2$ data fitting term  with EPLL image prior~\cite{zoran2011learning}.\vspace{.3em}

%Another interesting parameterization applied to video sequences appears in~\cite{chen2018reblur2deblur,hyun2015generalized} where per-pixel motion blur kernels are modeled as the combination of two line segments, obtained from the next- and previous-frame optical flow estimations. In ~\cite{chen2018reblur2deblur}, kernels are inferred by a neural network that predicts indexes in a pre-computed look-up table.

%%%%%%%%%%%%%%%%%%%%%

\begin{figure*}[t]
    \centering
    \includegraphics[width=\textwidth]{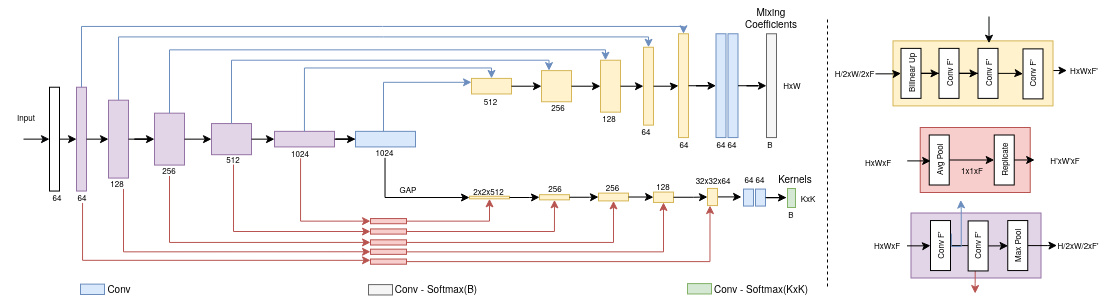}
    \caption{\textbf{Architecture Details} The proposed network is composed by an encoder and two decoders. The encoder takes as an input a blurry image. The decoders output the motion kernel basis and corresponding per-pixel mixing coefficients. }
    \label{fig:architecture}
\end{figure*}
%%%%%%%%%%%%%%%%%%%%

\noindent {\bf Kernel Prediction Networks.} Recently, Kernel Prediction Networks (KPN) have been proposed for low-level vision tasks such as burst denoising \cite{mildenhall2018burst,xia2019basis}, optical flow estimation, frame interpolation ~\cite{niklaus2017video,niklaus2017video2},
  ~stereo and video prediction \cite{jia2016dynamic}. 
 Several works have used KPNs in the context of burst denoising. Mildenhall \etal~\cite{mildenhall2018burst} produce denoised estimates at each pixel as a weighted average of observed local noisy pixels across all the input frames. These averaging weights, or kernels, are predicted from the noisy input burst using a CNN.  
To improve the computational efficiency, Xia \etal~\cite{xia2019basis} propose a basis prediction network that, given an input burst, predicts a set of
global basis kernels — shared within the image — and the
corresponding mixing coefficients, which are specific to individual pixels.\vspace{.3em} 

\noindent {\bf Relation with the Proposed Approach.} In this work, we propose a dense non-uniform motion blur estimation based on an efficient low-rank representation of the pixel-wise motion blur kernels. 
As in ~\cite{gupta2010singleimage,hirsch2011fastremoval,whyte2010nonuniform} the per-pixel kernels are modeled as linear combinations of base elements, but instead of having a single pre-computed basis of elements, we use a KPN to infer a non-parametric basis specific for each input image. Additionally, instead of learning the kernels to solve the image restoration problem (e.g. denoising in~\cite{xia2019basis}), we learn them to fit the forward \emph{degradation} model, enabling the utilization of existing techniques for solving inverse problems.

%Sentence from Plug and Play Denoising that could helpto motivate the use of model based approach ( i added herebecause for now i am not sure if you want to add the deblur-ring methods in introduction, state of the art, or separate itas we did in cvpr):It is easy to note that one main difference between model-based methods and learning-based methods is that, the for-mer are flexible to handle various IR tasks by simply speci-fying T and can directly optimize on the degraded image y,whereas the later require cumbersome training to learn themodel  before  testing  are  usually  restricted  by  specializedtasks.

%\section{Method for Non-Uniform Blur Estimation}\label{sec:method}

\section{Proposed Motion Blur Degradation Model} \label{sec:model}

Non-uniform motion blur can be modeled as the local convolution of a sharp image with a spatially varying filter, the \emph{motion blur field}. This simple model represents the integration, at each pixel, of photons arriving from different sources due to relative motion between the camera and the scene. 
%
% ALTERNATIVE TEXT (to paragraph below)
 Given a sharp image $\mathbf{u}$ of size $H\times W$, and a set of per-pixel blur kernels $\mathbf{k}_i$ of size $K\times K$, we will assume that the observed blurry image $\mathbf{v}$ is generated as
\begin{equation}
    {v}_i = \langle \mathbf{u}_{nn(i)}, \mathbf{k}_i \rangle + n_i,
        \label{eq:model}
\end{equation}
where $\mathbf{u}_{nn(i)}$ is a window of size $K \times K$ around pixel $i$ in image $\mathbf{u}$ and $n_{i}$ is additive noise. We assume that kernels are non-negative (no negative light) and of area one (conservation of energy).
%
%Formally, given a sharp image $\mathbf{u}$ of height $H$ and width $W$, and a set of blur kernels $\mathbf{k}_{i} \in 
%\left[0,1\right]^{K\times K}$, for $i=1\ldots H\times W$, the blurry image $\mathbf{v}$ is the result of applying the per-pixel operation:
%
%\begin{equation}
%    {v}_i = \langle \mathbf{u}_{nn(i)}, \mathbf{k}_i \rangle + n_i,
%        \label{eq:model}
%\end{equation}
%where $\mathbf{u}_{nn(i)}$ is a window of size $K \times K$ around pixel $i$ in the image $\mathbf{u}$ and $n_{i}$ is additive noise. We assume conservation of energy by imposing $||\mathbf{k}_i||_1=1$. 

Predicting the full motion field of per-pixel kernels $\mathbf{k}_i$ would lead to an estimation problem in a very high-dimensional space ($K^2 H W$), being computationally intractable for large images and kernels. We propose an efficient solution, based on the assumption that there exists significant redundancy between the kernels present in the image. We incorporate this assumption in our model by imposing a low-rank structure to the field of motion kernels.  We decompose the per-pixel kernels as linear combinations of elements in a much smaller basis. Specifically, if the blur kernel basis has $B$ elements, then, only $B$ coefficients $\mathbf{m}_i^b$ are required per pixel instead of the original $K \times K$. Additionally, the $B$ basis elements need to be estimated leading to an estimation problem of dimension $B(K^2 + HW)$. A notorious gain is obtained when $B\ll K^2$, particularly relevant for large blur kernels. The mixing coefficients are normalized so that they sum to one at each pixel location. Thus, the per-pixel kernel $\mathbf{k}_i$ results from the convex combination of the basis kernel, conservation of energy is guaranteed, and the degradation model becomes:
\begin{equation}
    {v}_{i} = \langle \mathbf{u}_{nn(i)}, \sum_{b=1}^B\mathbf{k}^b m^b_{i} \rangle + n_{i}.
    \label{eq:modelKernelBasis}
\end{equation}
Taking into account the sensor saturation, and the gamma correction performed by the camera, a more accurate model is given by
\begin{equation}
    v_i = R \left( \right \langle \mathbf{u}_{nn(i)}, \sum_{b=1}^B\mathbf{k}^b m^b_i \rangle + n_i )^{\frac{1}{\gamma}} ,
      \label{eq:model_sat}
\end{equation}
where $\gamma$ is the gamma correction coefficient and $R(\cdot)$ is the pixel saturation operator that clips  image values $v_{i}$ which are larger than 1. To avoid the non-differentiability of this function at $v_{i} = 1$, a smooth approximation for the captor response function is considered~\cite{whyte2014deblurring}:
\begin{equation}
    R(v_i) = v_i - \frac{1}{a}\log (1 + e^{a(v_i-1)}).
      \label{eq:sat_func}
\end{equation}
The parameter $a$ controls the smoothness of the approximation and is set to $a=50$. As for the gamma correction factor, a typical value of $\gamma = 2.2$ is used.

\begin{figure}
  \centering
\setlength{\tabcolsep}{2pt}

  \begin{tabular}{*{3}{c}}
    \includegraphics[trim=400 100 300 200,clip, width=0.152\textwidth]{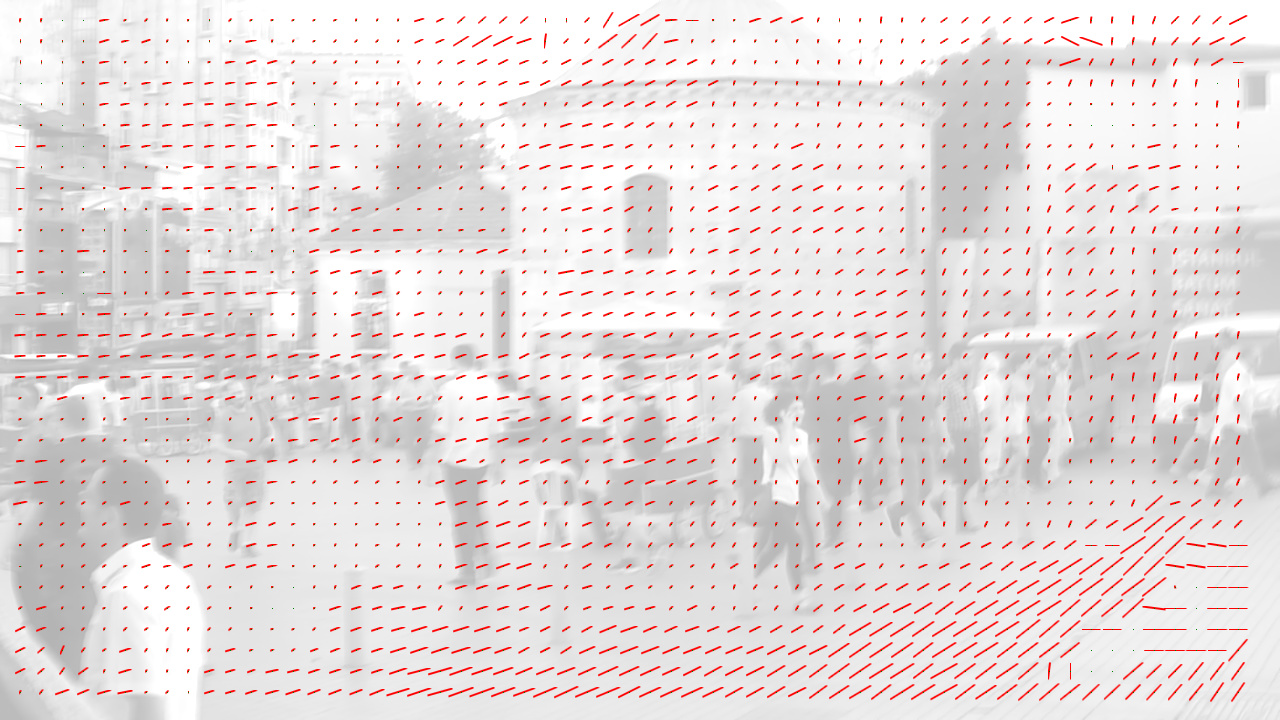}   &
      \includegraphics[trim=400 100 300 200,clip, width=0.152\textwidth]{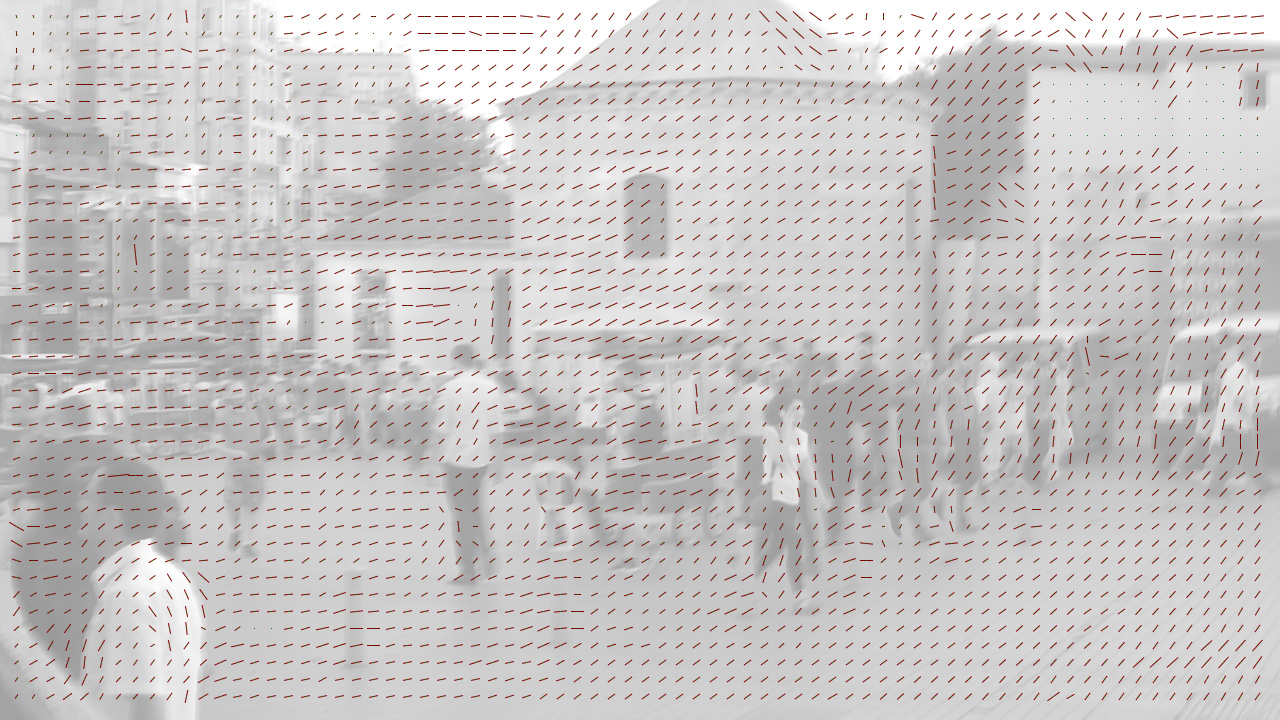}   &  
   \includegraphics[trim=400 100 300 200,clip, width=0.152\textwidth]{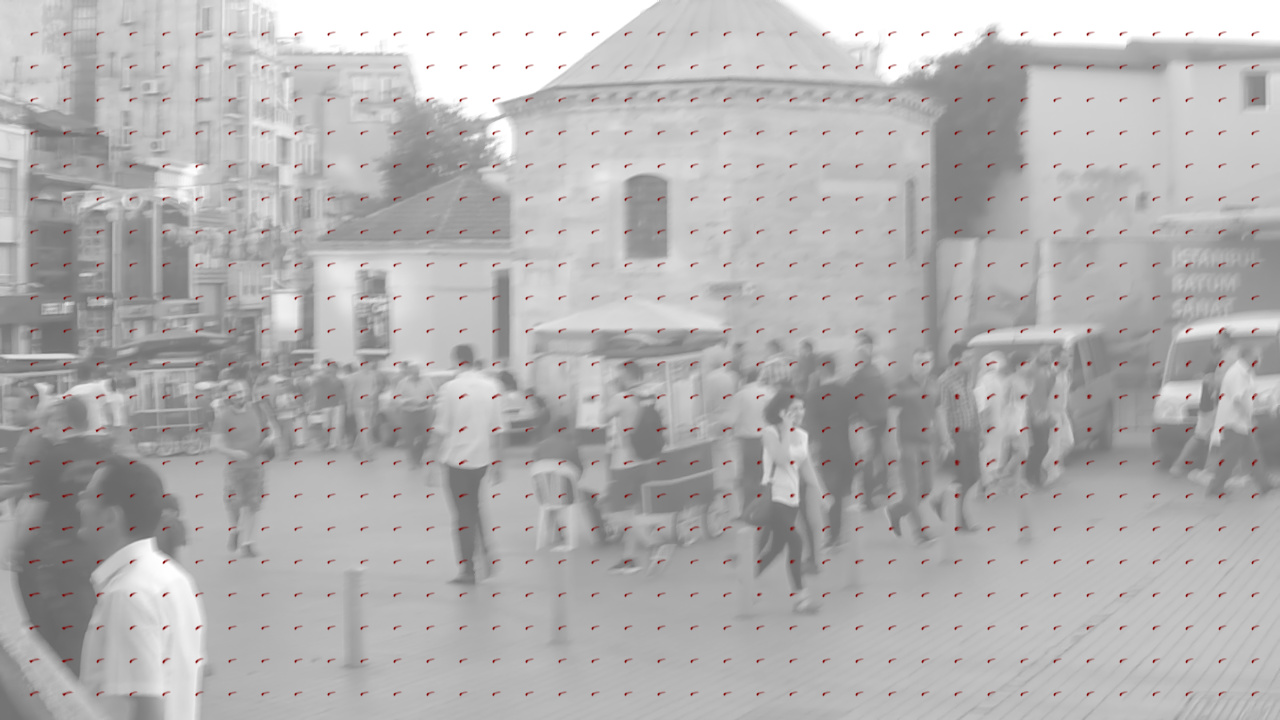}  \\
    
    \includegraphics[trim=50 320 700 50, clip,width=0.152\textwidth]{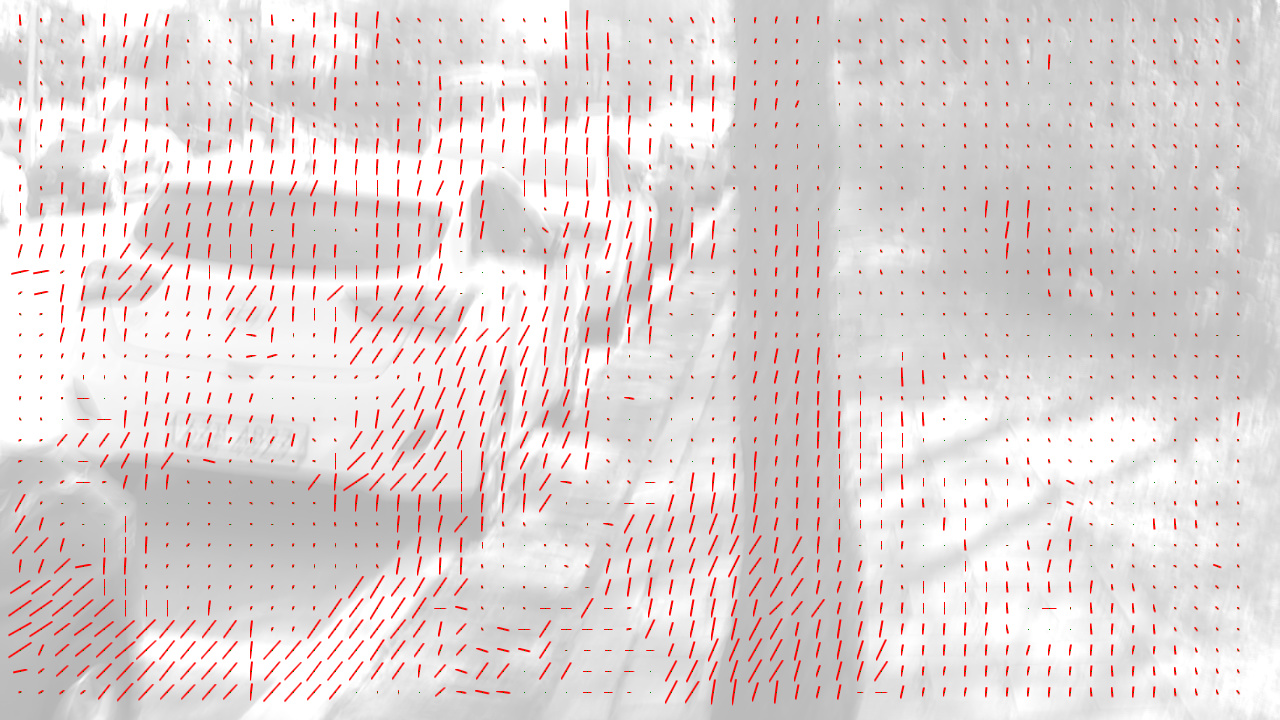}   &
      \includegraphics[trim=50 320 700 50, clip,width=0.152\textwidth]{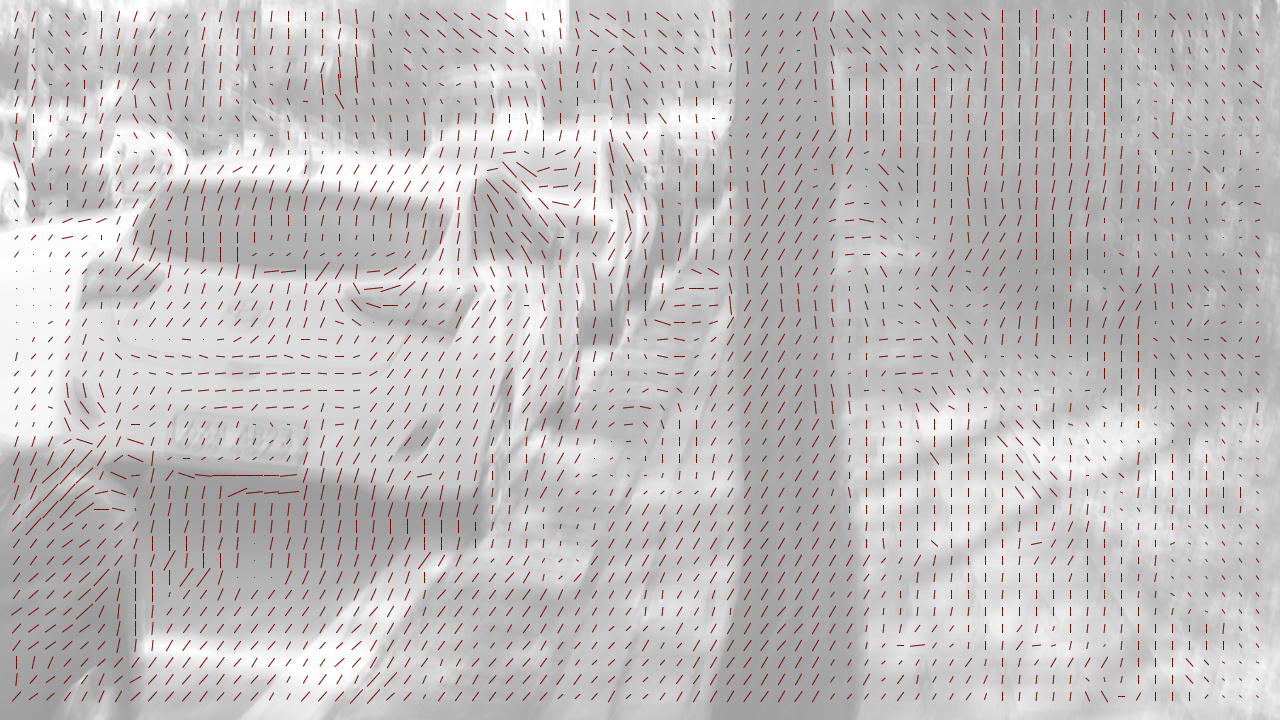} &  
        \includegraphics[trim=50 320 700 50, clip,width=0.152\textwidth]{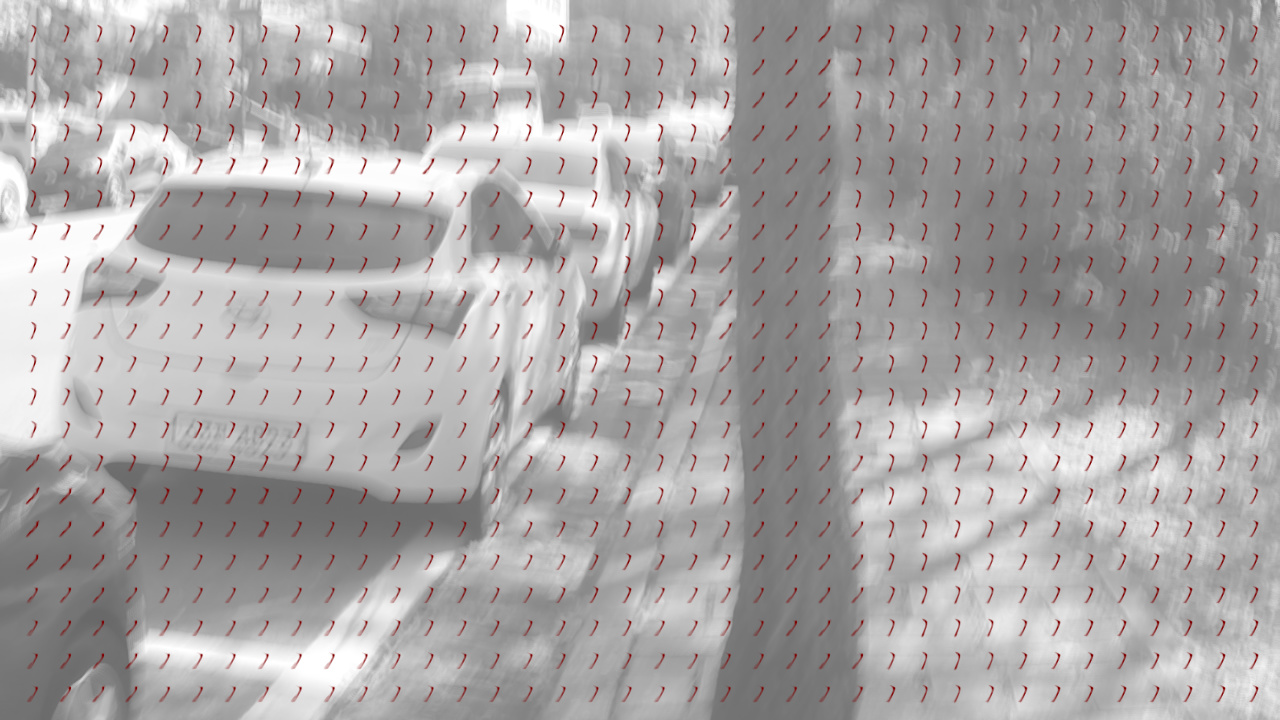}\\
    
    \includegraphics[trim=10 100 700 150, clip,width=0.152\textwidth]{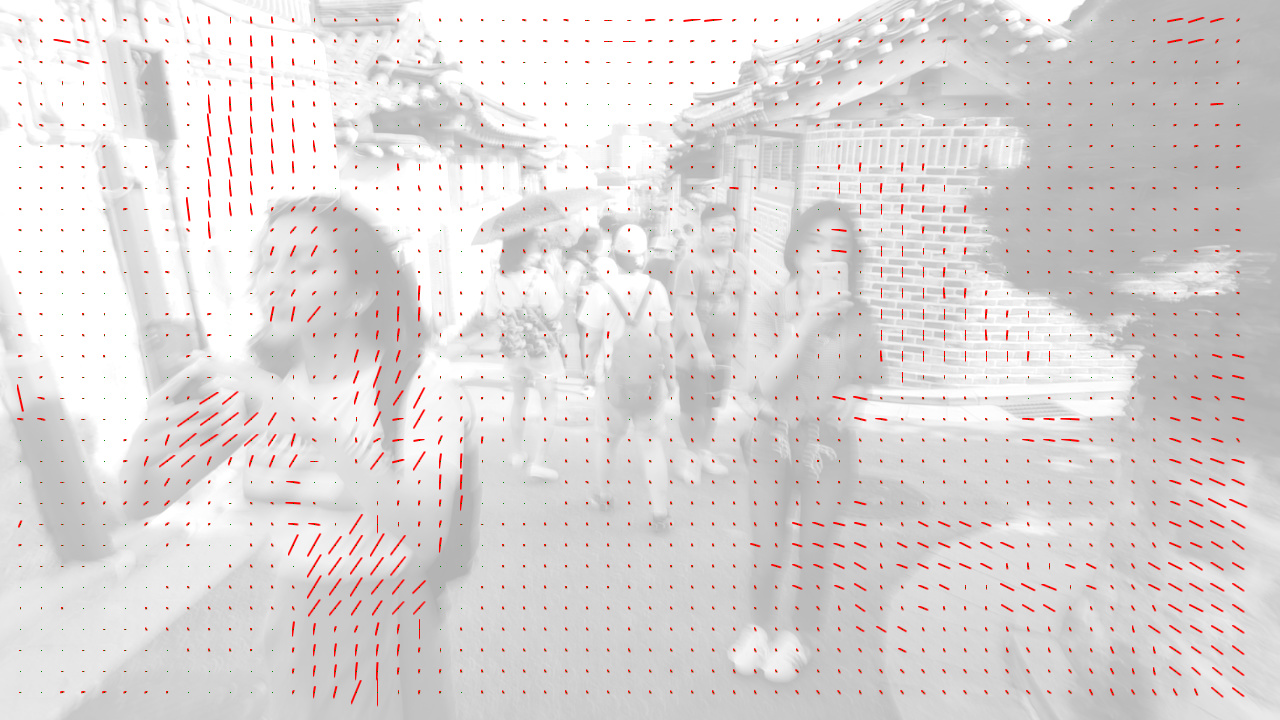}&
     \includegraphics[trim=10 100 700 150, clip,width=0.152\textwidth]{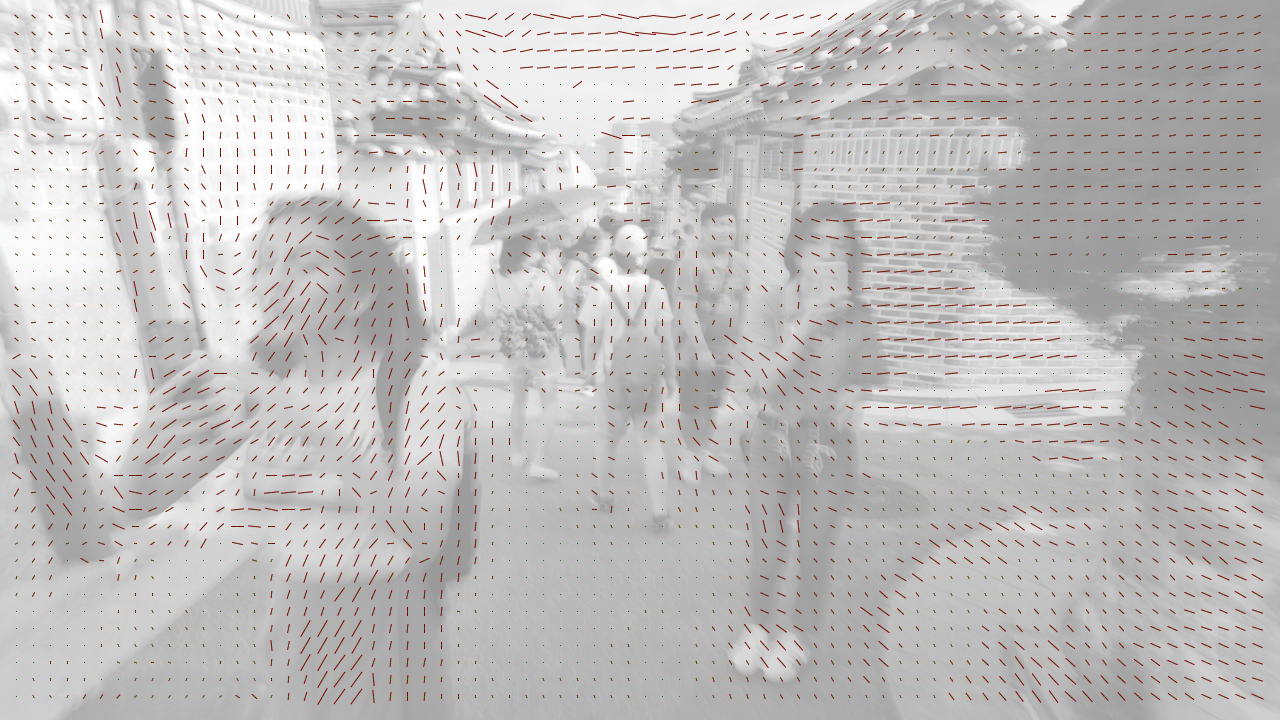}&   
     \includegraphics[trim=10 100 700 150, clip,width=0.152\textwidth]{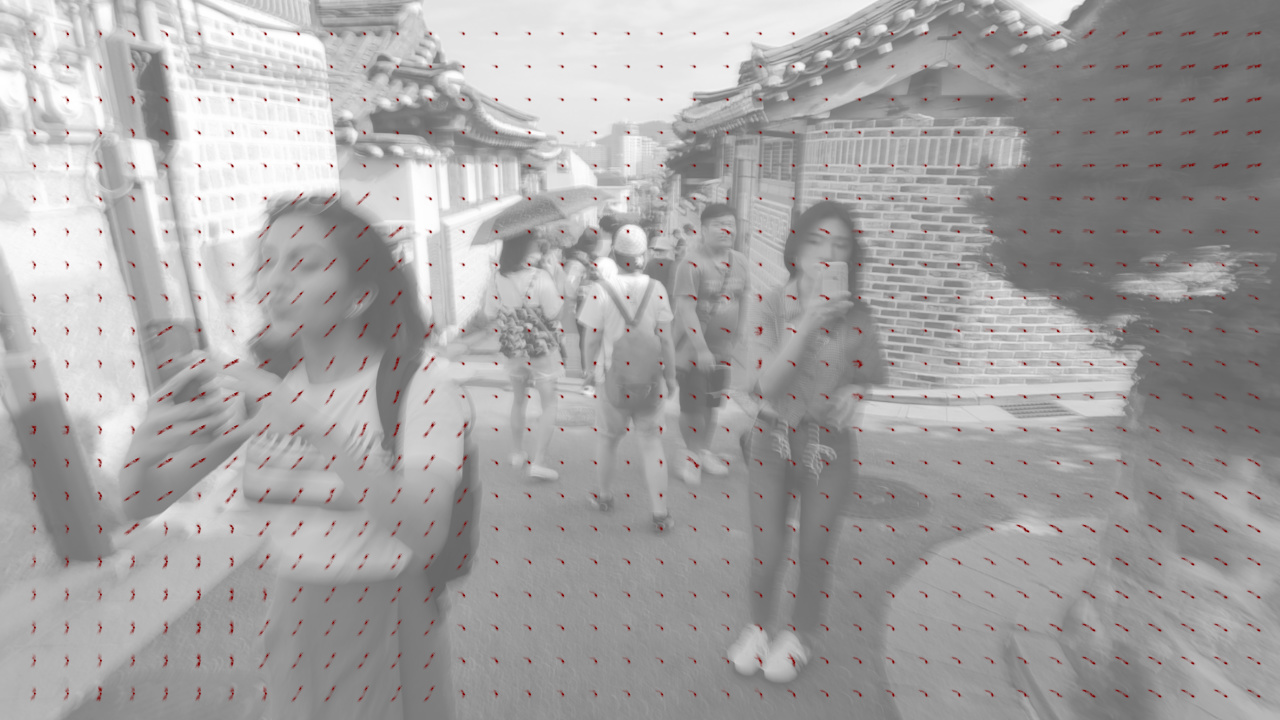} \\  
        \includegraphics[trim=10 100 10 10, clip,width=0.152\textwidth]{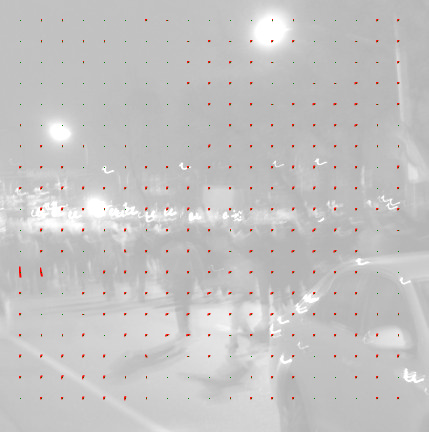}    &
     \includegraphics[trim=10 100 10 10, clip,width=0.152\textwidth]{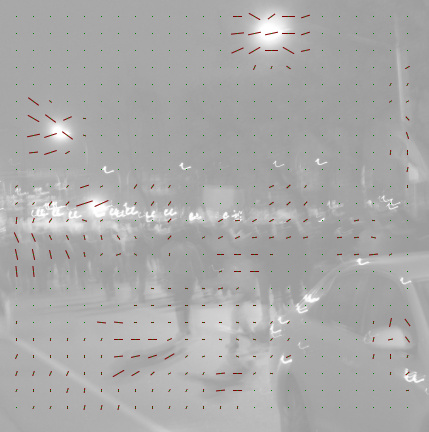}  &   
      \includegraphics[trim=10 100 10 10, clip,width=0.152\textwidth]{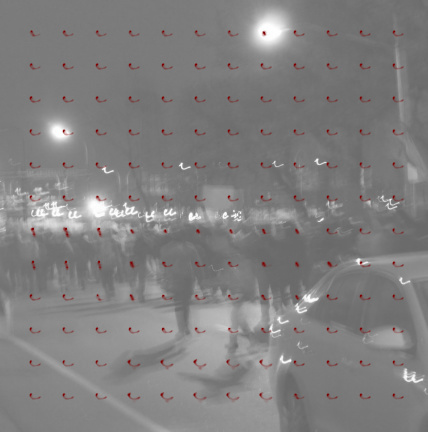}
  \end{tabular}
  \caption{\textbf{Visual comparison of non-uniform motion blur kernel estimation.} From left to right: Gong \etal~\cite{gong2017motion}, Sun \etal~\cite{sun2015learning} and ours. From top to bottom: two examples from the GoPro dataset~\cite{Nah_2017_CVPR}, one from REDs~\cite{Nah_2019_CVPR_Workshops_REDS} and one Lai's dataset~\cite{lai2016comparative}. Best viewed in electronic format. }
  \label{fig:KernelEstimationExamples}
\end{figure}

%%%%%%%%%%%%%%%%%%%%%

\begin{figure}[t]
    \centering
    \includegraphics[width=0.45\textwidth]{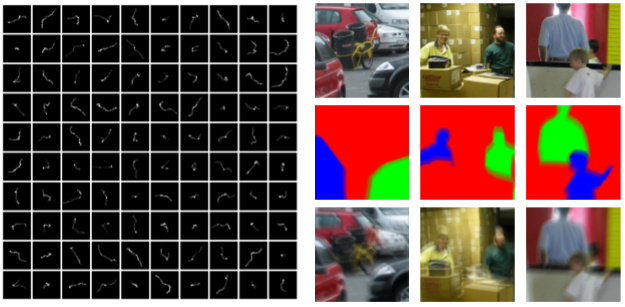}
    \caption{\textbf{Synthetic dataset for learning non-uniform motion kernel estimation.} Left: example random kernels. Right: different segments of the scene are convolved with different random kernels and are mixed with the (also convolved) segmentation masks.}
    \label{fig:dataset_gen}
\end{figure}
%%%%%%%%%%%%%%%%%%%%
\section{Synthetic Dataset Generation}\label{subsec:DatasetGen}
We build a synthetic dataset consisting of 5,888 images from the ADE20K semantic segmentation dataset~\cite{zhou2017scene}. To generate random motion kernels, we use a camera-shake kernel generator ~\cite{gavant2011physiological,delbracio2015removing} based on physiological hand tremor data and pre-compute 500,000 kernels with support smaller than $K\times K$, cf. Figure~\ref{fig:dataset_gen}.
In our experiments, we observed that training on this synthetic data generalizes remarkably well to real photographs with different types of scenes and motion. 

More specifically, for a random sharp image $\mathbf{u}$, we perform a convolution of the image with a random kernel $\mathbf{k}$. Additionally, each segmented object (if any) and its corresponding mask is convolved with a different random kernel. This ensures a soft transition between different blurry regions. To simplify, a maximum of three segmentation masks with a minimum size of 400 pixels are considered for each image. Finally, for each image, we obtain a tuple $\big(\mathbf{u}^{GT}, \mathbf{v}^{GT}, \{\mathbf{k}\}^{GT}, \{\mathbf{m}\}^{GT}\big)$ containing the sharp image, blurry image and the pairs of ground truth kernels and masks applied to generate the blurry image.  %To ensure a soft transition between different blurry regions, each segmentation mask is convolved with its corresponding kernel. 

\noindent\textbf{Data Augmentation}. %Blurry images may have saturated pixels. Generation of  examples with the procedure explained above is limited because  sharp image values are in the range [0,1] and the maximum value of the simulated blurry images is usually lower than the maximum of the corresponding sharp ones. 
To simulate latent scenes with saturating light sources, sharp images are first converted to \textit{hsv} color space and then the histogram is transformed by multiplying the {\em v}-channel by a random value between $[0.5, 1.5]$.  The transformed image is converted back to the  \textit{rgb} color space, and the blurry image is generated as described above. Finally, the blurry image is clipped to the $[0,1]$ range. We refer to Section~\ref{sec:app:dataset_generation}  for more details and examples from our dataset.

%\section{Learning the Motion Blur Kernels}
\section{Motion Blur Field Estimation}
We use a CNN to estimate, from a given input blurry image, both the $B$ basis motion kernels $\{\mathbf{k}^b\}_{b=1,...,B}$ and the mixing coefficients $\{\mathbf{m}^b\}_{b=1,...,B}$. Building upon recent work in kernel prediction networks~\cite{xia2019basis}, the network is composed of a shared backbone and two generator heads. We refer to Figure~\ref{fig:architecture} for an overview of the architecture. 
%The first decoder outputs a global per-image kernel basis of size $K\times K \times B$ (i.e., $B$ basis elements of size $K\times K$). The second decoder outputs $B$ maps of mixing coefficients of the same spatial resolution as the input image. 
Normalization of the blur kernels and the mixing coefficients is achieved by using Softmax layers.
In our experiments, we used $K=33$ and $B=25$. An ablation study exploring different values of $B$ is given in Table~\ref{tab:ablationstudy}.

\subsection{Objective Function}
%We propose two reconstruction losses to train the generators of basis and mixing coefficients. 

\noindent {\bf Reblur Loss.} Given a blurry image $\mathbf{v}^{GT}$, we aim to find the global  kernel basis  $\{\mathbf{k}^b\}$ and mixing coefficients $\{\mathbf{m}^b\}$ that minimize
\begin{equation} \label{eq:reblur_loss}
    \mathcal{L}_{reblur} = \sum_i  w_{i} ( v_{i} - v^{GT}_{i} )^2,
\end{equation}
where the $v_i$ are computed using~\eqref{eq:model_sat}. The re-blurring of the sharp image can be done efficiently by first convolving it with each of the kernels in the base, and then doing an element-wise blending of the $B$ resulting images with the corresponding mixing coefficients. To prevent a single kernel from dominating the loss, weights $w_{i}$ are computed as the inverse of the number of pixels that belong to the same segmented object. \vspace{.3em}
%
%Given corresponding blurry and sharp images, we first apply the predicted motion blur field to the sharp image. %Note that 
%The re-blurring of the sharp image can be done efficiently by first convolving it with each of the kernels in the base, and then doing an element-wise product of the $B$ resulting images with the corresponding mixing coefficients, and then adding the results. More precisely, given a blurry image $\mathbf{v}^{GT}$, we aim to find the global \emph{kernel basis} $\{\mathbf{k}^b\}$ and mixing coefficients $\{\mathbf{m}^b\}$ that minimize
%\begin{equation} \label{eq:reblur_loss}
%    \mathcal{L}_{reblur} = \sum_i  w_{i} ( v_{i} - v^{GT}_{i} )^2,
%\end{equation}
%where the $v_i$ are computed using~\eqref{eq:model_sat}.
%
%, and $w_{i,j}$ is a scalar used to weigh different regions in the image. %The effect of these weights will become more clear in Section~\ref{subsec:DatasetGen}, when training on synthetic data.  
%To prevent a single kernel from dominating the losses \eqref{eq:reblur_loss} and \eqref{eq:kernel_loss}, weights $w_{i}$ are computed as the inverse of the number of pixels that belong to the same segment.
%

\noindent {\bf Kernel Loss.}  Ground truth pixel-wise motion blur kernels are compared to the predicted per-pixel kernels. %This is the case when using synthetic blurry images, as described in Section~\ref{subsec:DatasetGen}. 
Given a ground truth per-pixel blur kernel $\{\mathbf{k}^{GT}_{i}\}$, the computed kernel basis $\{\mathbf{k}^b\}$ and mixing coefficients $\{m^b_{i}\}$, the \emph{kernel loss} is defined as:
\begin{equation} \label{eq:kernel_loss}
    \mathcal{L}_{kernel} = \sum_i  w_{i} \left\Vert \sum_{b=1}^Bm^b_{i}\mathbf{k}^b-\mathbf{k}^{GT}_{i} \right\Vert_p,
\end{equation}
where the weights $w_{i}$ are the same as in \emph{reblur loss}, and $p$ is 1 or 2, depending on the training strategy described next.

\subsection{Model Training}
We train the CNN using the sum of the \emph{reblur loss} \eqref{eq:reblur_loss} and the \emph{kernel loss} \eqref{eq:kernel_loss} with equal weights. %Training only with the \emph{Reblur} term would make the problem more challenging. % Note that the model can be trained without knowledge of the kernels, using only the  \emph{Reblur loss}. 
%Adding the \emph{Kernel loss} improves the convergence.   
Training a model to predict a per-pixel kernel estimation is a difficult task. In our case, the model needs to figure out an image-specific low-rank decomposition to approximate all the kernels present in the image. In our experiments we observed very slow convergence and only started to see well-shaped kernels after around 200 epochs. We found that using an $L^2$-norm on the kernels loss in~\eqref{eq:kernel_loss} was adequate to find a first approximation of the model. After 300 epochs, we switched to the more robust $L^1$-norm, which is harder to optimize but allows us to recover sharper kernels. In total, we trained our model for 1200 epochs using image patches of $256\times256$ pixels. Additional details on the training procedure and hyper-parameters are given in Section~\ref{sec:app:training}.

\section{Qualitative Evaluation}
In Figures~\ref{fig:KernelsAndMasks} and \ref{fig:KernelsAndMasks2}, we show examples of the set of kernel basis and corresponding mixing coefficients predicted for different images. Note that the predicted basis is image-dependent, specially for those kernel basis elements that are more active in the decomposition (i.e. the corresponding mixing coefficients have high values throughout the scene). 

Figure~\ref{fig:KernelEstimationExamples} shows some examples of non-uniform blur kernel estimates obtained by our method. We visually compare them with the results of two other existing DL-based non-uniform motion blur estimation methods, Gong \etal \cite{gong2017motion} and Sun \etal~\cite{sun2015learning}.
Despite being trained on synthetically blurred images, our method generalizes remarkably well to real blurry images (last column), as well as blurry images synthesized from video sequences as in GoPro~\cite{Nah_2017_CVPR} and REDs~\cite{Nah_2019_CVPR_Workshops_REDS} datasets.

 Our model is able to characterize different types of camera and object motions. Note also that the motion blur kernels estimated by the compared methods tend to correlate with the scene geometry, instead of revealing the underlying motion field. Moreover, our model predicts continuous free-form arbitrary motion kernels, whereas~\cite{sun2015learning} and~\cite{gong2017motion} are restricted to linear ones.% Further qualitative estimation results are shown in Figure~\ref{fig:blur_detection}.

\begin{figure*}[t!]
    \centering
            \setlength{\tabcolsep}{2pt}
        %\begin{tabular}{*{9}{c}}
        \begin{tabular}{c|cccc|cccc}
        & \multicolumn{4}{c|}{  End-to-end } & \multicolumn{4}{c}{  Model-based} \\[.25em]
Original &   D-GAN2 \cite{kupyn2019deblurgan} &  SRN~\cite{tao2018scale} & RealBlur \cite{rim_2020_ECCV} & MPRNet \cite{Zamir2021MPRNet} &  Whyte  \cite{whyte2010nonuniform}& Sun~\cite{sun2015learning}  & Gong~\cite{gong2017motion} & Ours  \\
\includegraphics[width=0.1\textwidth]{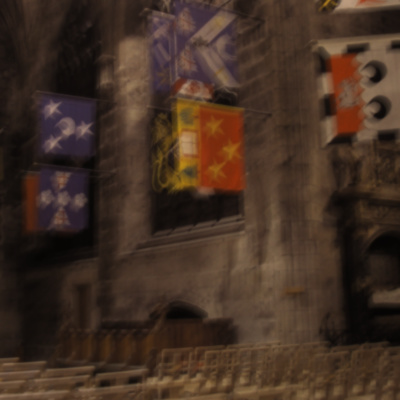} &

\includegraphics[width=0.1\textwidth]{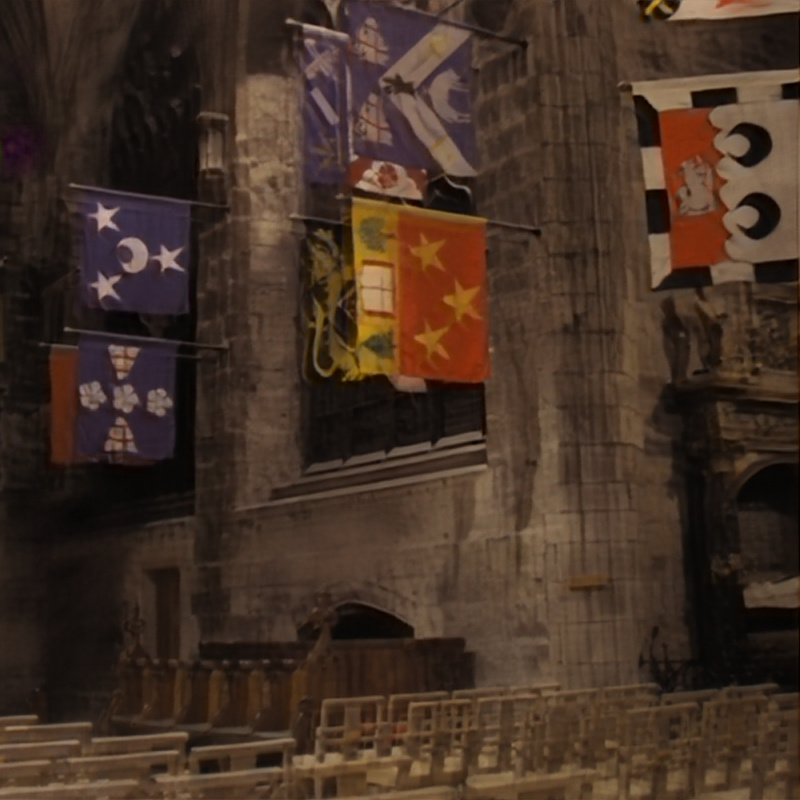} &   
\includegraphics[width=0.1\textwidth]{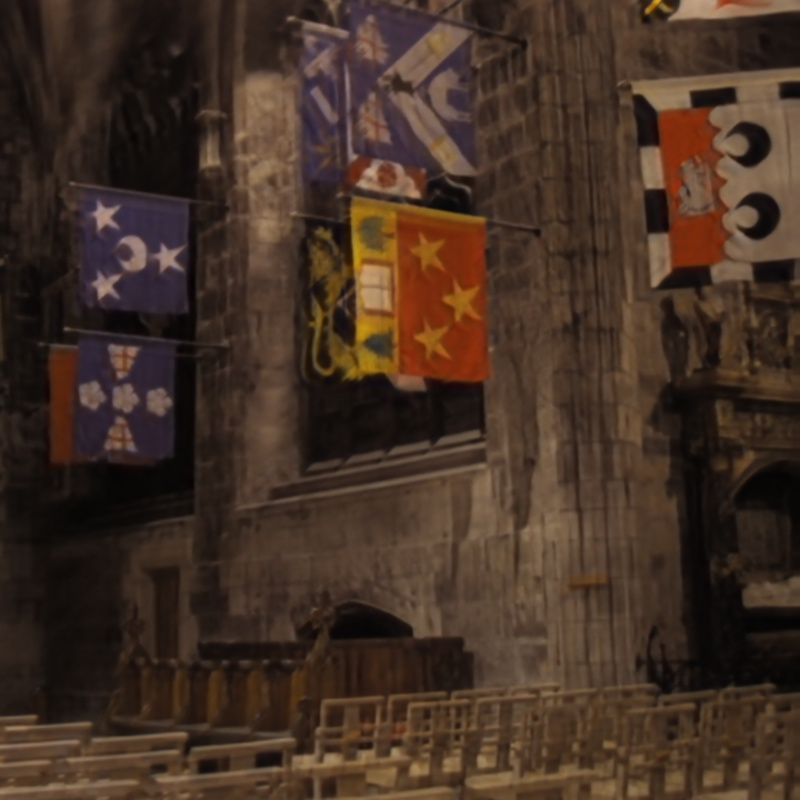} &
\includegraphics[width=0.1\textwidth]{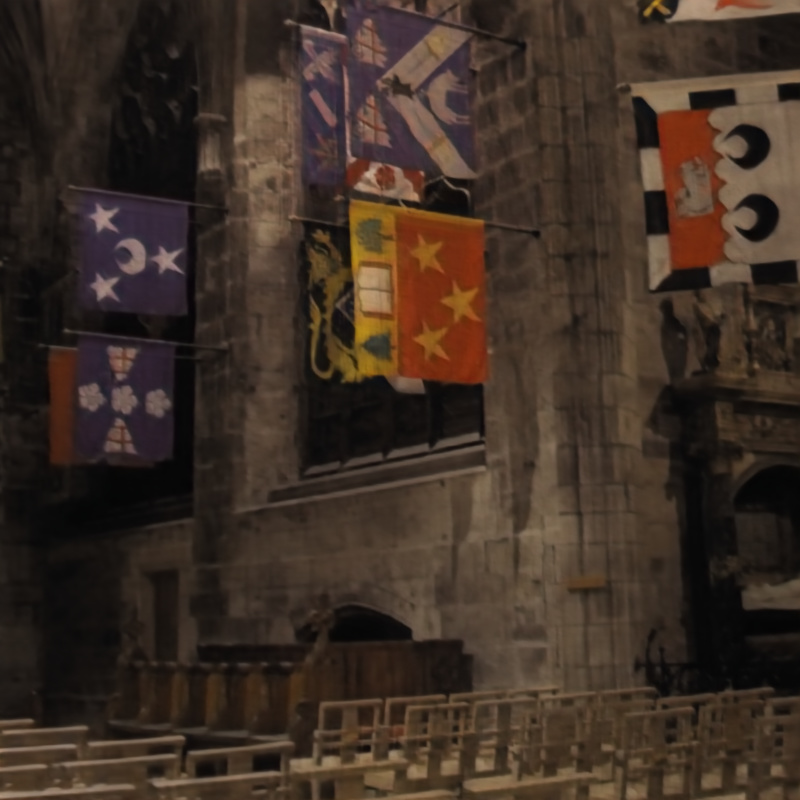} &
\includegraphics[width=0.1\textwidth]{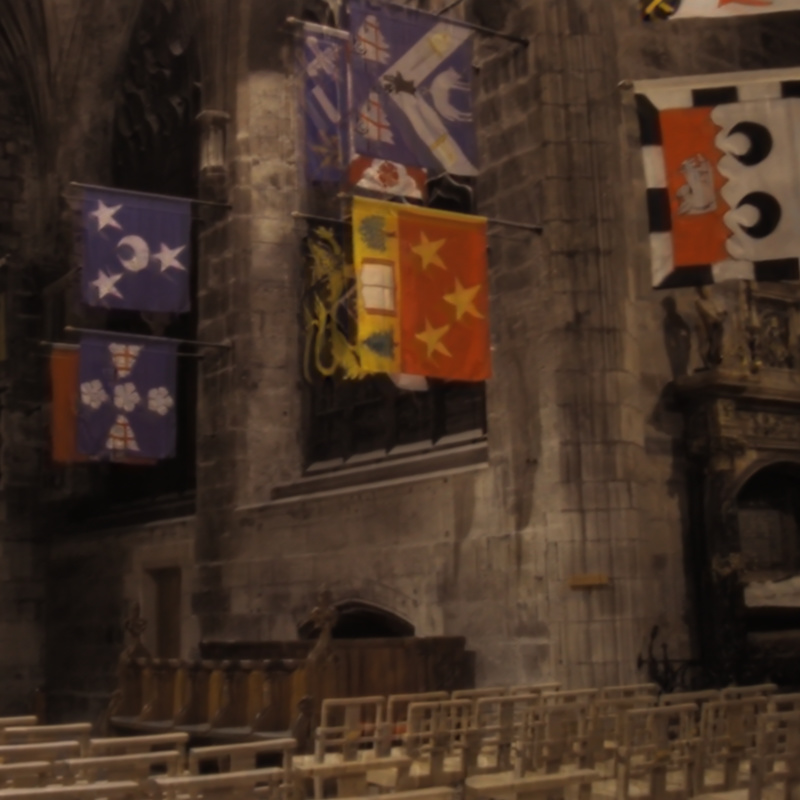}
& 
\includegraphics[width=0.1\textwidth]{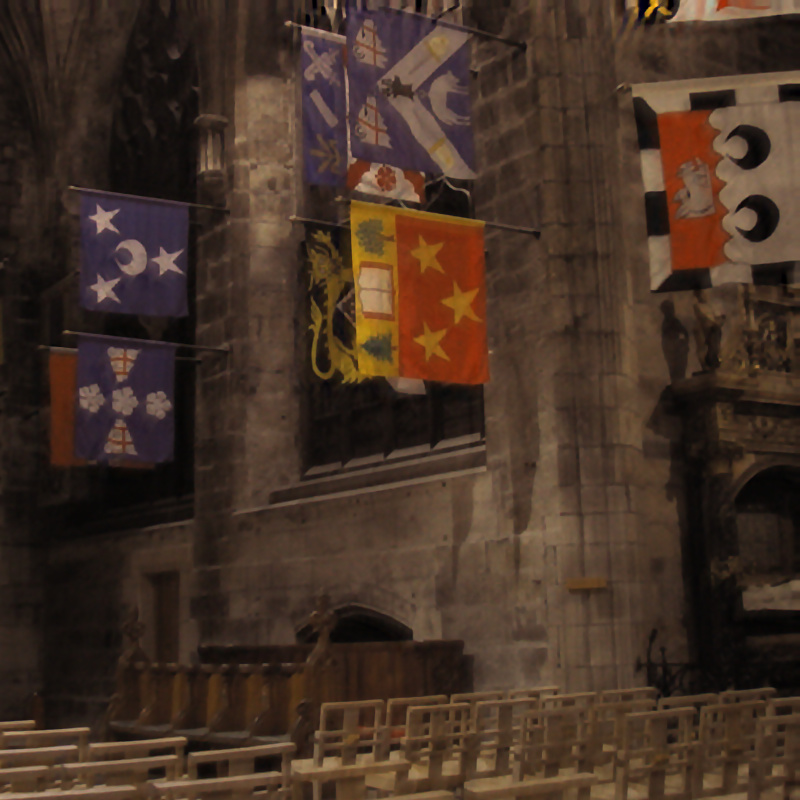}
&
\includegraphics[width=0.1\textwidth]{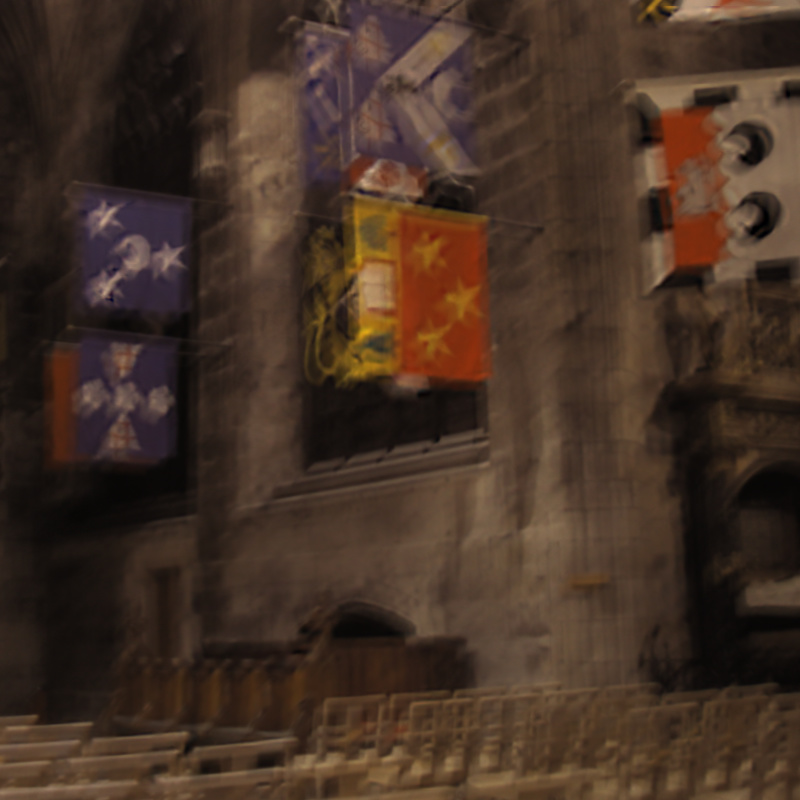}
& 
\includegraphics[width=0.1\textwidth]{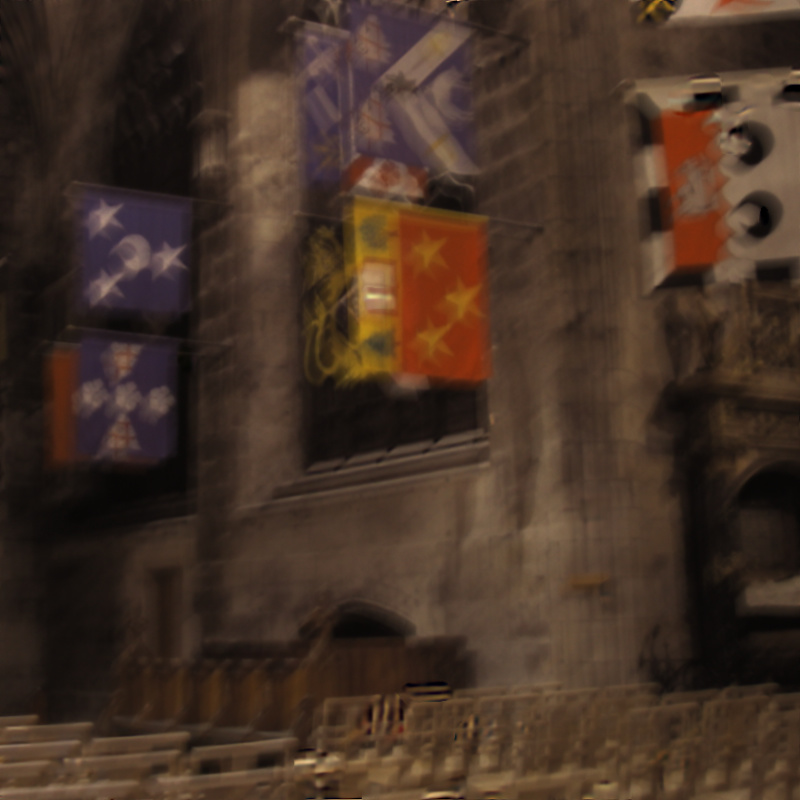} &
\includegraphics[width=0.1\textwidth]{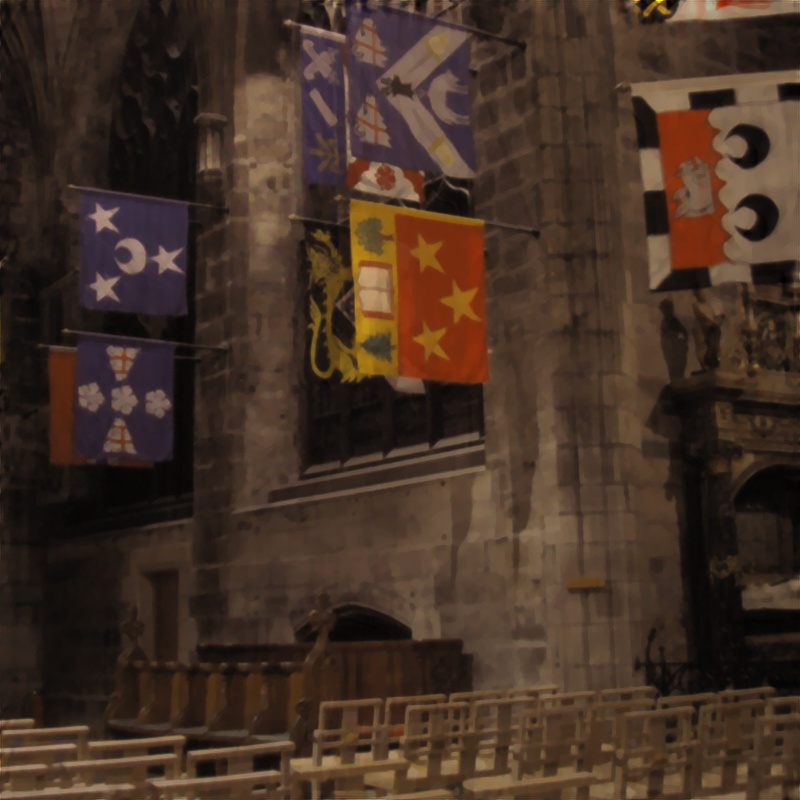}
\\ 
\includegraphics[trim=150 260 125 25, clip, width=0.1\textwidth]{latex/Kohler/Blurry/Blurry1_1.jpg} &

\includegraphics[trim=300 520 250 50, clip,width=0.1\textwidth]{latex/Kohler/DeblurGANv2Inception/Blurry1_1.jpg} & 
\includegraphics[trim=300 520 250 50, clip,width=0.1\textwidth]{latex/Kohler/SRN/Blurry1_1.jpg} &
\includegraphics[trim=300 520 250 50, clip,width=0.1\textwidth]{latex/Kohler/RealBlurJ_pre_trained+GOPRO+BSD500/Blurry1_1.jpg} &
\includegraphics[trim=300 520 250 50, clip,width=0.1\textwidth]{latex/Kohler/MPRNet/Blurry1_1.jpg} & 
\includegraphics[trim=300 520 250 50, clip,width=0.1\textwidth]{latex/Kohler/Whyte/Blurry1_1_result_whyteMAP-krishnan.jpg}
&
\includegraphics[trim=300 520 250 50, clip,width=0.1\textwidth]{latex/Kohler/Sun/Blurry1_1.jpg}
&
\includegraphics[trim=300 520 250 50, clip,width=0.1\textwidth]{latex/Kohler/gong/Blurry1_1_deblurred.jpg} &
\includegraphics[trim=300 520 250 50,
clip,width=0.1\textwidth]{latex/Kohler/our_no_FC/Blurry1_1_restored.jpg}
\\ 
27.58  dB & 30.88 dB &  31.51  dB & 33.06 dB & 33.37 dB &   34.00 dB  & 27.78 dB  &  27.45 dB&  35.19 dB \\

\includegraphics[width=0.1\textwidth]{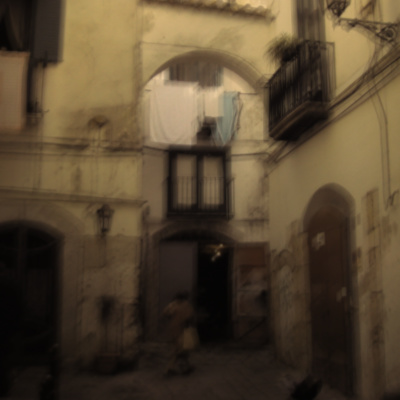} &

\includegraphics[width=0.1\textwidth]{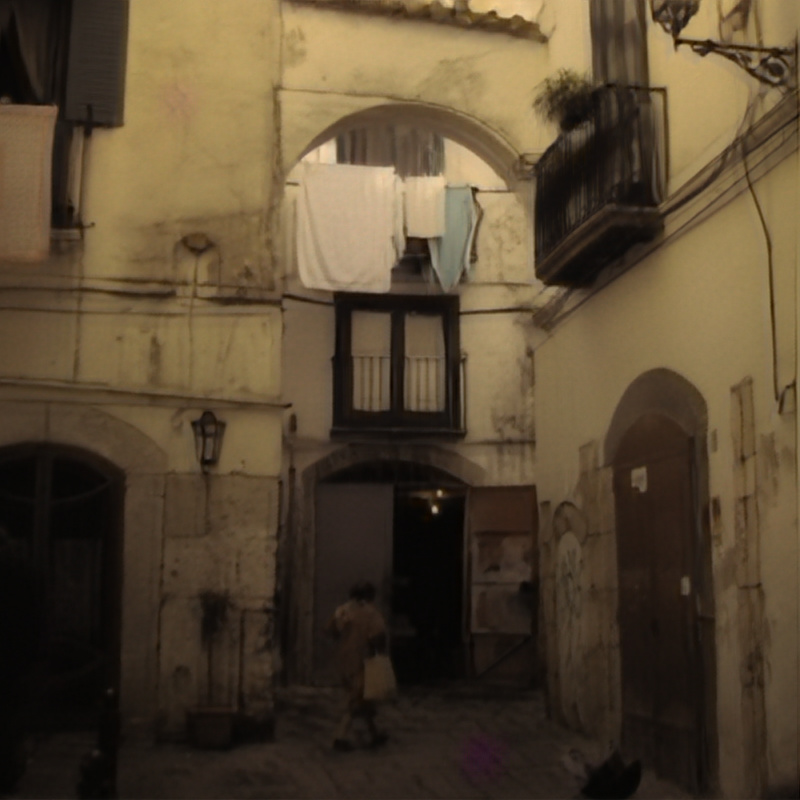} &   
\includegraphics[width=0.1\textwidth]{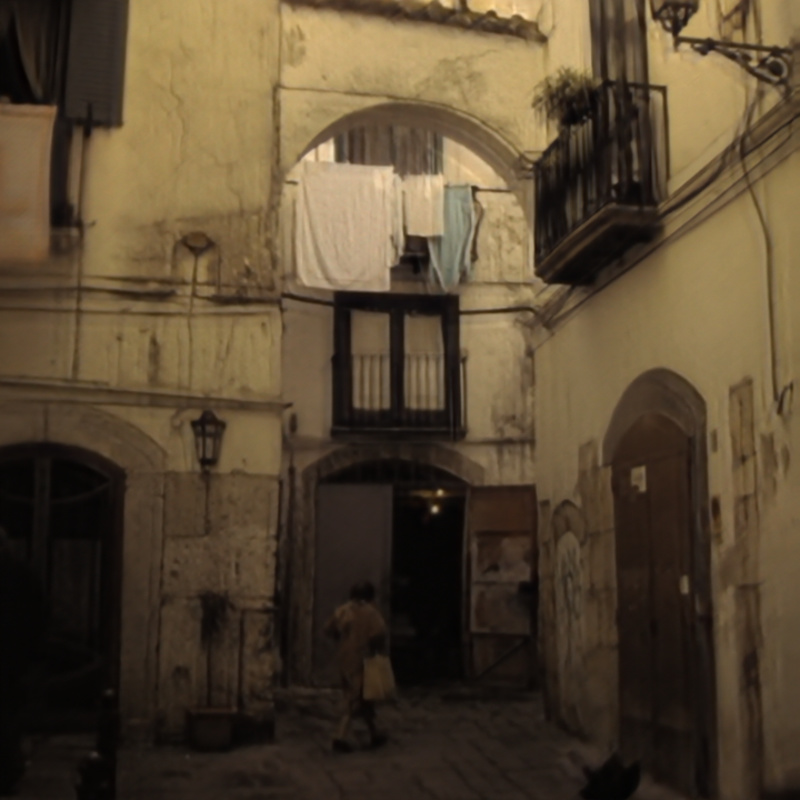} &
\includegraphics[width=0.1\textwidth]{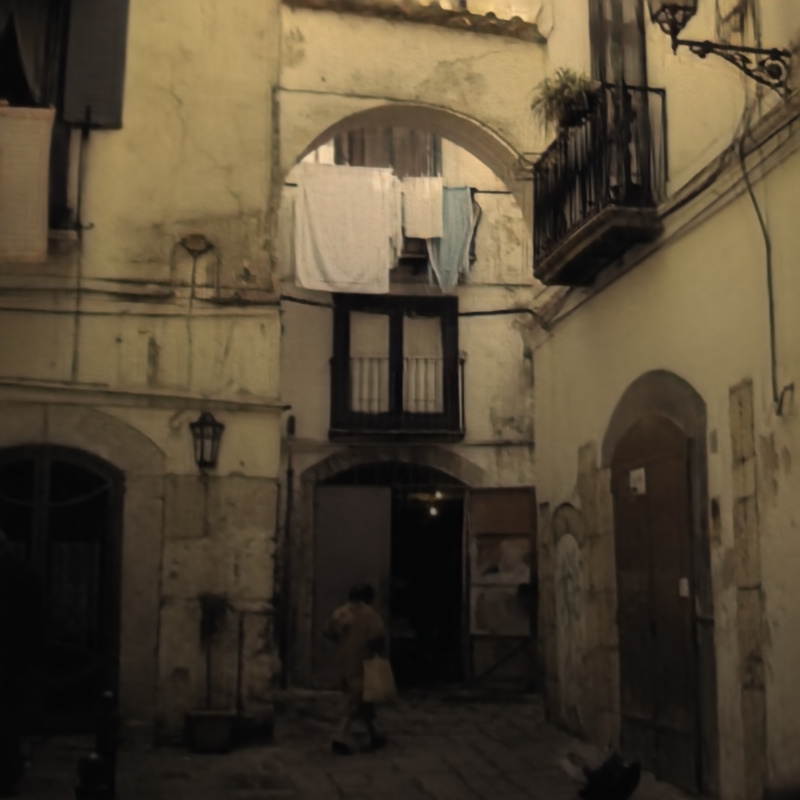} &
\includegraphics[width=0.1\textwidth]{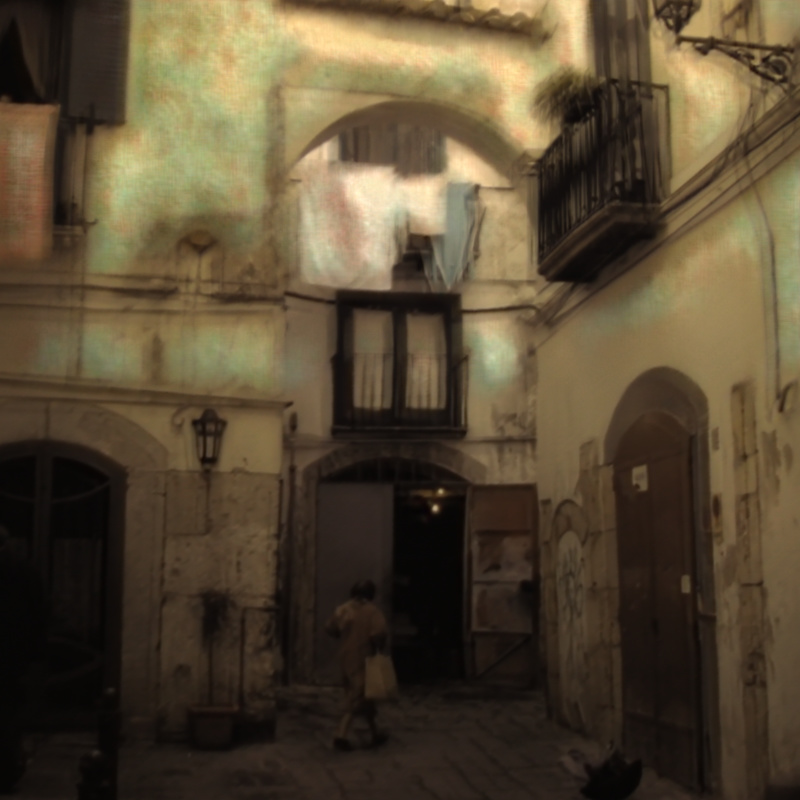} & 
\includegraphics[width=0.1\textwidth]{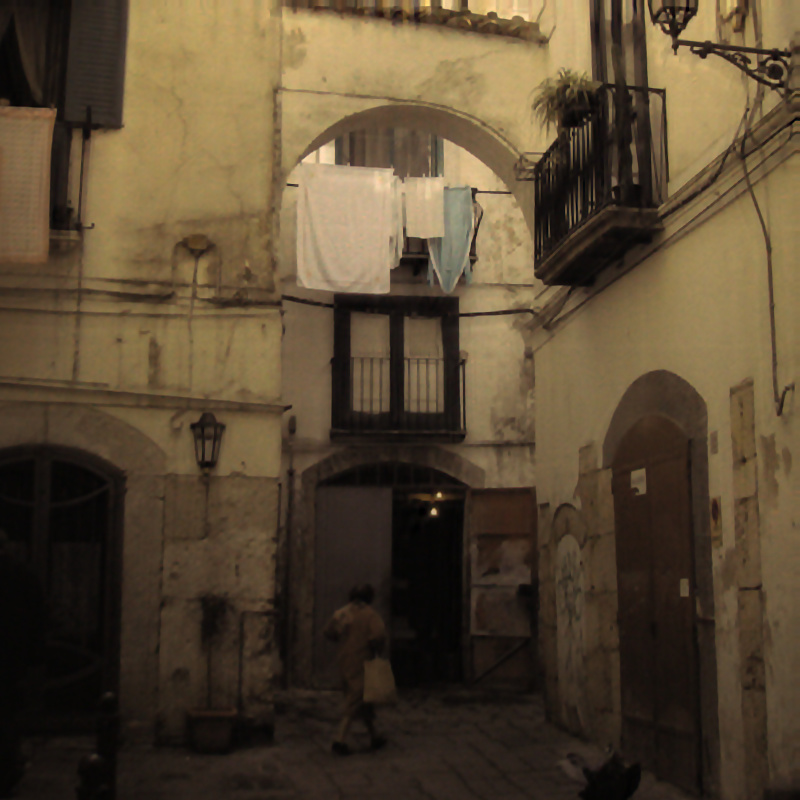} 
&
\includegraphics[width=0.1\textwidth]{latex/Kohler/SRN/Blurry3_1.jpg} 
&
\includegraphics[width=0.1\textwidth]{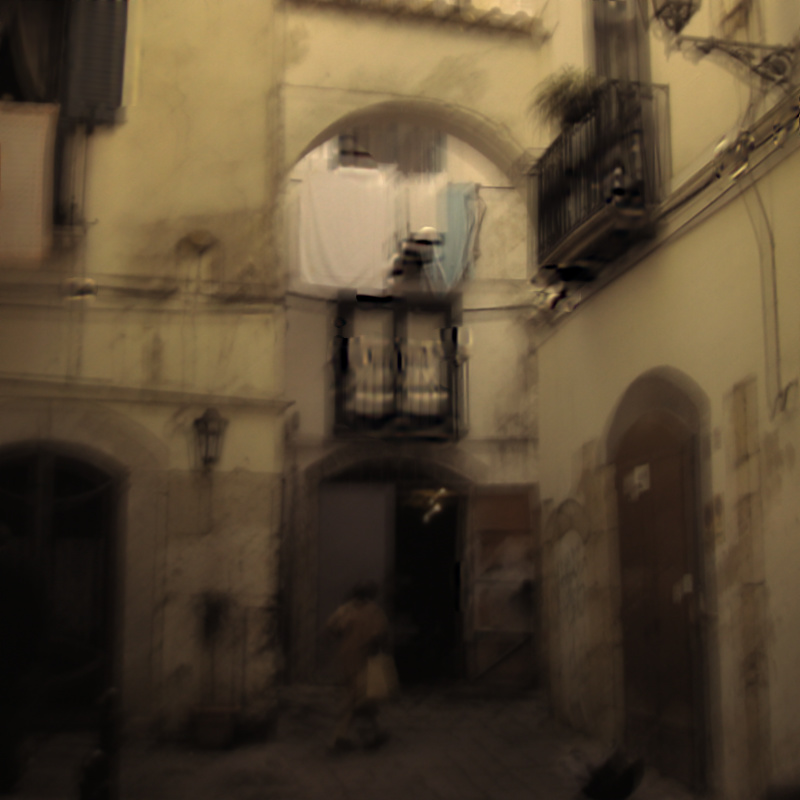} &
\includegraphics[width=0.1\textwidth]{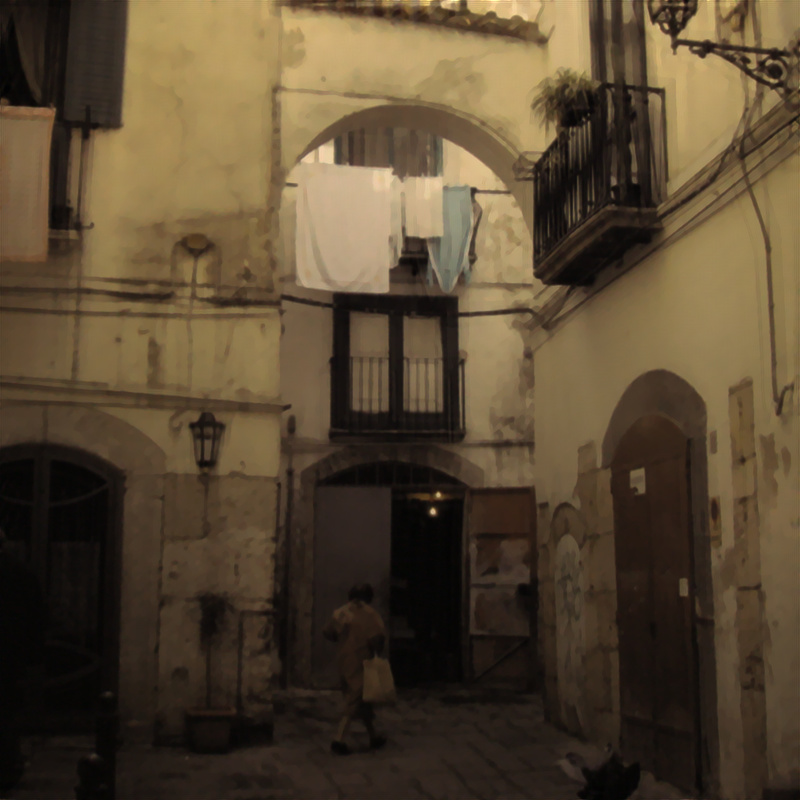}
\\ 
\includegraphics[trim=150 120 100 125, clip, width=0.1\textwidth]{latex/Kohler/Blurry/Blurry3_1.jpg} &
\includegraphics[trim=300 240 200 250, clip,width=0.1\textwidth]{latex/Kohler/DeblurGANv2Inception/Blurry3_1.jpg} &  
\includegraphics[trim=300 240 200 250, clip,width=0.1\textwidth]{latex/Kohler/SRN/Blurry3_1.jpg} &
\includegraphics[trim=300 240 200 250, clip,width=0.1\textwidth]{latex/Kohler/RealBlurJ_pre_trained+GOPRO+BSD500/Blurry3_1.jpg} &
\includegraphics[trim=300 240 200 250, clip,width=0.1\textwidth]{latex/Kohler/MPRNet/Blurry3_1.jpg}  & 
\includegraphics[trim=300 240 200 250, clip,width=0.1\textwidth]{latex/Kohler/Whyte/Blurry3_1_result_whyteMAP-krishnan.jpg}
&
\includegraphics[trim=300 240 200 250, clip,width=0.1\textwidth]{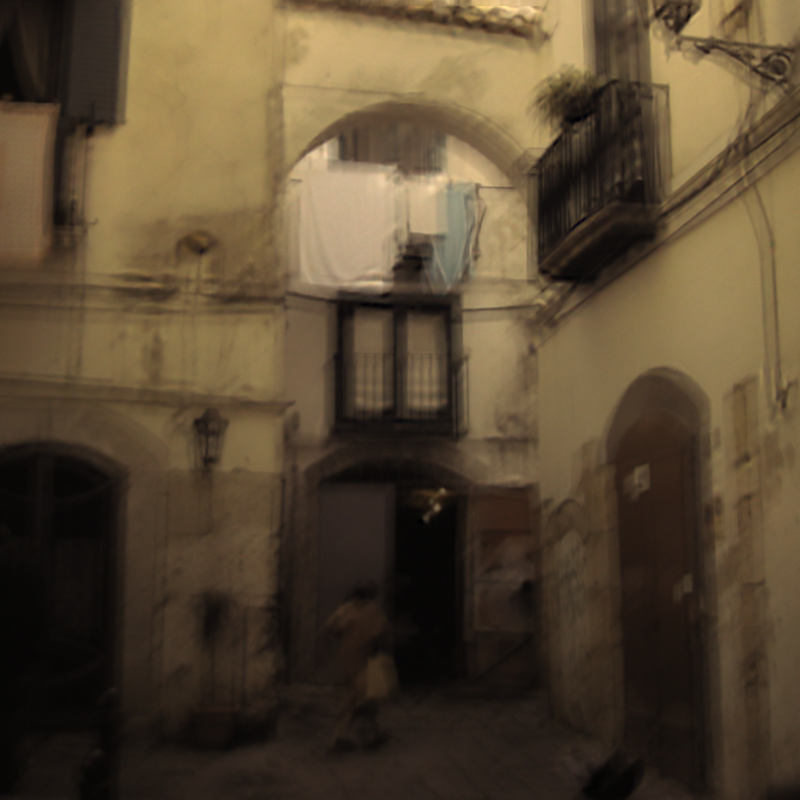}
& 
\includegraphics[trim=300 240 200 250, clip,width=0.1\textwidth]{latex/Kohler/gong/Blurry3_1_deblurred.jpg} &

\includegraphics[trim=300 220 200 250, clip,width=0.1\textwidth]{latex/Kohler/our_no_FC/Blurry3_1_restored.jpg}
\\ 

27.53 dB & 31.32 dB & 32.21 dB  & 33.11 dB & 26.54 dB &  36.25 dB & 27.94 dB & 27.55 dB & 34.93 dB  \\
    \end{tabular}
    \caption{\textbf{Qualitative comparison of different deblurring methods in K\"{o}hler Dataset~\cite{kohler2012recording} of  real blurred images.} %From left to right, DeblurGAN-v2, SRN, RealBlur and MPRNet  are DL end-to-end trained methods. Whyte \etal, Sun \etal,  Gong \etal and Ours are  model-based deblurring methods that  estimate a non-uniform blur motion field.  
    %The complete, full resolution images are available in the Appendix.  % JL: no me gusta esto, ya están en full resolution, es cuestión de hacer zoom
    }
    \label{fig:kohlerDataset}
\end{figure*}

\section{Application to Image Deblurring}

% We design the following strategy to assess the quality of the estimated motion blur kernels: {\em (i)} We compare the estimated kernels obtained by our method with the ones obtained by prior art ~\cite{gong2017motion,sun2015learning}; {\em (ii)} 

In this section, we demonstrate the use of our estimated motion kernels for solving the inverse problem of motion blur removal in real photographs.

\begin{figure*}[h]

    \setlength\tabcolsep{1.5pt} % default value: 6pt
    \centering
\begin{tabular}{cccccc}
\includegraphics[width=0.16\textwidth]{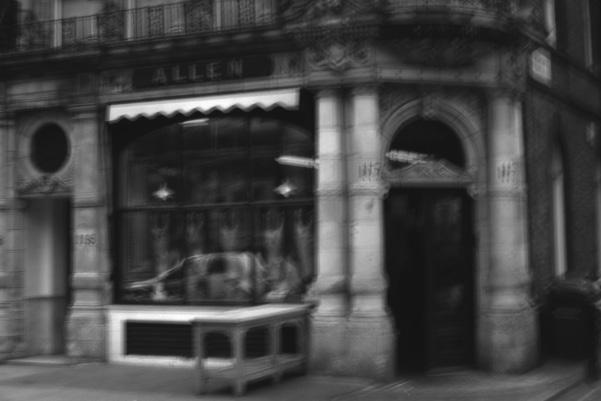}   &
\includegraphics[width=0.16\textwidth]{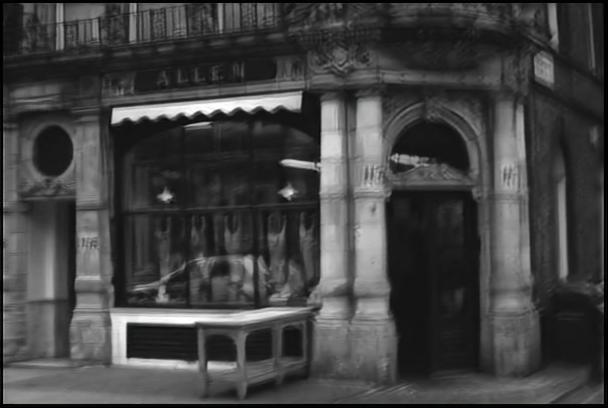} & 
\includegraphics[width=0.16\textwidth]{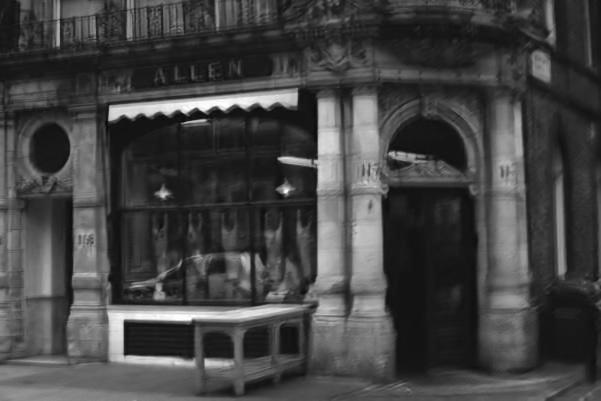}  &
\includegraphics[width=0.16\textwidth]{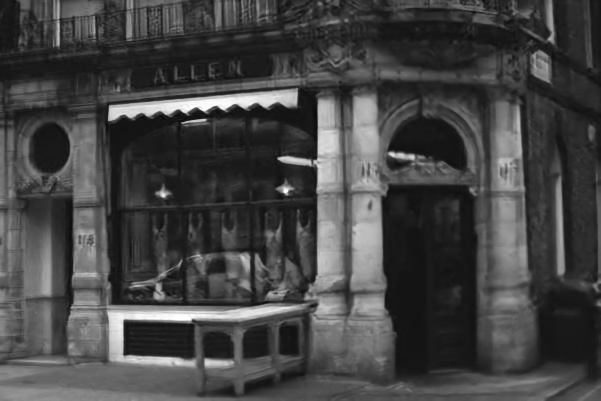}  &
\includegraphics[width=0.16\textwidth]{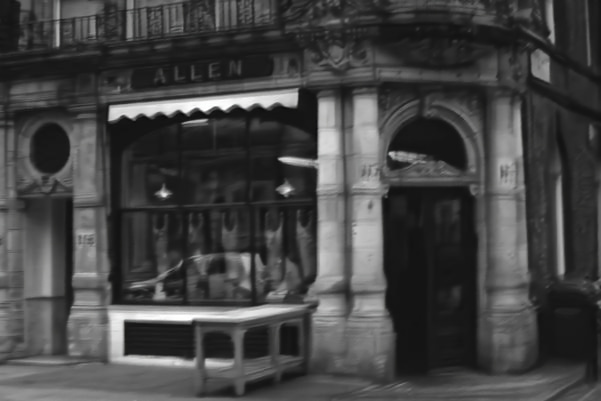}  &
\includegraphics[width=0.16\textwidth]{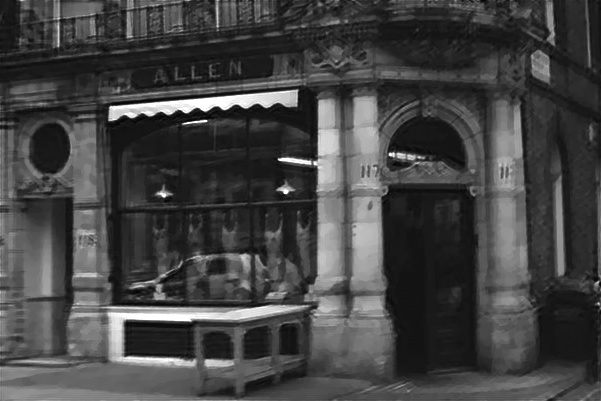}\\ 
\includegraphics[trim=50 140 450 220, clip,width=0.08\textwidth]{latex/Lai/Blurry/butchershop.jpg}  
\includegraphics[trim=325 210 175 150, clip,width=0.08\textwidth]{latex/Lai/Blurry/butchershop.jpg}  &
\includegraphics[trim=50 140 450 220, clip,width=0.08\textwidth]{latex/Lai/DMPHN/butchershop.jpg}  
\includegraphics[trim=325 210 175 150, clip,width=0.08\textwidth]{latex/Lai/DMPHN/butchershop.jpg}  &
\includegraphics[trim=50 140 450 220, clip,width=0.08\textwidth]{latex/Lai/SRN/butchershop.jpg}  
\includegraphics[trim=325 210 175 150, clip,width=0.08\textwidth]{latex/Lai/SRN/butchershop.jpg}  &
\includegraphics[trim=50 140 450 220, clip,width=0.08\textwidth]{latex/Lai/RealBlur/butchershop.jpg}
\includegraphics[trim=325 210 175 150, clip,width=0.08\textwidth]{latex/Lai/RealBlur/butchershop.jpg} &
\includegraphics[trim=50 140 450 220, clip,width=0.08\textwidth]{latex/Lai/MPRNet/butchershop.jpg}
\includegraphics[trim=325 210 175 150, clip,width=0.08\textwidth]{latex/Lai/MPRNet/butchershop.jpg} &
\includegraphics[trim=50 140 450 220, clip,width=0.08\textwidth]{latex/Lai/ours_iccv/butchershop_restored_no_gc.jpg}
\includegraphics[trim=325 210 175 150, clip,width=0.08\textwidth]{latex/Lai/ours_iccv/butchershop_restored_no_gc.jpg}\\

%%%%%%%%%%%
\includegraphics[width=0.165\textwidth]{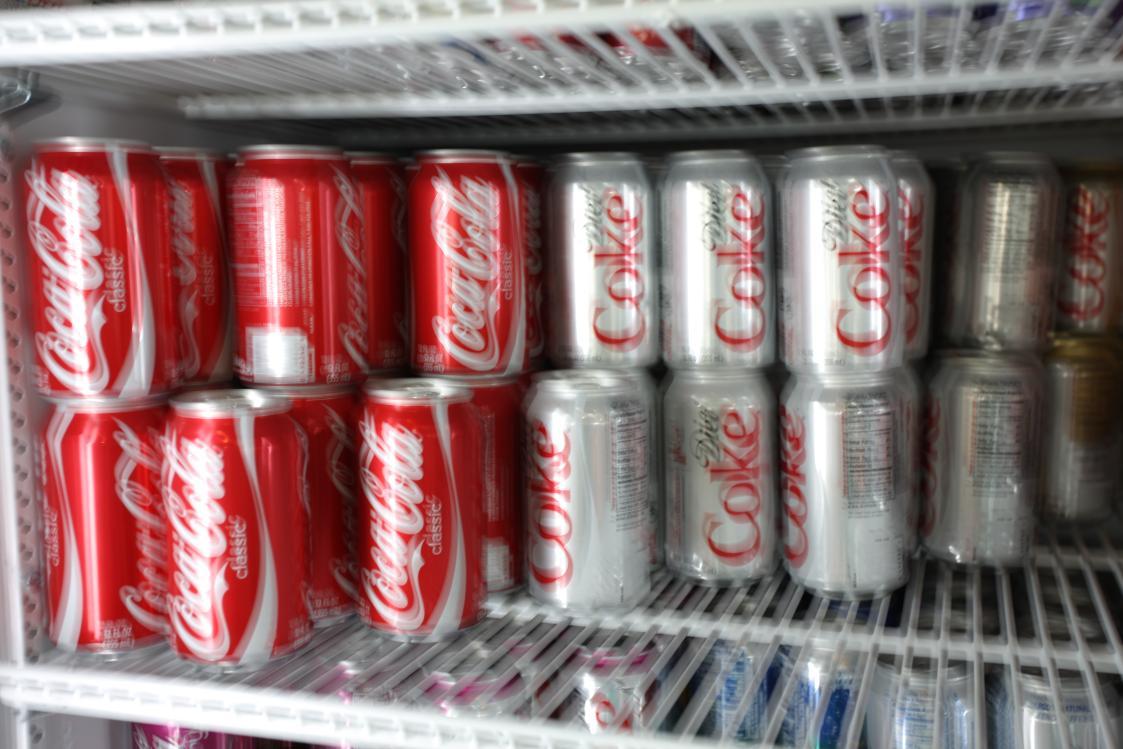}   &
\includegraphics[width=0.165\textwidth]{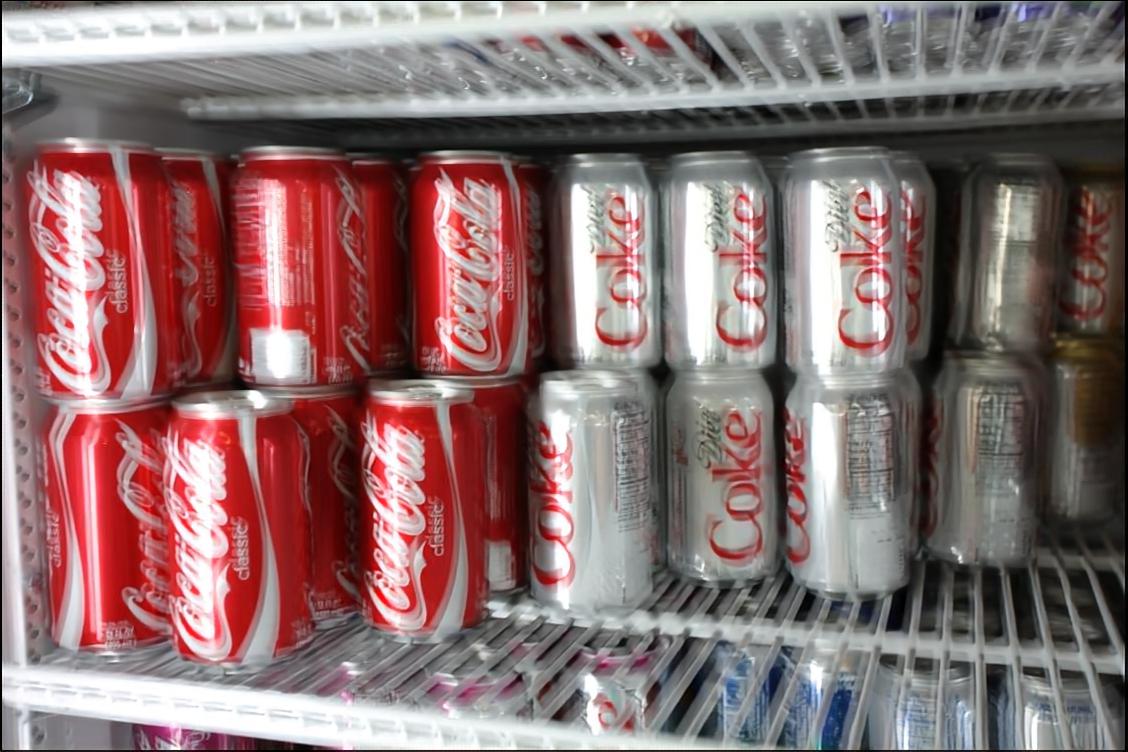} & 
\includegraphics[width=0.165\textwidth]{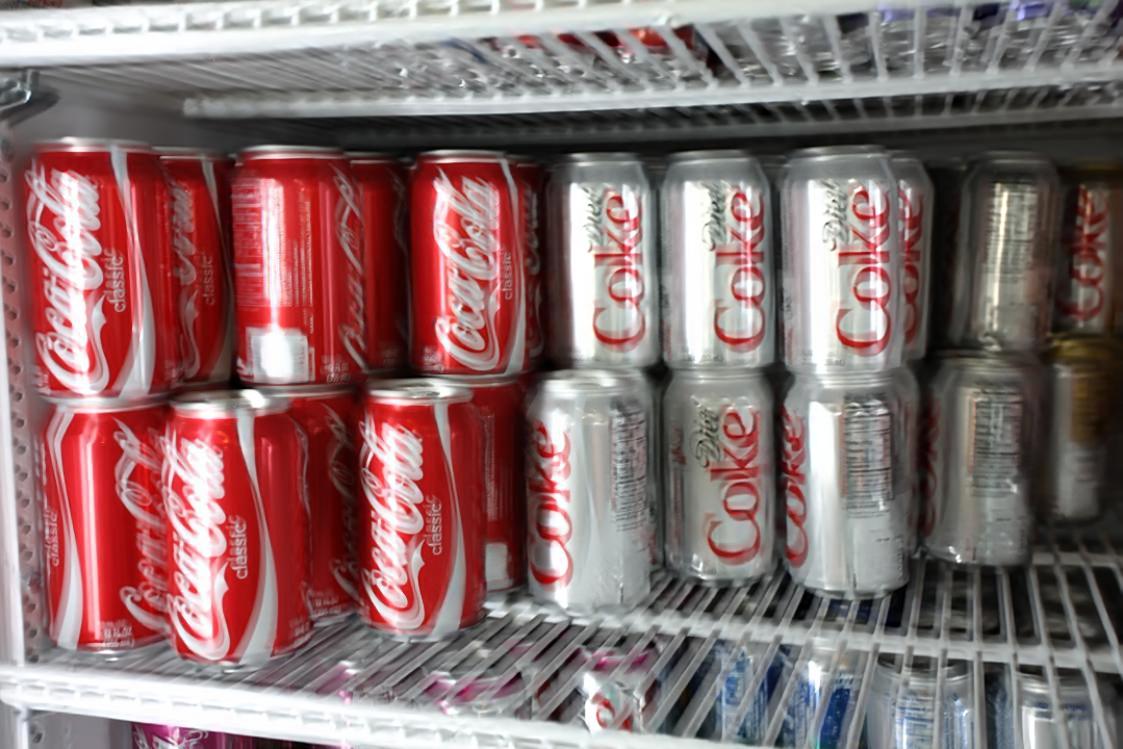}  &
\includegraphics[width=0.165\textwidth]{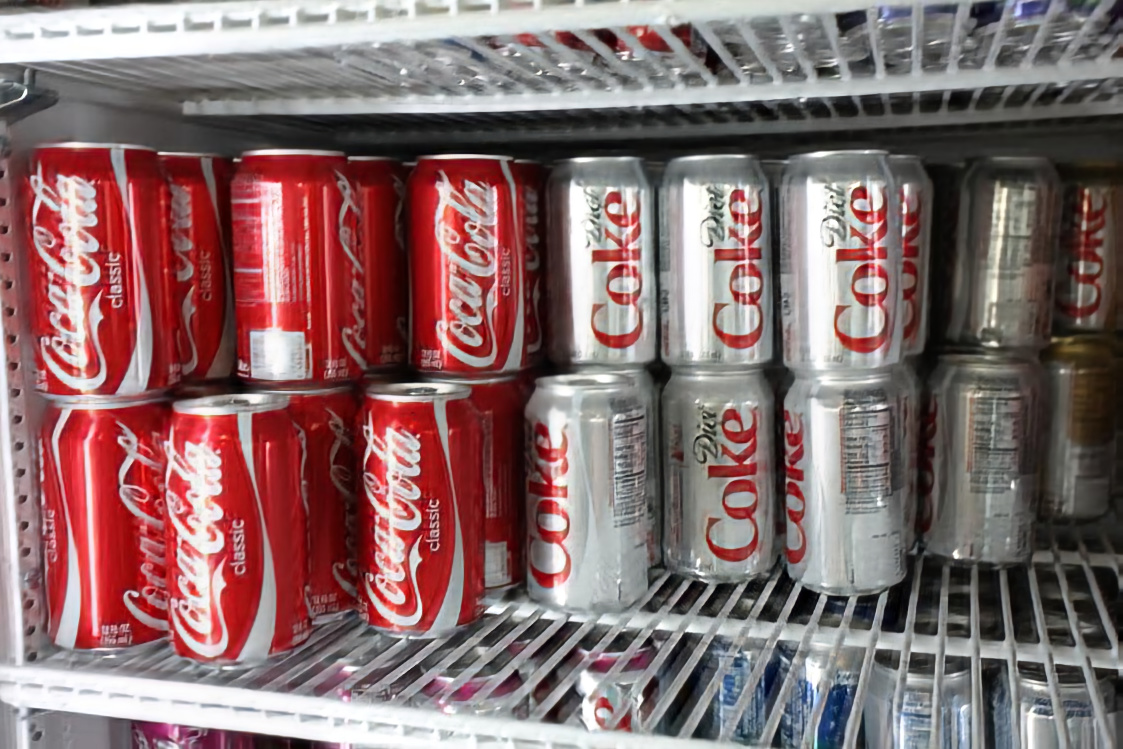}  &
\includegraphics[width=0.165\textwidth]{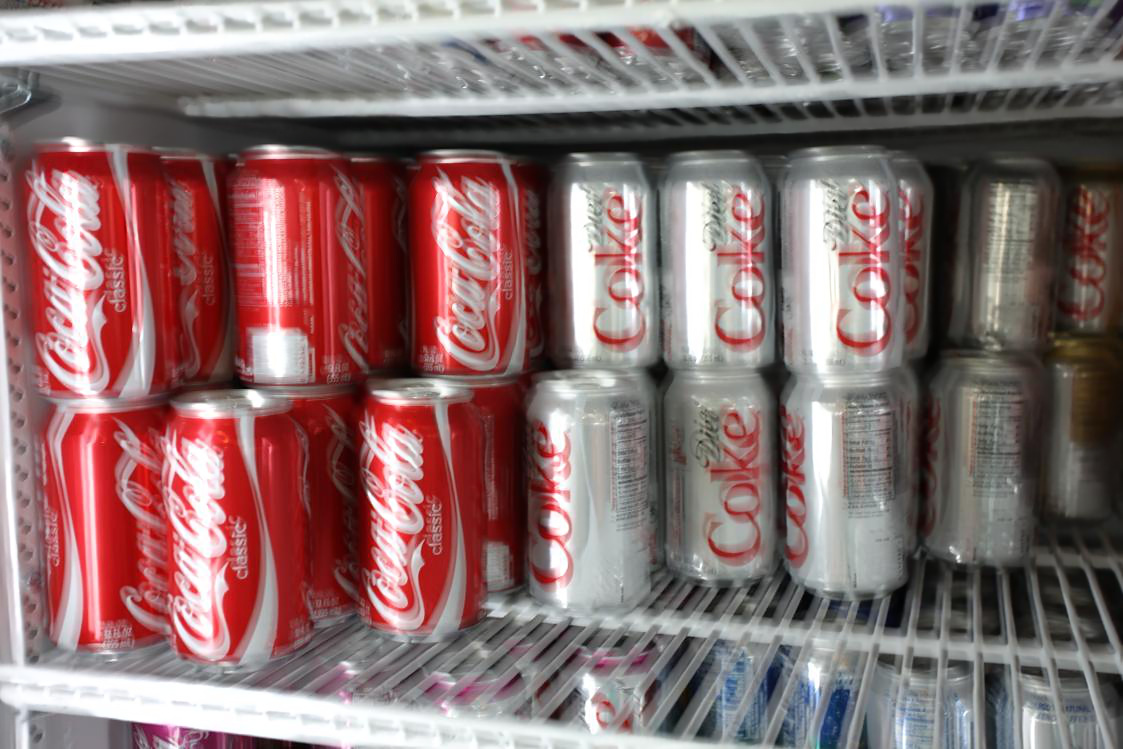}  &
\includegraphics[width=0.165\textwidth]{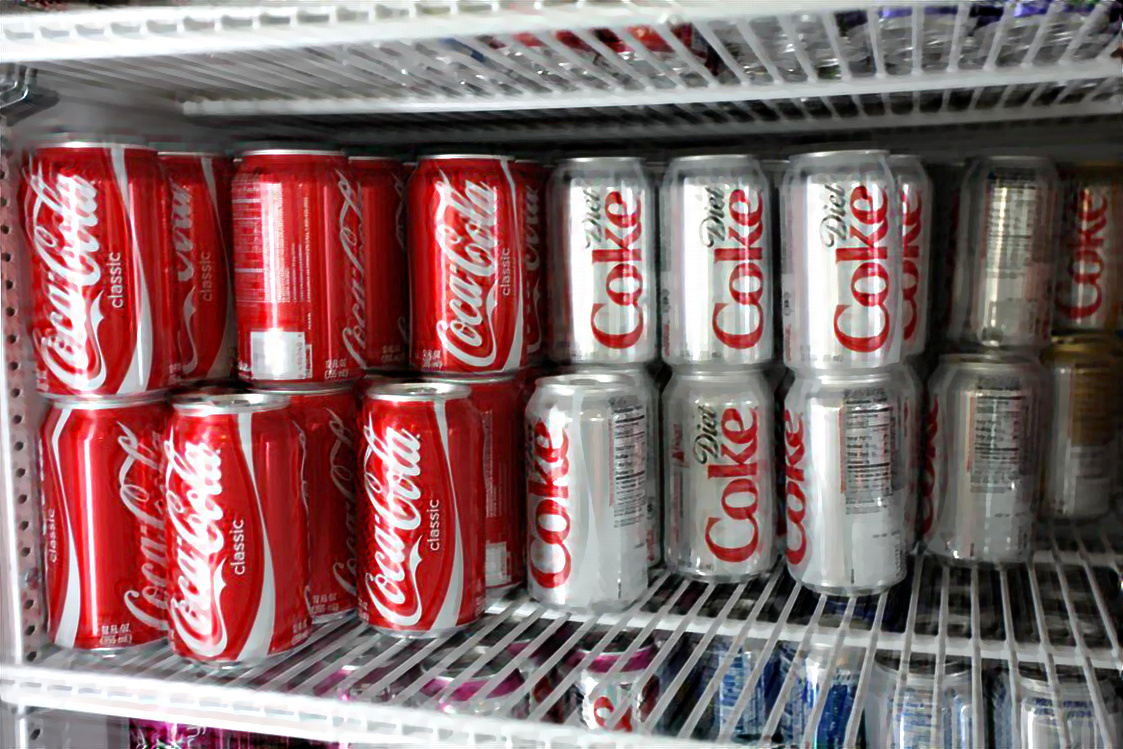}  \\
\includegraphics[trim=300 75 600 440, clip,width=0.08\textwidth]{latex/Lai/Blurry/coke.jpg}  
\includegraphics[trim=600 200 300 315, clip,width=0.08\textwidth]{latex/Lai/Blurry/coke.jpg}  &
\includegraphics[trim=300 75 600 440, clip,width=0.08\textwidth]{latex/Lai/DMPHN/coke.jpg}  
\includegraphics[trim=600 200 300 315, clip,width=0.08\textwidth]{latex/Lai/DMPHN/coke.jpg}  &
\includegraphics[trim=300 75 600 440, clip,width=0.08\textwidth]{latex/Lai/SRN/coke.jpg}  
\includegraphics[trim=600 200 300 315, clip,width=0.08\textwidth]{latex/Lai/SRN/coke.jpg}  &
\includegraphics[trim=300 75 600 440, clip,width=0.08\textwidth]{latex/Lai/RealBlur/coke.jpg}
\includegraphics[trim=600 200 300 315, clip,width=0.08\textwidth]{latex/Lai/RealBlur/coke.jpg} &
\includegraphics[trim=300 75 600 440, clip,width=0.08\textwidth]{latex/Lai/MPRNet/coke.jpg}
\includegraphics[trim=600 200 300 315, clip,width=0.08\textwidth]{latex/Lai/MPRNet/coke.jpg} &
\includegraphics[trim=300 75 600 440, clip,width=0.08\textwidth]{latex/Lai/ours_iccv/coke_restored.jpg}
\includegraphics[trim=600 200 300 315, clip,width=0.08\textwidth]{latex/Lai/ours_iccv/coke_restored.jpg}\\

% trim = left bottom right top
Blurred & DMPHN \cite{Zhang_2019_CVPR} &  SRN \cite{tao2018scale}&  RealBlur \cite{rim_2020_ECCV} & MPRNet \cite{Zamir2021MPRNet} & Ours \\
\end{tabular} 
    \caption{\textbf{Deblurring examples on real blurry images from Lai dataset  \cite{lai2016comparative}.} Our model-based approach is competitive with state-of-the-art data-driven methods. Best appreciated in electronic format.}
    \label{fig:LaiDeepLearning}
\end{figure*}

\subsection{Problem Formulation and Algorithm} 

The estimated per-pixel kernels form the \emph{forward model} of the blurring operation and can be used with any general linear inverse problem solver. Here we employ the classical Richardson-Lucy (RL) algorithm \cite{tai2011richardson-lucy}. In its basic form, RL recovers the latent sharp image as the \textit{maximum-likelihood} estimate under a \textit{Poisson noise} model. Following the notation in Section~\ref{sec:model}, the likelihood of the blurry image $\mathbf{v}$ as a function of the latent image $\mathbf{u}$ %under the Poisson noise model 
can be written as % the following per-pixel multiplication
\begin{equation}
    p\left( \mathbf{u} \vert \mathbf{v} \right) = \prod_i \frac{\hat{v_i}^{v_i} e^{ -\hat{v_i}}}{v_i!},
    \label{ec:poisson_likelihood_main}
\end{equation}
where
\begin{equation}
    \hat{v_i} =  \langle \mathbf{u}_{nn(i)}, \mathbf{k}_{i} \rangle  =  \langle \mathbf{u}_{nn(i)}, \sum_{b=1}^B\mathbf{k}^b m^b_{i} \rangle = \left(\mathbf{H}\mathbf{u}\right)_i,
\end{equation}
being $\mathbf{H}$ the \textit{degradation} matrix of size $N_v \times N_v$ for a $N_v$ pixels image. Each row in matrix $\mathbf{H}$ contains the corresponding per-pixel kernel.
Maximizing \eqref{ec:poisson_likelihood_main} subject to the non-negativity constraint on $\mathbf{u}$ gives rise to the RL update:
\begin{equation}
    \hat{\mathbf{u}}^{t+1} = \hat{\mathbf{u}}^t \circ \mathbf{H}^T \left( \frac{\mathbf{v}}{\mathbf{H}\hat{\mathbf{u}}^t} \right), %
    \label{ec:basicRL_main}
\end{equation}
where $\hat{\mathbf{u}}^t$ is the estimate at iteration $t$,  $\circ$ represents the Hadamard product and the quotient is element-wise. The complete derivation is included in Section~\ref{sec:app:rl}. \vspace{.3em}

\noindent {\bf Non-uniform Richardson-Lucy.}
To apply the RL update with the proposed non-uniform model, it is necessary to compute the matrix-vector products with matrices $\mathbf{H}$ and  $\mathbf{H^T}$ in \eqref{ec:basicRL_main}. For this, we benefit from the low-rank decomposition of the kernel field. Denoting $\mathbf{H} = \sum_b \mathbf{M}_b \mathbf{K}_b$, where $\mathbf{M}_b$ is a diagonal matrix containing the per-pixel mixing coefficient of the $b^{th}$ basis kernel and $\mathbf{K}_b$ is a Toeplitz matrix of the convolution with the basis element $\mathbf{k}_b$,
\begin{equation}
    \mathbf{Hx} = \sum_{b=1}^B \mathbf{M}_b \mathbf{K}_b\mathbf{x} \text{,~~~~} \mathbf{H^Tx} = \sum_{b=1}^B \mathbf{K}^T_b \mathbf{M}_b\mathbf{x}.
\label{eq:nonuniformRL1}
\end{equation}
%\begin{equation}
%    \mathbf{H^Tx} = \sum_{b=1}^B \mathbf{K}^T_b \mathbf{M}_b\mathbf{x}.
%\label{eq:nonuniformRL2}
%\end{equation}
%\begin{equation}
%    \hat{v}_{i} = \left(\mathbf {Hu}\right)_i =  \langle \mathbf{u}_{nn(i)}, \sum_{b=1}^B\mathbf{k}^b m^b_{i} \rangle
%        \label{eq:forward_RL}
%\end{equation}
% \begin{equation}
%     \left(\mathbf {H^Tx}\right)_i =  \langle 
%     \hat{\mathbf{x}}_{nn(i)}, \sum_{b=1}^B\mathbf{k_f}^b m^b_{i} \rangle
%         \label{eq:forward_RL}
% \end{equation}
Here we leverage the fact that the product of $\mathbf{K}_b$ or $\mathbf{K}^T_b$ with any vector $\mathbf{x}$ can be computed efficiently via the DFT.
%The computational complexity is thus reduced from $\mathcal{O}((HW)^2)$ to $\mathcal{O}(B\cdot HW \cdot log(HW))$. 
A similar procedure was  proposed recently in a concurrent work~\cite{Gwak2020}, in the context of modeling nonstationary lens blur.\vspace{.3em}

%
%\begin{equation}
%    \mathcal{L}_{TV\_prior} =- \lambda_{TV} \Vert \nabla \mathbf{u} \Vert 
%    \label{ec:TV_reg}
%\end{equation}
%and the update rule becomes
%\begin{equation}
%    \hat{\mathbf{u}}^{t+1} = \frac{\hat{\mathbf{u}}^t}{1 -  \lambda_{TV} div \left( \frac{ \nabla \hat{\mathbf{u}}^t}{\Vert \nabla \hat{\mathbf{u}}^t \Vert} \right)  } \circ \mathbf{H}^T \left( \frac{\mathbf{v}}{\mathbf{H}\hat{\mathbf{u}}^t} \right) 
%    \label{ec:regularizedRL}
%\end{equation}

% With every RL update, a new estimation $\hat{\mathbf{u}}$ of the latent image $\mathbf{u}$ is generated. Some of its pixels may take values greater than 1. This is handled by the \textit{forward model} with the inclusion of the saturation function \eqref{ec:sat_func}. 

% % The derivation of the RL algorithm is modified in  \ref{ec:RL_der1}

% % \begin{equation}
% % \frac{\partial \hat{v_i} }{\partial u_j} = H_{ij} R'\left( \left(\mathbf{Hu}\right)_i \right)
% % \end{equation}

% The update rule turns into

% \begin{equation}
%     \mathbf{u}^{t+1}= \mathbf{u}^{t} \circ \mathbf{H}^T \left( \frac{\mathbf{v} \circ R'\left( \mathbf{Hu}\right) \circ \mathbf{z} }{ \mathcal{R}\left( \mathbf{H}\mathbf{u}^{t}\right) } + 1 - \mathbf{z}   \right)
% \end{equation}

\noindent {\bf Richardson-Lucy in the Presence of  Saturated Pixels.}
%The presence of a saturated pixel in the latent sharp image may be an indicator of loss of information during the acquisition.  
The presence of saturated pixels in the blurry image may be an indicator of information loss during acquisition, and prevents the perfect recovery of the latent saturated pixels. Additionally, this generates a chain reaction that affects the restoration of nearby pixels. Underestimation of latent saturated pixels due to saturated blurry pixels is one of the main causes of ringing artifacts \cite{whyte2014deblurring}.    

To prevent the error propagation caused by the loss of information in blurry saturated pixels, Whyte \etal \cite{whyte2014deblurring} proposed to decouple the estimation of ``bright'' pixels in $\mathbf{u}$ from the latent pixels in reliable regions by separating them, in each RL update, into two different regions $\mathcal{S}$ and  $\mathcal{U}$, respectively.
%latent image pixels are separated into two regions: saturated $\mathcal{S}$ and unsaturated $\mathcal{U}$.
%Since it is not known before-hand which parts of $\mathbf{u}$ belong in $\mathcal{U}$ and which in $\mathcal{S}$, 
The segmentation is performed in each iteration using a threshold in the current latent image. In this work, we considered a threshold of 0.99. We applied a smoothing to prevent artifacts due to discontinuities between both regions.
The latent image in terms of those disjoint sets is written as: $\mathbf{u}=\mathbf{u_{\mathcal{U}}} + \mathbf{u_{\mathcal{S}}}$, where $\mathbf{u_{\mathcal{U}}}$ is the image where pixels in  $\mathcal{S}$ are $0$ and vice-versa.
% The forward model is
%\begin{equation}
%    \mathbf{v} = \mathbf{H} \mathbf{u}_{\mathcal{U}} + \mathbf{H} \mathbf{v}_{\mathcal{U}}.
%\end{equation}
%
Following \cite{whyte2012removing}, the update rule for unsaturated pixels is
\begin{equation}
    \hat{\mathbf{u}}_{\mathcal{U}}^{t+1}= \hat{\mathbf{u}}_{\mathcal{U}}^{t} \circ \mathbf{H}^T \left( \frac{\mathbf{v} \circ R'\left( \mathbf{H}\hat{\mathbf{u}}^{t} \right) \circ \mathbf{z} }{ R\left( \mathbf{H}\hat{\mathbf{u}}^t\right) } + \mathbf{1} - R'\left( \mathbf{H}\hat{\mathbf{u}}^t\right) \circ \mathbf{z}   \right),
    \label{ec:non_sat_pixels}
\end{equation}
and for saturated pixels,
\begin{equation}
    \hat{\mathbf{u}}_{\mathcal{S}}^{t+1}= \hat{\mathbf{u}}_{\mathcal{S}}^{t} \circ \mathbf{H}^T \left( \frac{\mathbf{v} \circ R'\left( \mathbf{H}\hat{\mathbf{u}}^t\right) }{ R\left( \mathbf{H}\hat{\mathbf{u}}^{t}\right) } + \mathbf{1} - R'\left( \mathbf{H}\hat{\mathbf{u}}^t\right)\right).  
    \label{ec:sat_pixels}
\end{equation}
Function $R'$ is the derivative of $R$ defined in \eqref{eq:sat_func}, while $\mathbf{z}$ is a binary mask whose values are 0 for pixels affected by the loss of information and 1 otherwise~\cite{whyte2012removing}.
%We build $\mathbf{z}$ by eroding  $\mathcal{U}$ with a disk of size 3.      
Intuitively, the main difference between \eqref{ec:non_sat_pixels} and  \eqref{ec:sat_pixels} is that to prevent the propagation of ringing, in the former only pixels with no loss of information are used to estimate the latent image while in the latter all the data is used.  $R'(x) \simeq 1$ for saturated pixels in the blurry images and therefore those pixels are not taken into account to recover the latent image. \\

\begin{figure*}
  \centering
\setlength{\tabcolsep}{2pt}

  \begin{tabular}{*{6}{c}}
    
    \includegraphics[width=0.15\textwidth]{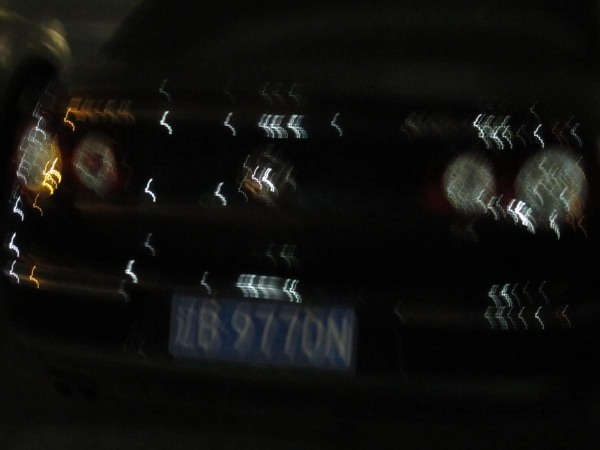}&
    \includegraphics[width=0.15\textwidth]{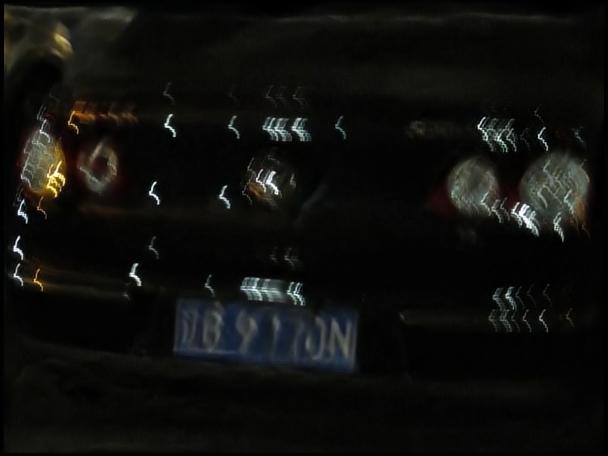}   &
        \includegraphics[width=0.15\textwidth]{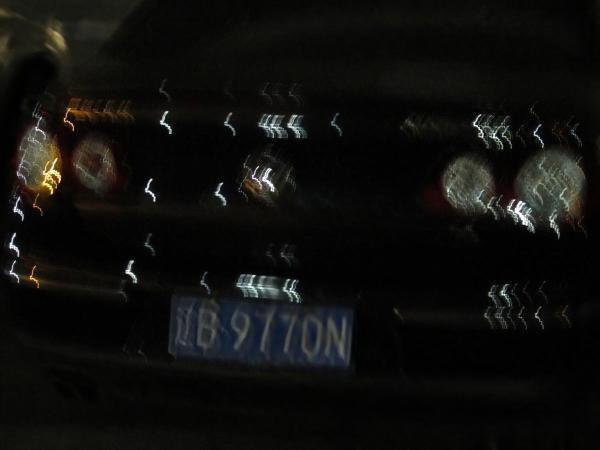} &
    \includegraphics[width=0.15\textwidth]{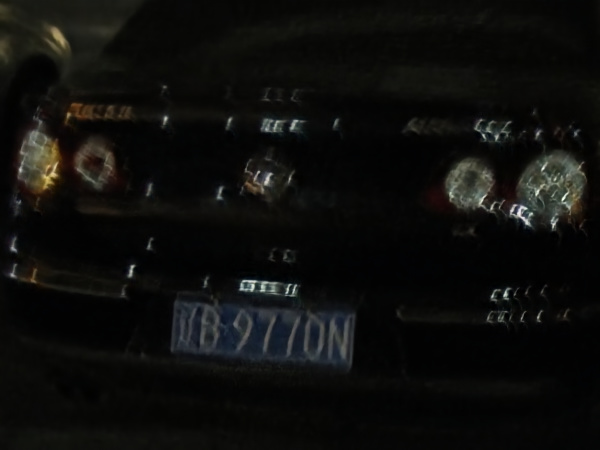}&
    \includegraphics[width=0.15\textwidth]{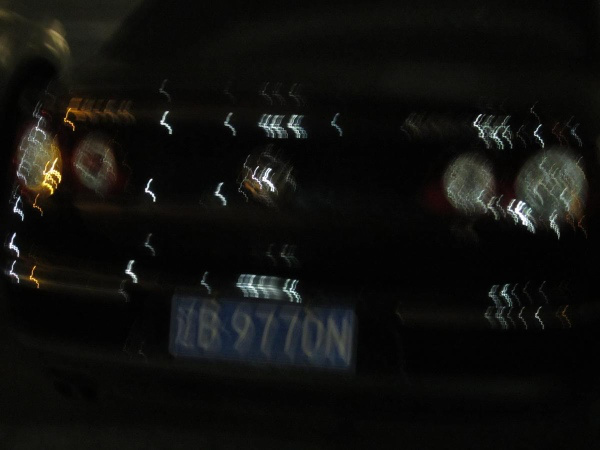}   &
    \includegraphics[width=0.15\textwidth]{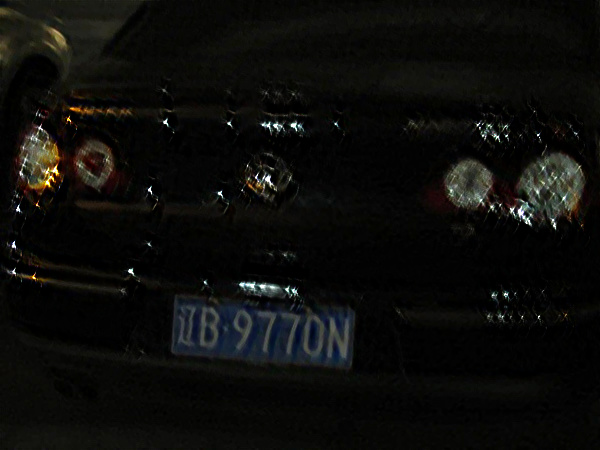}\\
    \includegraphics[width=0.15\textwidth]{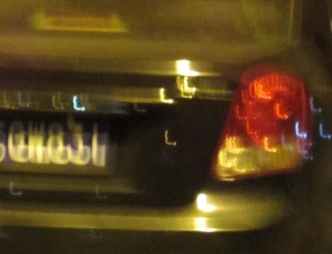}&
    \includegraphics[width=0.15\textwidth]{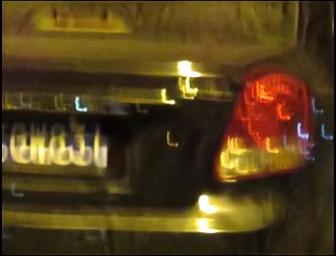}   &
        \includegraphics[width=0.15\textwidth]{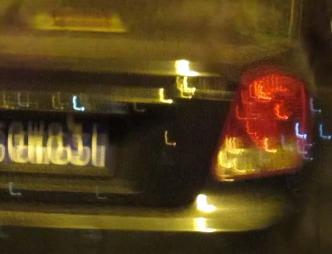} &
    \includegraphics[width=0.15\textwidth]{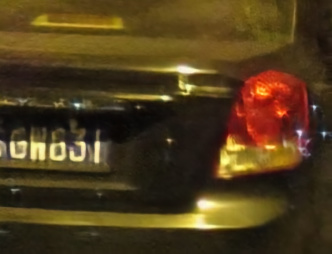}&
    \includegraphics[width=0.15\textwidth]{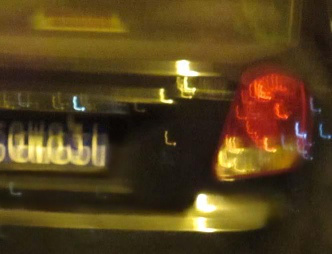}   &
    \includegraphics[width=0.15\textwidth]{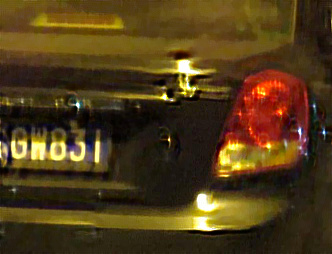}\\
    \includegraphics[width=0.15\textwidth]{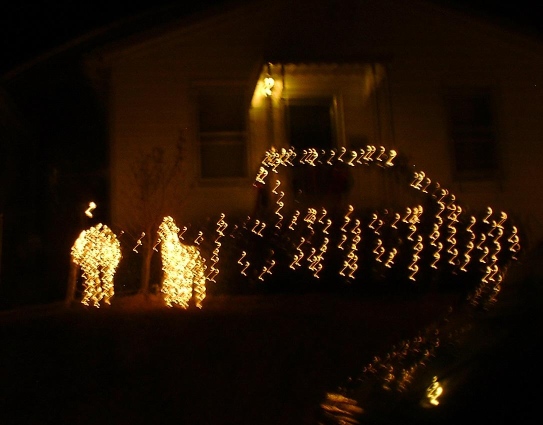}&
    \includegraphics[width=0.15\textwidth]{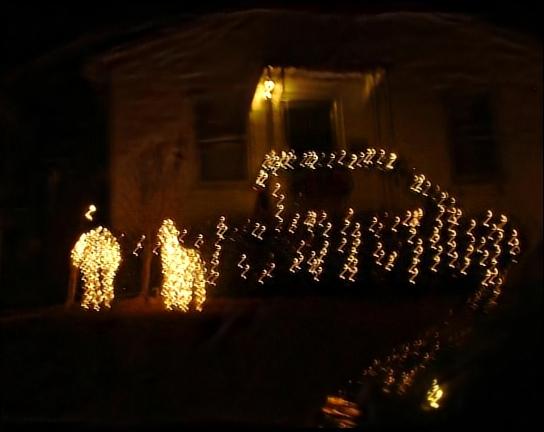}   &
        \includegraphics[width=0.15\textwidth]{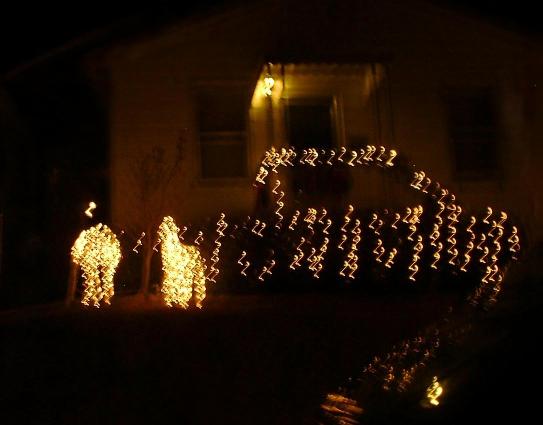} &
    \includegraphics[width=0.15\textwidth]{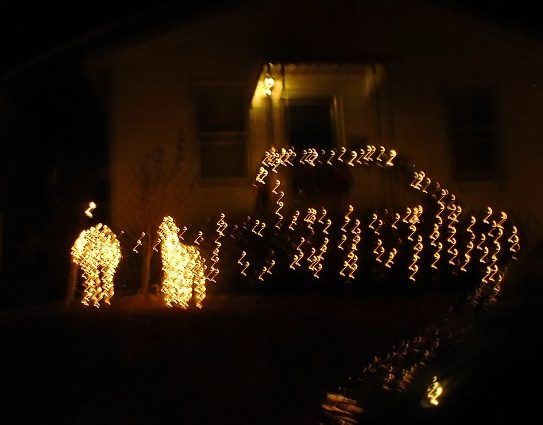}&
    \includegraphics[width=0.15\textwidth]{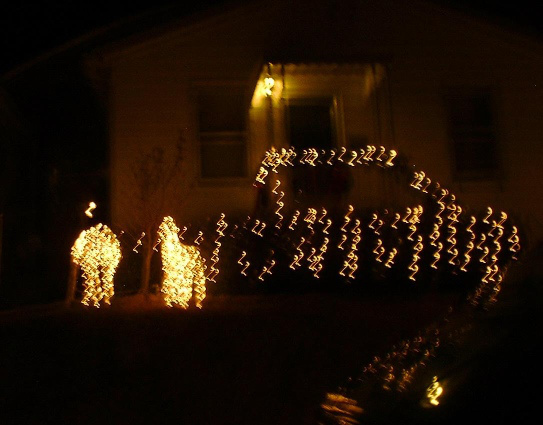}   &
    \includegraphics[width=0.15\textwidth]{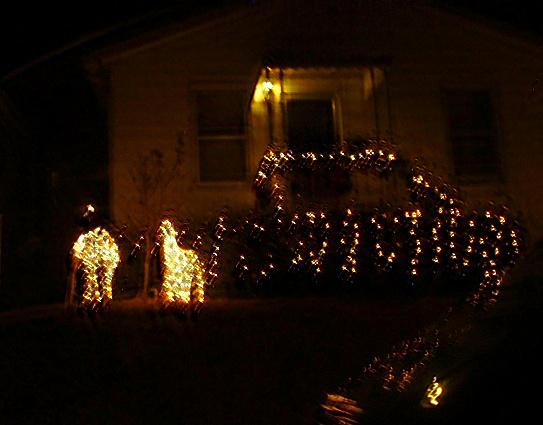}\\
Blurred & DMPHN \cite{Zhang_2019_CVPR} &  SRN \cite{tao2018scale}&  RealBlur \cite{rim_2020_ECCV} & MPRNet \cite{Zamir2021MPRNet} & Ours \\ 
  \end{tabular}
  \caption{\textbf{Deblurring examples on real images with saturated regions from Lai dataset \cite{lai2016comparative}.} While most of the methods struggle to handle saturated regions, our method is able to recover them. Corresponding kernel maps can be found in the Figure~\ref{fig:app:lai}. Best viewed in electronic format. }
  \label{fig:LaiSaturation}
\end{figure*}

\noindent {\bf Regularized Richardson-Lucy.} A natural image \textit{prior} can be easily added to the RL formulation \cite{dey20043d, tai2011richardson-lucy}. The \textit{Maximum a Posteriori} (MAP) solution implies modifying  \eqref{ec:basicRL_main} as in~\cite{dey20043d}:
% \begin{equation}
%     \hat{\mathbf{u}}^{t+1} = \frac{\hat{\mathbf{u}}^t}{1 + \nabla_{\mathbf{u}} \mathcal{L}_{prior}(\hat{\mathbf{u}}^t)} \circ \mathbf{H}^T \left( \frac{\mathbf{v}}{\mathbf{H}\hat{\mathbf{u}}^t} \right). 
%     \label{ec:regularizedRL}
% \end{equation}
%
\begin{align}
   & \hat{\mathbf{u}}_{unreg}^{t+1} = {\hat{\mathbf{u}}_{\mathcal{S}}^{t+1} + \hat{\mathbf{u}}_{\mathcal{U}}^{t+1}}  
    \label{ec:regularizedRL}   \\
    &    \hat{\mathbf{u}}^{t+1} = \frac{ \hat{\mathbf{u}}_{unreg}^{t+1}}{1 + \nabla_{\mathbf{u}} \mathcal{L}_{prior}(\hat{\mathbf{u}}^t)}.   
\end{align}

%where  $\nabla_{\mathbf{u}}\mathcal{L}_{prior}(\mathbf{u})$ is the gradient of $\mathcal{L}_{prior}(\mathbf{u})$ w.r.t. $\mathbf{u}$. 
Here we consider a Total Variation regularization $\mathcal{L}_{prior} = \lambda_{TV} \Vert \nabla \mathbf{u} \Vert$. \\ %therefore

The RL iteration is intialized with the blurry image, and ran until the reblur loss does not improve or until it reaches 30 iterations.
\subsection{Experimental Results in Real Images}

In this section we evaluate the ability of our deblurring procedure to generalize to real motion blurred photographs.  We compare the deblurring results on the standard datasets of real blurred images: Köhler \cite{kohler2012recording}, Lai \cite{lai2016comparative} and RealBlur \cite{rim_2020_ECCV}. \\

\begin{table}[h]
  \centering
  \begin{tabular}{lcc}
    \toprule
     &   \multicolumn{2}{c}{K\"ohler} \\
     Method &    PSNR & SSIM \\
    \midrule
    DeblurGAN \cite{kupyn2018deblurgan}          &     26.05 & 0.75          \\
    GoPro K=2 \cite{Nah_2017_CVPR}      &     (26.02) & (\underline{0.81})           \\
    SRN \cite{tao2018scale}             &     27.18 & 0.79          \\
    DMPHN 1-2-4 \cite{Zhang_2019_CVPR}     &     25.69&0.75        \\
    DeblurGANv2 Inc. \cite{kupyn2019deblurgan}    &      27.25&0.79       \\
    DeblurGANv2 Mob. \cite{kupyn2019deblurgan}  &       25.88&0.74       \\
    %\hline
    SRN$^1$\cite{tao2018scale}        &       (26.57)&(0.80)           \\
    SRN$^2$ \cite{tao2018scale}   &  
    \underline{27.85}& \underline{0.81}     \\
    MPRNet  \cite{Zamir2021MPRNet}   &  
    26.57 & 0.78    \\
    \hline                                                                    
    Sun \textit{et. al } \cite{sun2015learning}   &      (25.22)&(0.77)$^1$         \\
    Gong \cite{gong2017motion}         &       25.23&0.74      \\
    \hline                                                                    
    Ours         &      \textbf{28.39}&\textbf{0.82}          \\
    \bottomrule
  \end{tabular}
  \caption{\textbf{Quantitative comparison for image deblurring (PSNR/SSIM) in a real dataset.}  When possible, we reproduced the results using  available code, otherwise parenthesis are used and values are taken from their own paper.  $^1$ is trained with RealBlurJ and $^2$ with GoPro, BSD, RealBlurJ as in \cite{rim_2020_ECCV}.  }
    \label{tab:quantitative_deblurring_results}
\end{table}

\noindent {\bf Köhler Dataset \cite{kohler2012recording}.} Quantitative and qualitative results are presented in Table~\ref{tab:quantitative_deblurring_results} and Figure~\ref{fig:kohlerDataset}. Our method compares favorably to state-of-the-art end-to-end DL methods (gain between 0,54 and 2,51 dB), that fail to generalize from the synthetic dataset they were trained on to real blurred images  
(gain of 3.17 and 3.16 dB). \\

\noindent {\bf Lai Dataset \cite{lai2016comparative}} is a standard benchmark of real blurry images with no corresponding ground truth, allowing only visual comparison, which we show in Figure~\ref{fig:LaiDeepLearning}, \ref{fig:LaiSaturation} and in the Appendix. Note that our model-based method is competitive with state-of-the-art end-to-end approaches, outperforming most of them. Figure \ref{fig:LaiSaturation} shows the capacity of our model  to deal with real scenes with saturated regions in low-light conditions. Note that most of the compared methods fall short in this case. \\

\begin{table}[h]
    \centering
    \begin{tabular}{lcccc}
    \toprule
    &  \multicolumn{2}{c}{ RealBlur-R }  &  \multicolumn{2}{c}{ RealBlur-J }  \\
    Method & PSNR & SSIM& PSNR & SSIM \\
    \midrule
    Hu et al. \cite{hu2014deblurring}  &   33.67 & 0.916 & 26.41 & 0.803 \\
           Pan et al. \cite{pan2016blind}  &   34.01 & 0.916 & 27.22 & 0.790 \\
        Xu et al. \cite{xu2013unnatural}  &   34.46 & 0.937 & 27.14 & 0.830 \\
        \hline 
     Nah et al. \cite{Nah_2017_CVPR} &   32.51 & 0.841 & 27.87 & 0.827 \\
      DeblurGAN \cite{kupyn2018deblurgan}  &   33.79 & 0.903 & 27.97 & 0.834 \\
        DeblurGAN-v2 \cite{kupyn2019deblurgan} & 35.26 & 0.944 & 28.70 & 0.866 \\
Zhang et al. \cite{zhang2018dynamic} & 35.48 & 0.947 & 27.80 & 0.847 \\
SRN \cite{tao2018scale} & 35.66 & 0.947 & 28.56 & 0.867\\
DMPHN \cite{Zhang_2019_CVPR} & 35.70 & 0.948 & 28.42 & 0.860\\
MPRNet \cite{Zamir2021MPRNet}  & 35.99 & \textbf{0.952} &28.70 & \textbf{0.873} \\
Ours & \textbf{36.17} & 0.946 &\textbf{28.95} & 0.865\\
%\midrule
%\hline
%DeblurGAN-v2 \cite{kupyn2019deblurgan}  & 36.44 & 0.935& 29.69 &0.870 %\\
%SRN \cite{tao2018scale} & 38.65 & 0.965 & 31.38 & 0.909 \\
%MPRNet  \cite{Zamir2021MPRNet} & 39.31 & 0.972 & 31.76 & 0.922 \\
%Ours & 36.55 & 0.949 & 29.07 & 0.865 \\
 %   \hline

    \bottomrule
    \end{tabular}
    \caption{\textbf{Cross-dataset quantitative evaluation.} The first three methods are classical variational methods. DL-based methods were trained on synthetic datasets and evaluated on real blurry images. The PSNR/SSIM scores for other methods are taken from the RealBlur benchmark \cite{rim_2020_ECCV,Zamir2021MPRNet}.
    }
    \label{tab:RealBlurDataset}
\end{table}

\noindent {\bf RealBlur Dataset~\cite{rim_2020_ECCV}. } We evaluate our method following the  benchmark presented in \cite{rim_2020_ECCV}. 
%The deblurred and sharp ground truth images are aligned using a homography estimated by the enhanced correlation coefficients method~\cite{evangelidis2008parametric}. 
The comparison to state-of-the-art end-to-end DL-based and classical variational methods is shown in Table~\ref{tab:RealBlurDataset}. 
All the learning-based compared methods were trained in the GoPro dataset \cite{Nah_2017_CVPR} and ours in our dataset presented in Section~\ref{subsec:DatasetGen}. All methods are evaluated in RealBlur \cite{rim_2020_ECCV}. This cross-dataset evaluation allows to  validate the generalization of the deblurring performance and to avoid evaluating dataset-specific mappings. Our method outperforms previous methods in terms of PSNR. We conjecture that this is due to the heavy overfitting of end-to-end DL methods to specifics of the training databases. 
Qualitative comparisons on representative examples can be found in Section~\ref{sec:app:qualitative}.

\subsection{Blurring to Deblur}
Chen \etal \cite{chen2018reblur2deblur} proposed to impose cycle-consistency to deblurring models using a forward model estimated from consecutive frames in a video. The motivation was to prevent a deblurring conditional GAN \cite{kupyn2018deblurgan} from hallucinating image content. In the same spirit, and to validate our estimated kernels, we fine-tuned a pre-trained DeblurGAN~\cite{kupyn2018deblurgan} network, using our estimated kernels for imposing the forward-model consistency. Results shown in Table~\ref{tab:Reblurresults} prove that our fine-tuning is useful to improve a DeblurGAN model, slightly outperforming~\cite{chen2018reblur2deblur}. Also we note that different to \cite{chen2018reblur2deblur}, our method only requires a single frame instead of a video.

\begin{table}[h]
    \centering
    \begin{tabular}{lcc}
    \toprule
     &  \multicolumn{2}{c}{ Gopro }  \\
     Method & PSNR & SSIM   \\ 
    \midrule
    DeblurGAN &   27.25 & 0.81 \\
    \midrule
    DeblurGAN (resume training) &  27.57 & 0.83   \\
    DeblurGAN + Reblur2Deblur\cite{chen2018reblur2deblur} &  28.03$^1$   &  \textbf{0.90}$^1$ \\
    DeblurGAN + ours & \textbf{28.08} &  0.85\\
    \bottomrule
    \end{tabular}
    \caption{\textbf{Blurring to deblur.} Comparison between Reblur2Deblur \cite{chen2018reblur2deblur} and the proposed method in GoPro \cite{Nah_2017_CVPR}. The incorporation of a reblur loss in training produces better results than just resuming training. $^1$ No code available. 
    }
    \label{tab:Reblurresults}
\end{table}

\section{Conclusions}
%In this work we address the problem of non-uniform motion blur kernel estimation. 
We introduced a novel method that predicts a dense field of motion blur kernels, via an efficient image-dependent decomposition. This results in a compact, non-parametric definition of the non-uniform motion field. Extensive experimental results validate the estimated kernels and show that when combined with a non-blind variational deblurring algorithm, the method's generalization to real blurry images is superior to that of state-of-the-art end-to-end deep learning methods. This work contributes to bridging a longstanding gap between model-based and data-driven methods for motion blur removal.

\section*{Acknowledgements}
GC was supported partially by Agencia Nacional de Investigación e Innovación (ANII, Uruguay) grant POS\_FCE\_2018\_1\_1007783  and PV by the  MICINN/FEDER UE project under Grant PGC2018-098625-B-I0; H2020-MSCA-RISE-2017 under Grant 777826 NoMADS and  Spanish Ministry of Economy and Competitiveness under the Maria de Maetzu Units of Excellence Programme (MDM-2015-0502).
The experiments presented in this paper were carried out using ClusterUY (site: https://cluster.uy) and GPUs donated by NVIDIA Corporation.
We also thank Juan F. Montesinos for his help during the experimental phase.

{\small
\bibliographystyle{ieee_fullname}
\bibliography{main}

\begin{thebibliography}{10}\itemsep=-1pt

\bibitem{chen2018reblur2deblur}
H. Chen, J. Gu, O. Gallo, M. Liu, A. Veeraraghavan, and J. Kautz.
\newblock Reblur2deblur: Deblurring videos via self-supervised learning.
\newblock In {\em 2018 IEEE International Conference on Computational
  Photography (ICCP)}, pages 1--9, Los Alamitos, CA, USA, may 2018. IEEE
  Computer Society.

\bibitem{delbracio2015removing}
Mauricio Delbracio and Guillermo Sapiro.
\newblock Removing camera shake via weighted fourier burst accumulation.
\newblock {\em IEEE Trans. Pattern Anal. Mach. Intell.}, 12(1):234--778, 2002.

\bibitem{dey20043d}
Nicolas Dey, Laure Blanc-F{\'e}raud, Christophe Zimmer, Pascal Roux, Zvi Kam,
  Jean-Christophe Olivo-Marin, and Josiane Zerubia.
\newblock {\em 3D microscopy deconvolution using Richardson-Lucy algorithm with
  total variation regularization}.
\newblock PhD thesis, INRIA, 2004.

\bibitem{gavant2011physiological}
Fabien Gavant, Laurent Alacoque, Antoine Dupret, and Dominique David.
\newblock A physiological camera shake model for image stabilization systems.
\newblock In {\em SENSORS, 2011 IEEE}, pages 1461--1464, 2014.

\bibitem{gong2017motion}
Dong Gong, Jie Yang, Lingqiao Liu, Yanning Zhang, Ian Reid, Chunhua Shen, Anton
  Van Den~Hengel, and Qinfeng Shi.
\newblock From motion blur to motion flow: a deep learning solution for
  removing heterogeneous motion blur.
\newblock In {\em Proceedings of the IEEE Conference on Computer Vision and
  Pattern Recognition}, pages 2319--2328, 2017.

\bibitem{gupta2010singleimage}
Ankit Gupta, Neel Joshi, C.~Lawrence Zitnick, Michael Cohen, and Brian Curless.
\newblock Single image deblurring using motion density functions.
\newblock In {\em Proceedings of the 11th European Conference on Computer
  Vision: Part I}, ECCV'10, page 171–184, Berlin, Heidelberg, 2010.
  Springer-Verlag.

\bibitem{Gwak2020}
Moonsung Gwak and Seungjoon Yang.
\newblock Modeling nonstationary lens blur using eigen blur kernels for
  restoration.
\newblock {\em Opt. Express}, 28(26):39501--39523, Dec 2020.

\bibitem{hirsch2011fastremoval}
M. {Hirsch}, C.~J. {Schuler}, S. {Harmeling}, and B. {Schölkopf}.
\newblock Fast removal of non-uniform camera shake.
\newblock In {\em 2011 International Conference on Computer Vision}, pages
  463--470, 2011.

\bibitem{hu2014deblurring}
Zhe Hu, Sunghyun Cho, Jue Wang, and Ming-Hsuan Yang.
\newblock Deblurring low-light images with light streaks.
\newblock In {\em Proceedings of the IEEE Conference on Computer Vision and
  Pattern Recognition}, pages 3382--3389, 2014.

\bibitem{jia2016dynamic}
Xu Jia, Bert De~Brabandere, Tinne Tuytelaars, and Luc~V Gool.
\newblock Dynamic filter networks.
\newblock In {\em Advances in neural information processing systems}, pages
  667--675, 2016.

\bibitem{kim2013dynamic}
T.~H. {Kim}, B. {Ahn}, and K.~M. {Lee}.
\newblock Dynamic scene deblurring.
\newblock In {\em 2013 IEEE International Conference on Computer Vision}, pages
  3160--3167, 2013.

\bibitem{kim2014segmentation-free}
T.~H. {Kim} and K.~M. {Lee}.
\newblock Segmentation-free dynamic scene deblurring.
\newblock In {\em 2014 IEEE Conference on Computer Vision and Pattern
  Recognition}, pages 2766--2773, 2014.

\bibitem{kohler2012recording}
Rolf K{\"o}hler, Michael Hirsch, Betty Mohler, Bernhard Sch{\"o}lkopf, and
  Stefan Harmeling.
\newblock Recording and playback of camera shake: Benchmarking blind
  deconvolution with a real-world database.
\newblock In {\em European conference on computer vision}, pages 27--40.
  Springer, 2012.

\bibitem{kupyn2018deblurgan}
Orest Kupyn, Volodymyr Budzan, Mykola Mykhailych, Dmytro Mishkin, and
  Ji{\v{r}}{\'\i} Matas.
\newblock Deblurgan: Blind motion deblurring using conditional adversarial
  networks.
\newblock In {\em Proceedings of the IEEE conference on computer vision and
  pattern recognition}, pages 8183--8192, 2018.

\bibitem{kupyn2019deblurgan}
Orest Kupyn, Tetiana Martyniuk, Junru Wu, and Zhangyang Wang.
\newblock Deblurgan-v2: Deblurring (orders-of-magnitude) faster and better.
\newblock In {\em Proceedings of the IEEE International Conference on Computer
  Vision}, pages 8878--8887, 2019.

\bibitem{lai2016comparative}
Wei-Sheng Lai, Jia-Bin Huang, Zhe Hu, Narendra Ahuja, and Ming-Hsuan Yang.
\newblock A comparative study for single image blind deblurring.
\newblock In {\em Proceedings of the IEEE Conference on Computer Vision and
  Pattern Recog.}, pages 1701--1709, 2016.

\bibitem{Lucy1974}
B.~L. Lucy.
\newblock An iterative technique for the rectification of observed
  distributions.
\newblock {\em Astronomical Journal}, 1974.

\bibitem{mildenhall2018burst}
Ben Mildenhall, Jonathan~T Barron, Jiawen Chen, Dillon Sharlet, Ren Ng, and
  Robert Carroll.
\newblock Burst denoising with kernel prediction networks.
\newblock In {\em Proceedings of the IEEE Conference on Computer Vision and
  Pattern Recognition}, pages 2502--2510, 2018.

\bibitem{Nah_2019_CVPR_Workshops_REDS}
Seungjun Nah, Sungyong Baik, Seokil Hong, Gyeongsik Moon, Sanghyun Son, Radu
  Timofte, and Kyoung~Mu Lee.
\newblock Ntire 2019 challenge on video deblurring and super-resolution:
  Dataset and study.
\newblock In {\em The IEEE Conference on Computer Vision and Pattern
  Recognition (CVPR) Workshops}, June 2019.

\bibitem{Nah_2017_CVPR}
Seungjun Nah, Tae~Hyun Kim, and Kyoung~Mu Lee.
\newblock Deep multi-scale convolutional neural network for dynamic scene
  deblurring.
\newblock In {\em The IEEE Conference on Computer Vision and Pattern
  Recognition (CVPR)}, July 2017.

\bibitem{niklaus2017video}
Simon Niklaus, Long Mai, and Feng Liu.
\newblock Video frame interpolation via adaptive convolution.
\newblock In {\em Proceedings of the IEEE Conference on Computer Vision and
  Pattern Recognition}, pages 670--679, 2017.

\bibitem{niklaus2017video2}
Simon Niklaus, Long Mai, and Feng Liu.
\newblock Video frame interpolation via adaptive separable convolution.
\newblock In {\em Proceedings of the IEEE International Conference on Computer
  Vision}, pages 261--270, 2017.

\bibitem{pan2016soft}
Jin{-}shan Pan, Zhe Hu, Zhixun Su, Hsin{-}Ying Lee, and Ming{-}Hsuan Yang.
\newblock Soft-segmentation guided object motion deblurring.
\newblock In {\em 2016 {IEEE} Conference on Computer Vision and Pattern
  Recognition, {CVPR} 2016, Las Vegas, NV, USA, June 27-30, 2016}, pages
  459--468. {IEEE} Computer Society, 2016.

\bibitem{pan2016blind}
Jinshan Pan, Deqing Sun, Hanspeter Pfister, and Ming-Hsuan Yang.
\newblock Blind image deblurring using dark channel prior.
\newblock In {\em Proceedings of the IEEE Conference on Computer Vision and
  Pattern Recognition}, pages 1628--1636, 2016.

\bibitem{Richardson1972}
William~Hadley Richardson.
\newblock Bayesian-based iterative method of image restoration$\ast$.
\newblock {\em J. Opt. Soc. Am.}, 62(1):55--59, Jan 1972.

\bibitem{rim_2020_ECCV}
Jaesung Rim, Haeyun Lee, Jucheol Won, and Sunghyun Cho.
\newblock Real-world blur dataset for learning and benchmarking deblurring
  algorithms.
\newblock In {\em Proceedings of the European Conference on Computer Vision
  (ECCV)}, 2020.

\bibitem{dai2008motion}
{Shengyang Dai} and {Ying Wu}.
\newblock Motion from blur.
\newblock In {\em 2008 IEEE Conference on Computer Vision and Pattern
  Recognition}, pages 1--8, 2008.

\bibitem{su2017deep}
Shuochen Su, Mauricio Delbracio, Jue Wang, Guillermo Sapiro, Wolfgang Heidrich,
  and Oliver Wang.
\newblock Deep video deblurring for hand-held cameras.
\newblock In {\em Proceedings of the IEEE Conference on Computer Vision and
  Pattern Recognition}, pages 1279--1288, 2017.

\bibitem{sun2015learning}
Jian Sun, Wenfei Cao, Zongben Xu, and Jean Ponce.
\newblock Learning a convolutional neural network for non-uniform motion blur
  removal.
\newblock In {\em Proceedings of the IEEE Conference on Comput.er Vis.}, 2015.

\bibitem{tai2011richardson-lucy}
Y. {Tai}, P. {Tan}, and M.~S. {Brown}.
\newblock Richardson-lucy deblurring for scenes under a projective motion path.
\newblock {\em IEEE Transactions on Pattern Analysis and Machine Intelligence},
  33(8):1603--1618, 2011.

\bibitem{tao2018scale}
Xin Tao, Hongyun Gao, Xiaoyong Shen, Jue Wang, and Jiaya Jia.
\newblock Scale-recurrent network for deep image deblurring.
\newblock In {\em Proceedings of the IEEE Conference on Comput.er Vis.ion and
  Pattern Recog. Worksh.}, 2018.

\bibitem{whyte2012removing}
Oliver Whyte.
\newblock {\em Removing camera shake blur and unwanted occluders from
  photographs}.
\newblock PhD thesis, {\'E}cole normale sup{\'e}rieure de Cachan-ENS Cachan,
  2012.

\bibitem{whyte2014deblurring}
Oliver Whyte, Josef Sivic, and Andrew Zisserman.
\newblock Deblurring shaken and partially saturated images.
\newblock {\em International journal of computer vision}, 110(2):185--201,
  2014.

\bibitem{whyte2010nonuniform}
O. {Whyte}, J. {Sivic}, A. {Zisserman}, and J. {Ponce}.
\newblock c.
\newblock In {\em 2010 IEEE Computer Society Conference on Computer Vision and
  Pattern Recognition}, pages 491--498, 2010.

\bibitem{xia2019basis}
Zhihao Xia, Federico Perazzi, Micha{\"e}l Gharbi, Kalyan Sunkavalli, and Ayan
  Chakrabarti.
\newblock Basis prediction networks for effective burst denoising with large
  kernels.
\newblock {\em arXiv preprint arXiv:1912.04421}, 2019.

\bibitem{xu2013unnatural}
Li Xu, Shicheng Zheng, and Jiaya Jia.
\newblock Unnatural l0 sparse representation for natural image deblurring.
\newblock In {\em Proceedings of the IEEE conference on computer vision and
  pattern recognition}, pages 1107--1114, 2013.

\bibitem{Zamir2021MPRNet}
Syed~Waqas Zamir, Aditya Arora, Salman Khan, Munawar Hayat, Fahad~Shahbaz Khan,
  Ming-Hsuan Yang, and Ling Shao.
\newblock Multi-stage progressive image restoration.
\newblock In {\em CVPR (to appear)}, 2021.

\bibitem{Zhang_2019_CVPR}
Hongguang Zhang, Yuchao Dai, Hongdong Li, and Piotr Koniusz.
\newblock Deep stacked hierarchical multi-patch network for image deblurring.
\newblock In {\em The IEEE Conference on Computer Vision and Pattern
  Recognition (CVPR)}, June 200319.

\bibitem{zhang2018dynamic}
Jiawei Zhang, Jinshan Pan, Jimmy Ren, Yibing Song, Linchao Bao, Rynson~WH Lau,
  and Ming-Hsuan Yang.
\newblock Dynamic scene deblurring using spatially variant recurrent neural
  networks.
\newblock In {\em Proceedings of the IEEE Conference on Computer Vision and
  Pattern Recognition}, pages 2521--2529, 2018.

\bibitem{zhou2017scene}
Bolei Zhou, Hang Zhao, Xavier Puig, Sanja Fidler, Adela Barriuso, and Antonio
  Torralba.
\newblock Scene parsing through ade20k dataset.
\newblock In {\em Proceedings of the IEEE Conference on Computer Vision and
  Pattern Recognition}, 2017.

\bibitem{zoran2011learning}
Daniel Zoran and Yair Weiss.
\newblock From learning models of natural image patches to whole image
  restoration.
\newblock In {\em 2011 International Conference on Computer Vision}, pages
  479--486. IEEE, 2011.

\end{thebibliography}
}

\clearpage
%%%%%%%%%%%%%%%%%%%%%
\appendix
\section{Appendix}

\subsection{Architecture Details}

Our network is built upon a Kernel Prediction Network (KPN) proposed by~\cite{xia2019basis}. The network architecture is shown in Figure 3 of the submission. It is composed of one \textit{contractive path} and two expansive paths that yield the kernel basis and mixing coefficients. We call them \textit{kernel head} and \textit{mixing coefficients head}, respectively.

The \textit{contractive path} is composed of five \textit{down-sampling} blocks. Each \textit{down-sampling} block is composed of two convolutions followed by a max-pooling layer. After each \textit{down-sampling} block the spatial size is divided by two. The output is a feature vector, that is fed into both the \textit{kernel head} and \textit{mixing coefficients head}. 

The \textit{mixing coefficients head} together with the \textit{contractive path} follow a U-Net architecture with skip connections. After the last convolution, softmax is applied along the $B$-channels dimension to ensure that the sum of the per-pixel mixing coefficients associated with the kernels adds up to one. 

The \textit{kernel head} aims to produce $B$ basis kernels, each of size $K \times K$. In our work $B=25$ and $K=33$.  The first operation performed  is a  Global Average Pooling which takes into account that, unlike the \textit{mixing coefficients head}, there is no correspondence between the spatial positions of input and output pixels. Then, the feature vector goes through five \textit{up-sampling} blocks. Each \textit{up-sampling} is composed of a bilinear upsampling followed by three convolutional layers. Additionally, there are skip connections between the second convolution in the \textit{down-sampling} blocks and the second convolution in the \textit{up-sampling} blocks. Finally, two 64-channels convolutional layers followed by a $B$-channels convolutional layer with softmax are applied to ensure the $B$ kernels of size $K \times K$ are positive and add up to one. Convolutional kernel sizes are 3 with 1-padding except for the first 64-channel convolutions, whose kernel sizes are 2.  

Figure~\ref{fig:KernelsAndMasks} shows additional examples of kernels and corresponding mixing coefficients generated with the proposed network. The non-uniform motion blur represented by the set of basis kernels and mixing coefficients allows to re-blur the corresponding sharp image by first
convolving it with each basis kernel, and then performing a weighted sum of the results using the mixing coefficients. 

\begin{figure*}[ht!]
\setlength\tabcolsep{1.0pt} % default value: 6pt
\centering
 \begin{tabular}{c@{\hspace{1em}}c}
 \multirow{2}{*}[0.26cm]{\includegraphics[width=0.07\textwidth]{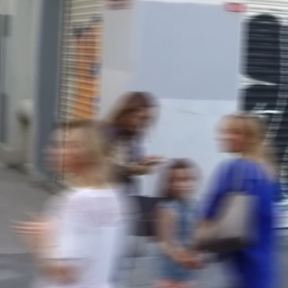}} &
 \includegraphics[width=0.9\textwidth]{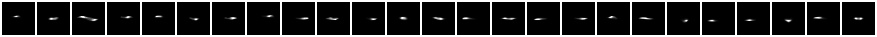}\\
 %\addlinespace
 & \includegraphics[width=0.896\textwidth]{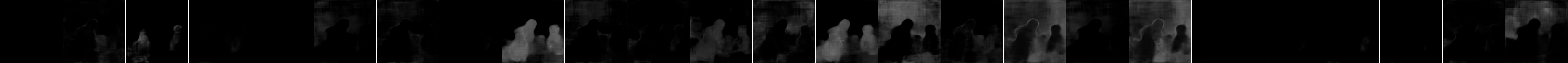}\\
 \end{tabular}
 \begin{tabular}{c@{\hspace{1em}}c}
 \multirow{2}{*}[0.26cm]{\includegraphics[width=0.07\textwidth]{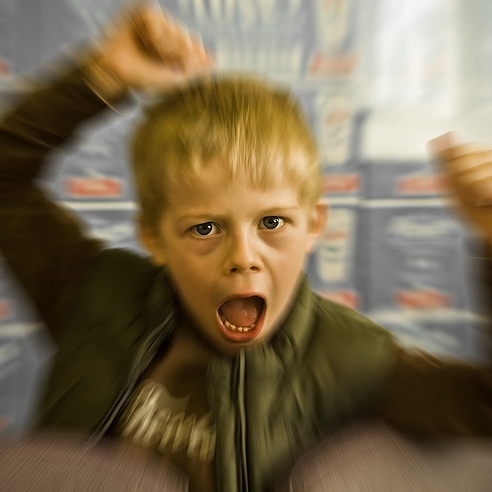}} & \includegraphics[width=0.9\textwidth]{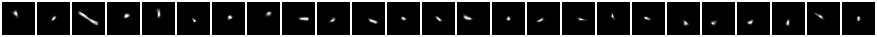}\\
 %\addlinespace
 & \includegraphics[width=0.896\textwidth]{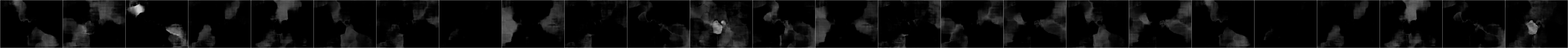}\\
 \end{tabular}
 \begin{tabular}{c@{\hspace{1em}}c}
 \multirow{2}{*}[0.26cm]{\includegraphics[width=0.07\textwidth]{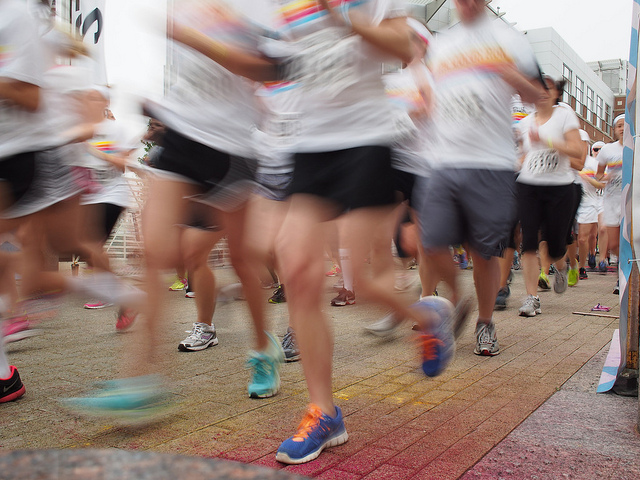}} & \includegraphics[width=0.9\textwidth]{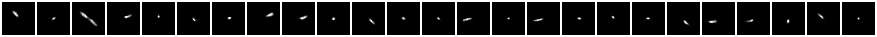}\\
 %\addlinespace
 & \includegraphics[width=0.896\textwidth]{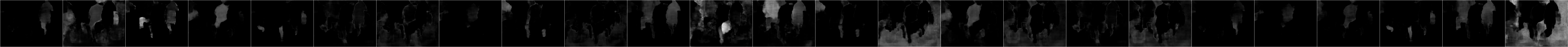}\\
 \end{tabular}
 \begin{tabular}{c@{\hspace{1em}}c}
 \multirow{2}{*}[0.26cm]{\includegraphics[width=0.07\textwidth]{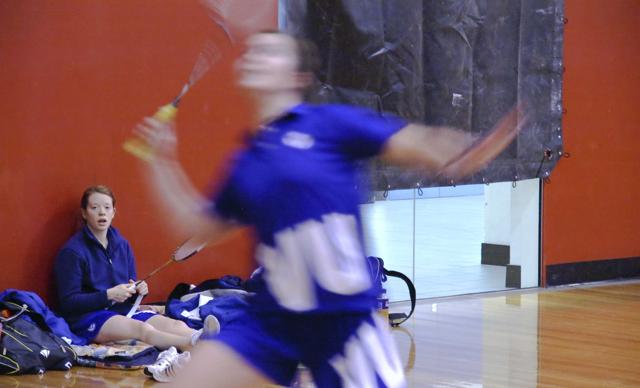}} & \includegraphics[width=0.9\textwidth]{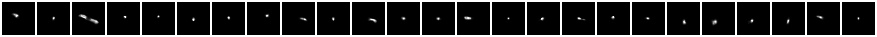}\\
 %\addlinespace
 & \includegraphics[width=0.896\textwidth]{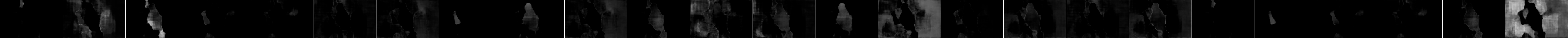}\\
 \end{tabular}
 \begin{tabular}{c@{\hspace{1em}}c}
 \multirow{2}{*}[0.26cm]{\includegraphics[width=0.07\textwidth]{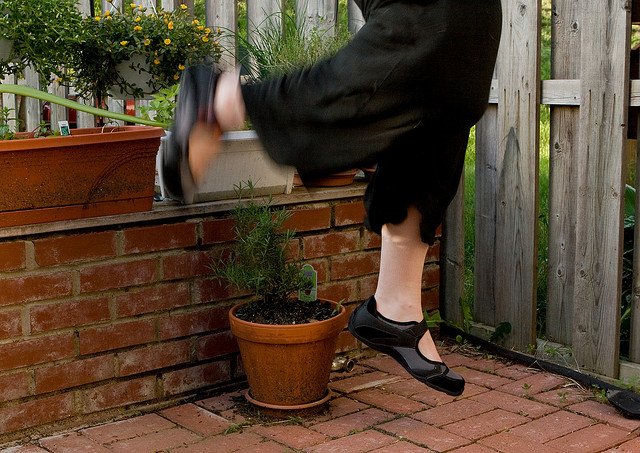}}& \includegraphics[width=0.9\textwidth]{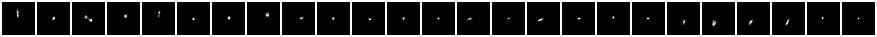}\\
 %\addlinespace
 & \includegraphics[width=0.896\textwidth]{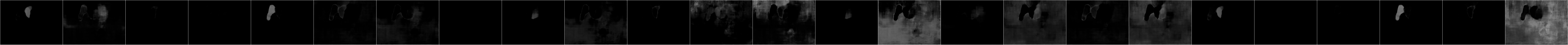}\\
 \end{tabular}
 \begin{tabular}{c@{\hspace{1em}}c}
 \multirow{2}{*}[0.26cm]{\includegraphics[width=0.07\textwidth]{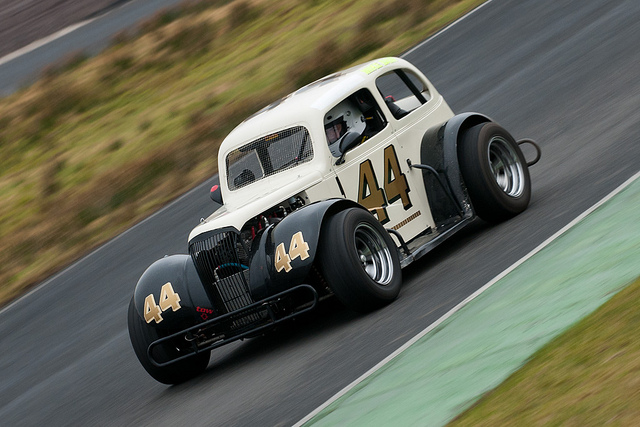}}& \includegraphics[width=0.9\textwidth]{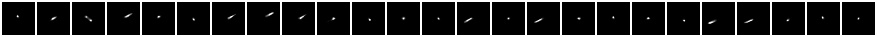}\\
 %\addlinespace
 & \includegraphics[width=0.896\textwidth]{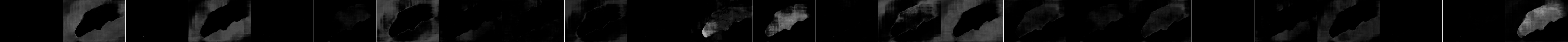}\\
 \end{tabular}
\caption{\textbf{Examples of generated kernel basis $\{\mathbf{k}^b\}$  and corresponding mixing coefficients $\{\mathbf{m}^b\}$} predicted from the blurry images shown on the left. The adaptation to the input is more notorious for the elements that have significant weights. }
    \label{fig:KernelsAndMasks2}
\end{figure*}

\subsection{Training Details}\label{sec:app:training}
We train the proposed network with a combination of the $L^2$-norm  and $L^1$-norm for a total number of 1200 epochs in two steps. In the first 300 epochs, the network is trained using an $L^2$-norm in the \emph{kernels loss}, continued by 900 epochs switching to $L^1$-norm. We minimize our objective loss using Adam optimizer with standard hyperparameters. We start with a learning rate equal to 1e-4 and halve it every 150 epochs.  

Table~\ref{tab:ablationstudy} compares the performance obtained for different numbers of basis elements and different training schemes. 

\begin{table}[h]
    \centering
    \begin{tabular}{c|c|c|c}
      Method  &   B=15 & B=25  & B=40  \\
    \hline
    L2 300 epochs  & 28.52  & 28.64  & 28.56  \\
    L2 450 epochs   &  28.64 & 28.70  & 28.86  \\
    L2 300 epochs+L1 150 epochs &  28.95 &  28.85  & 28.93      \\
    \end{tabular}
    \caption{Comparison of restoration performance for models trained with different values of $B$. Models were evaluated in the  RealBlurJ test dataset using the same restoration algorithm. Results shown are PSNRs  values.  }
    \label{tab:ablationstudy}
\end{table}

\subsection{Dataset Generation}\label{sec:app:dataset_generation}
To generate the training set for the kernel estimation network, we propose a method that generates non-uniform motion blurred images, based on the training set of the ADE20K semantic segmentation dataset~\cite{zhou2017scene}. We choose a subset of 5888 images containing at least one 
segmented person or car having a minimum size of 400 pixels. To blur the images, we use a dataset of 500,000 kernels with a maximum exposure time of one second. The dataset of kernels was generated by a camera-shake kernel generator~\cite{gavant2011physiological,delbracio2015removing} based on physiological hand tremor data. In this way we obtain for each image a tuple $\big(\mathbf{u}^{GT}, \mathbf{v}^{GT}, \{\mathbf{k}\}^{GT}, \{\mathbf{m}\}^{GT}\big)$ composed by the ground truth sharp image, the blurry image, the pairs of ground truth blur kernels and masks, respectively. Notice that the kernels $\{\mathbf{k}\}^{GT}$ are not the kernel basis as computed by our network but the resulting ground truth kernels present in the image.   A pseudo-code of the blurry synthetic dataset generation is presented Algorithm~\ref{alg:SyntheticDataset}.% Additionally,  Figure~\ref{fig:ExamplesDataset} provides some examples from our synthetic dataset.
%\begin{itemize}
 %   \item Ejemplos de nuestro dataset.
   % \item Algoritmo: explicar solo para una iamagen, require explicar que es cada cosa, comentar el codigo. en el return no hace falta devoler u. 
  %  \item añadir parte que se le aplica el blur a la mascara -> Guille :)
    %\item compararlo con el código original.
%\end{itemize}

\begin{algorithm}
\caption{Synthetic Image Generation.}

\begin{algorithmic}
\State \textbf{Input}
\State $\mathbf{u},\{\mathbf{k}\}$ \Comment{Sharp Image and dataset of kernels}
\Procedure{BlurImage}{$\mathbf{u},\{\mathbf{k}\}$}
\State $\mathbf{k}_u.append(Random(\{\mathbf{k}\}))$\Comment{ Background kernel}
\State $\mathbf{m}_u.append(ones(size(\mathbf{u})))$\Comment{Background mask}
\State $\mathbf{v}=zeros(size(\mathbf{u}))$\Comment{Initialize blurry image}
\For{$\mathbf{m}\in SegmentedObjectMasks(\mathbf{u})$}
    \State $\mathbf{k} =Random(\{\mathbf{k}\})$ \Comment{Object kernel} 
    \State $\mathbf{m}_u[0] = \mathbf{m}_u[0] -\mathbf{m}$ \Comment{Update background mask}
    \State $\mathbf{k}_u.append(\mathbf{k})$
    \State $\mathbf{m}_u.append(\mathbf{m})$
\EndFor
\For{$\mathbf{k},\mathbf{m} \in \mathbf{k}_u,\mathbf{m}_u$}
    \State $\mathbf{m} = \mathbf{m}*\mathbf{k} $\Comment{Smooth mask}
    \State $\mathbf{v} =  \mathbf{v} + \mathbf{m}(\mathbf{k} * \mathbf{u})$\Comment{Output  image update}
\EndFor
\EndProcedure
\State \textbf{return }$\mathbf{v},\mathbf{k}_u,\mathbf{m}_u $\Comment{Blurred Image, kernels and masks}
\end{algorithmic}
\label{alg:SyntheticDataset}
\end{algorithm}

\begin{figure*}[htbp]
\centering\begin{tabular}{cccc}
%\hline

 ADE dataset image &  Segmentation Masks & Kernels & Simulated Blurry Image \\ 
 & & & \\
 
\multirow{3}{*}[23pt]{\includegraphics[height=3.7cm]{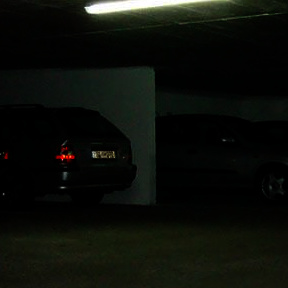}}
& 
\multirow{3}{*}[23pt]{{\includegraphics[height=3.7cm]{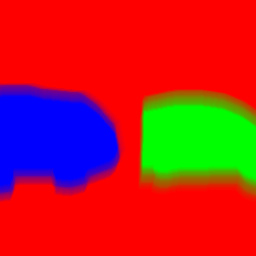}}}
 & {\fcolorbox{red}{red}{\includegraphics[height=1cm]{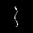}}}
 &  \multirow{3}{*}[23pt]{{\includegraphics[height=3.7cm]{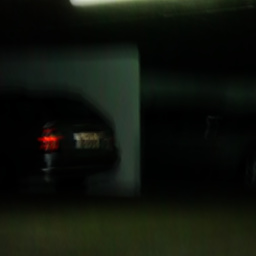}}} \\
 & & {\fcolorbox{green}{green}{\includegraphics[height=1cm]{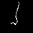}}} & \\
 & & {\fcolorbox{blue}{blue}{\includegraphics[height=1cm]{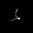}}} & \\
  & & & \\
 
 \multirow{3}{*}[23pt]{\includegraphics[height=3.7cm]{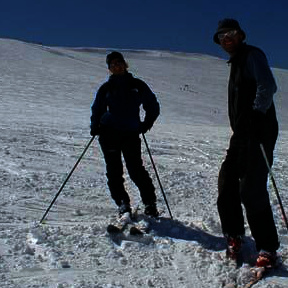}}
& 
\multirow{3}{*}[23pt]{{\includegraphics[height=3.7cm]{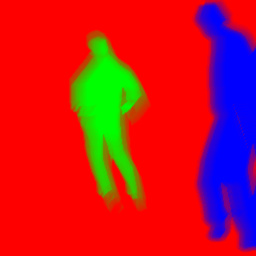}}}
 & {\fcolorbox{red}{red}{\includegraphics[height=1cm]{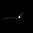}}}
 &  \multirow{3}{*}[23pt]{{\includegraphics[height=3.7cm]{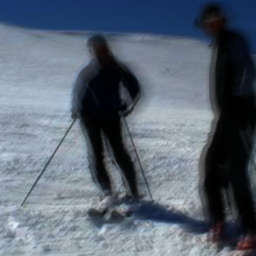}}} \\
 & & {\fcolorbox{green}{green}{\includegraphics[height=1cm]{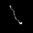}}} & \\
 & & {\fcolorbox{blue}{blue}{\includegraphics[height=1cm]{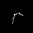}}} & \\
 
 & & & \\
 
  \multirow{3}{*}[23pt]{\includegraphics[height=3.7cm]{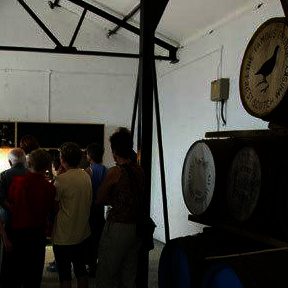}}
& 
\multirow{3}{*}[23pt]{{\includegraphics[height=3.7cm]{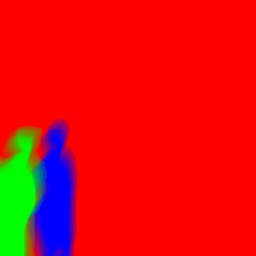}}} 
 & {\fcolorbox{red}{red}{\includegraphics[height=1cm]{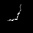}}} 
 &  \multirow{3}{*}[23pt]{{\includegraphics[height=3.7cm]{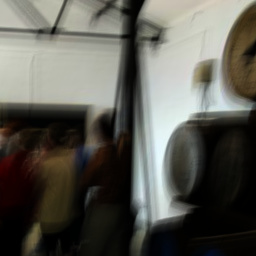}}} \\
 & & {\fcolorbox{green}{green}{\includegraphics[height=1cm]{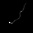}}} & \\
 & & {\fcolorbox{blue}{blue}{\includegraphics[height=1cm]{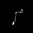}}} & \\
 
 & & & \\
 
  \multirow{3}{*}[23pt]{\includegraphics[height=3.7cm]{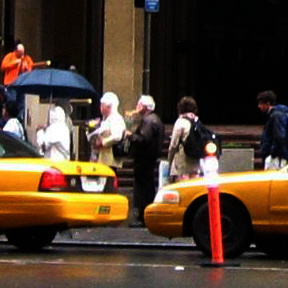}}
& 
\multirow{3}{*}[23pt]{{\includegraphics[height=3.7cm]{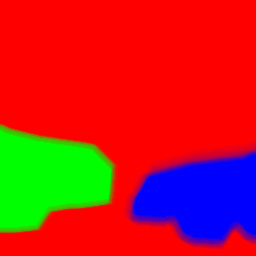}}}
 & {\fcolorbox{red}{red}{\includegraphics[height=1cm]{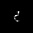}}}
 &  \multirow{3}{*}[23pt]{{\includegraphics[height=3.7cm]{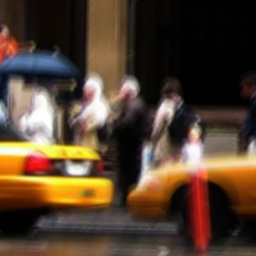}}} \\
 & & {\fcolorbox{green}{green}{\includegraphics[height=1cm]{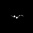}}} & \\
 & & {\fcolorbox{blue}{blue}{\includegraphics[height=1cm]{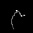}}} & \\
 \end{tabular}
  \caption{Examples of synthetic blurred images generated by the proposed procedure. From left to right: camera-shake motion kernels from physiological tremor model; image from the ADE20K segmentation dataset; corresponding ADE20K segmentation masks; resulting motion blurred image, obtained by convolving each region in the image with the corresponding kernel.}
    \label{fig:training_samples}
\end{figure*}

\subsection{Kernel Generation and Image Restoration on Real Images}\label{sec:app:qualitative}

\begin{landscape}
\begin{figure}[h]
    \setlength\tabcolsep{1.5pt} % default value: 6pt
    \centering
\begin{tabular}{cccccc}
Blurred & DMPHN~\cite{Zhang_2019_CVPR} &  SRN~\cite{tao2018scale}&  RealBlur~\cite{rim_2020_ECCV} & MPRNet~\cite{Zamir2021MPRNet} &  Ours \\ 
\includegraphics[width=0.21\textwidth]{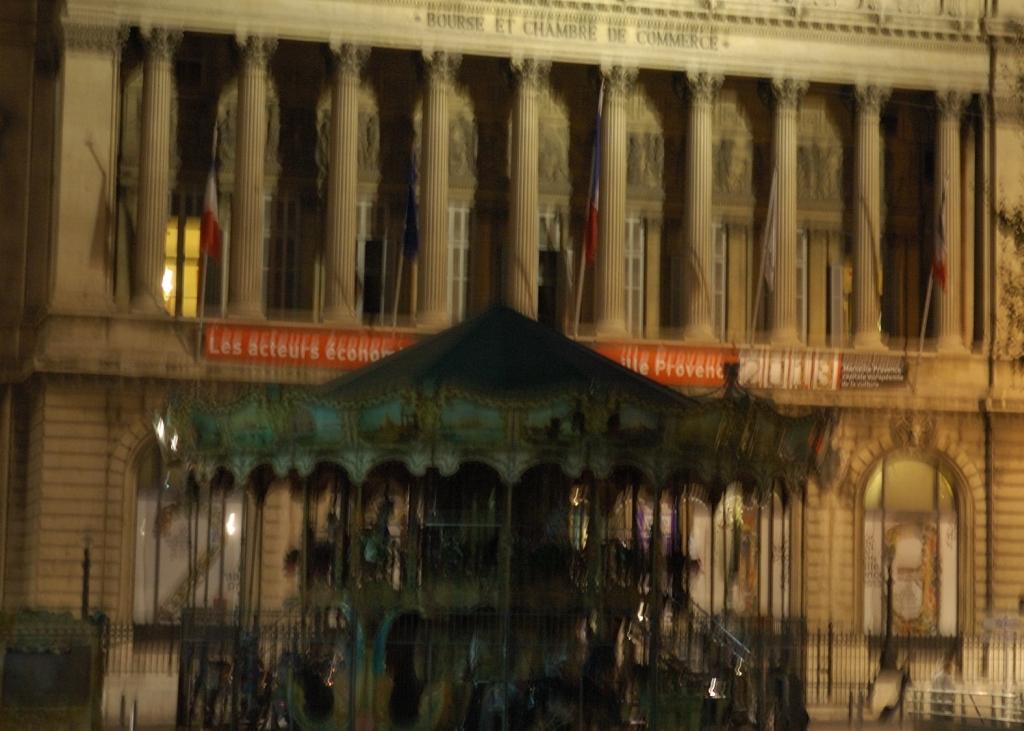}   &
\includegraphics[width=0.21\textwidth]{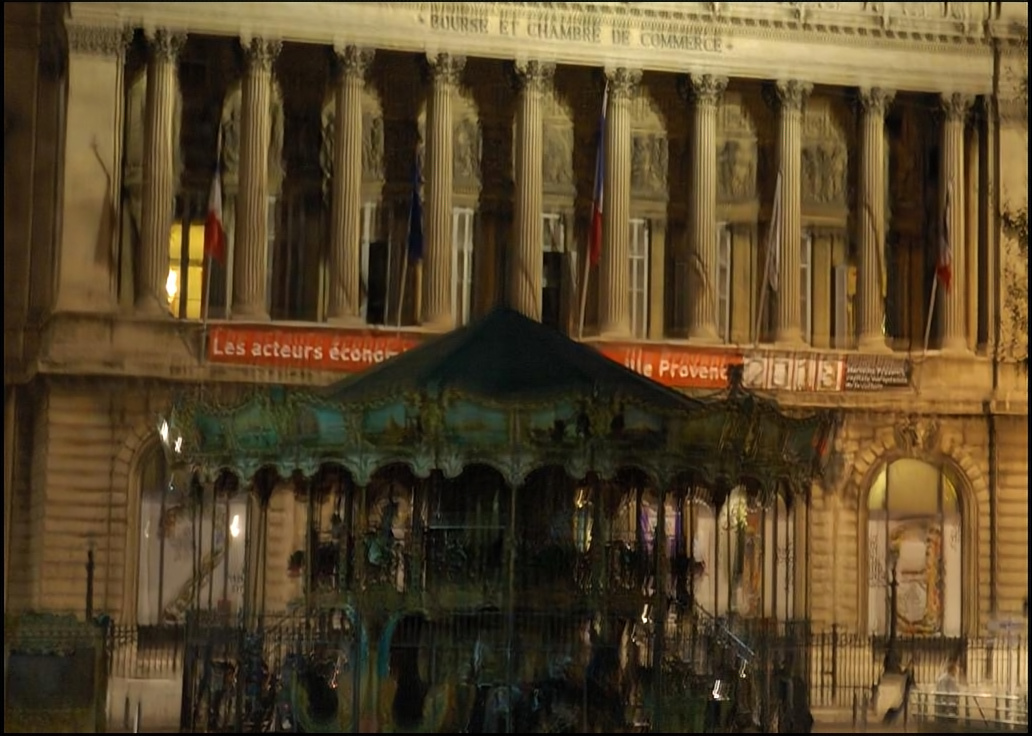} & 
\includegraphics[width=0.21\textwidth]{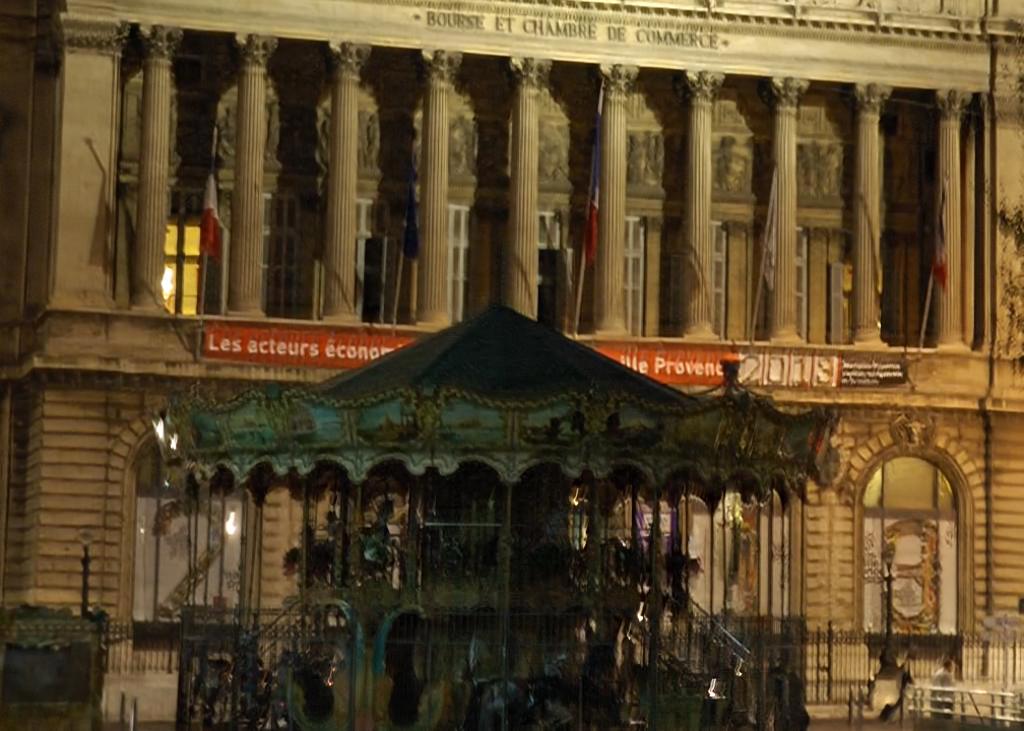}  &
\includegraphics[width=0.21\textwidth]{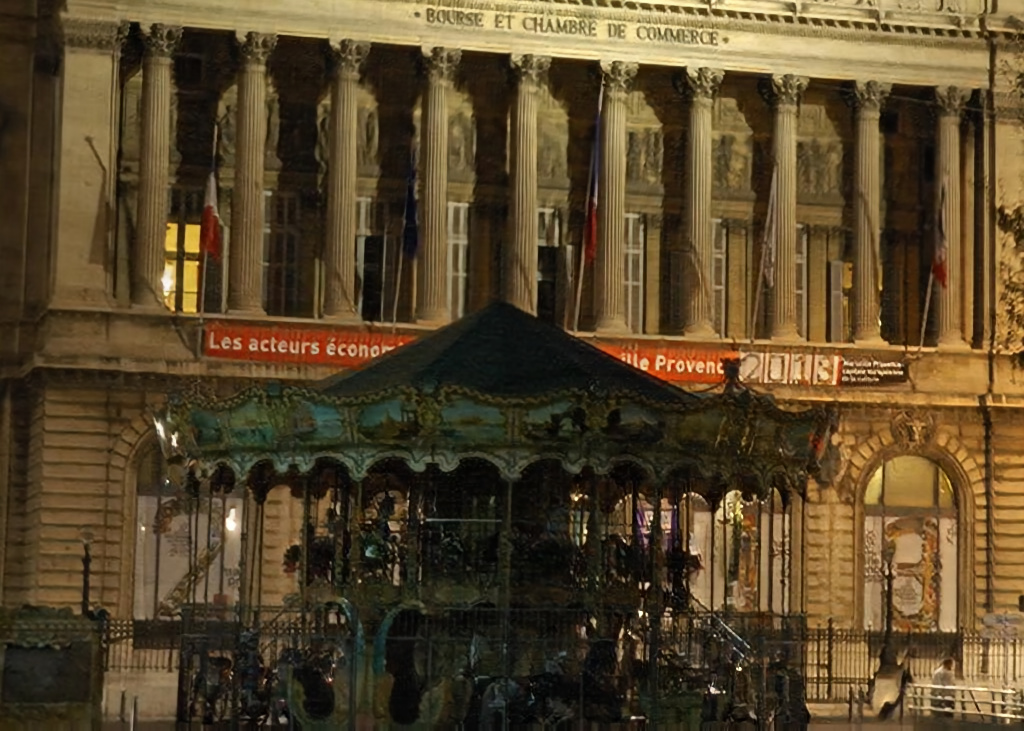}  &
\includegraphics[width=0.21\textwidth]{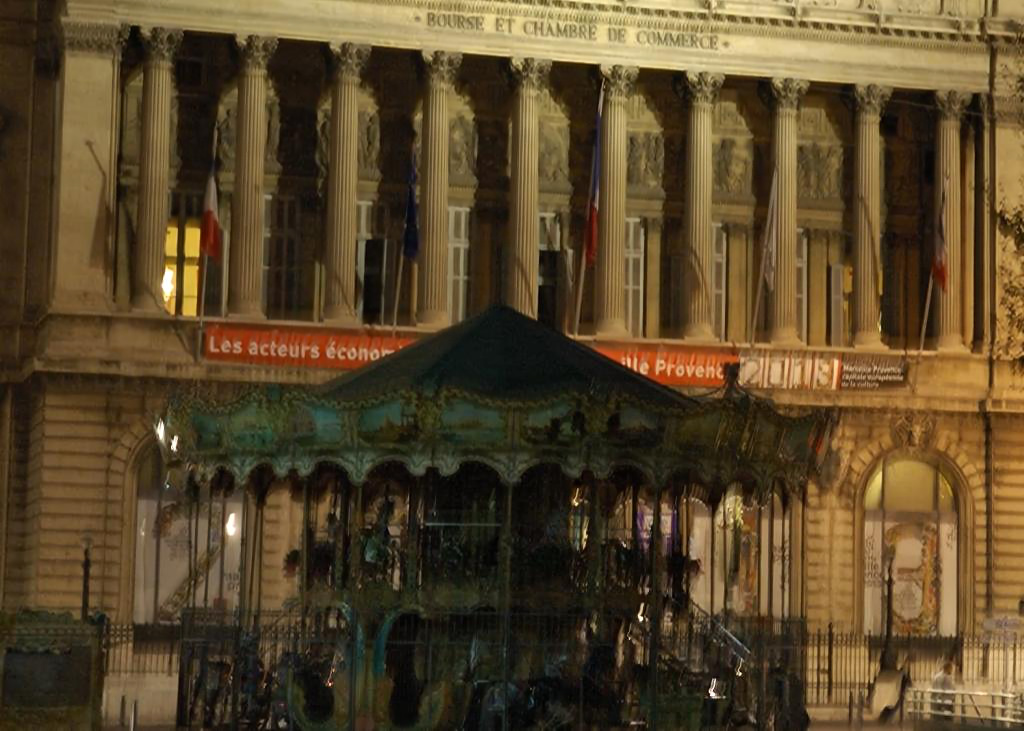}  &
\includegraphics[width=0.21\textwidth]{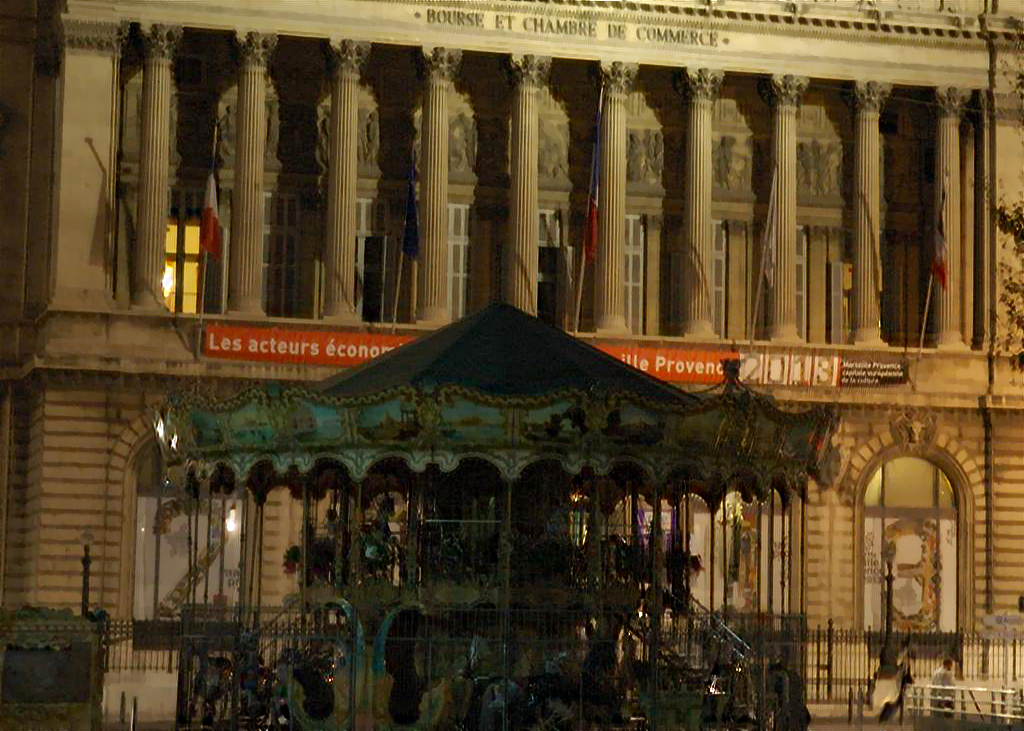}\\ % trim = left bottom right top
%%%%%%
%%%%%
%%%%%%%%%
% new one here
\includegraphics[trim=120 320 540 300,                    clip,width=0.105\textwidth]{latex/Lai/Blurry/building1.jpg}  
\includegraphics[trim=400 620 260 0, clip,width=0.105\textwidth]{latex/Lai/Blurry/building1.jpg}  &
\includegraphics[trim=120 320 540 300, clip,width=0.105\textwidth]{latex/Lai/DMPHN/building1.jpg}  
\includegraphics[trim=400 620 260 0, clip,width=0.105\textwidth]{latex/Lai/DMPHN/building1.jpg}  &
\includegraphics[trim=120 320 540 300, clip,width=0.105\textwidth]{latex/Lai/SRN/building1.jpg}  
\includegraphics[trim=400 620 260 0, clip,width=0.105\textwidth]{latex/Lai/SRN/building1.jpg}  &
\includegraphics[trim=120 320 540 300, clip,width=0.105\textwidth]{latex/Lai/RealBlur/building1.jpg} 
\includegraphics[trim=400 620 260 0, clip,width=0.105\textwidth]{latex/Lai/RealBlur/building1.jpg} &
\includegraphics[trim=120 320 540 300, clip,width=0.0925\textwidth]{latex/Lai/MPRNet/building1.jpg}
\includegraphics[trim=400 620 260 0, clip,width=0.105\textwidth]{latex/Lai/MPRNet/building1.jpg} &
\includegraphics[trim=120 320 540 300, clip,width=0.105\textwidth]{latex/Lai/ours_iccv/building1_restored.jpg}
\includegraphics[trim=400 620 260 0, clip,width=0.105\textwidth]{latex/Lai/ours_iccv/building1_restored.jpg}\\
%%%%%%%%%%%
\includegraphics[width=0.21\textwidth]{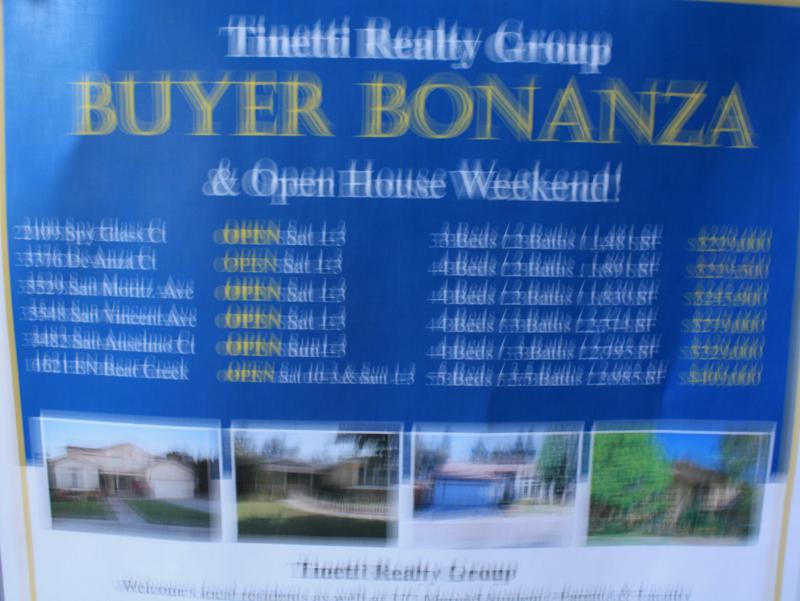}   &
\includegraphics[width=0.21\textwidth]{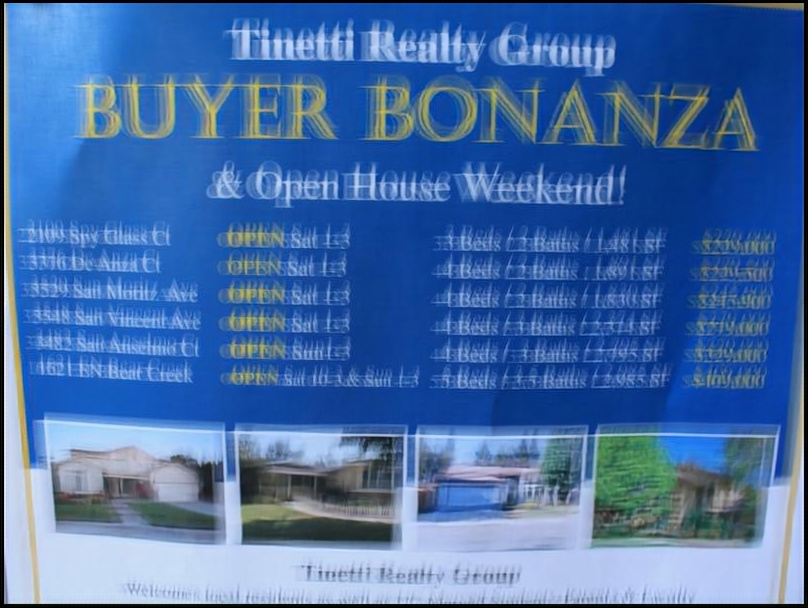} & 
\includegraphics[width=0.21\textwidth]{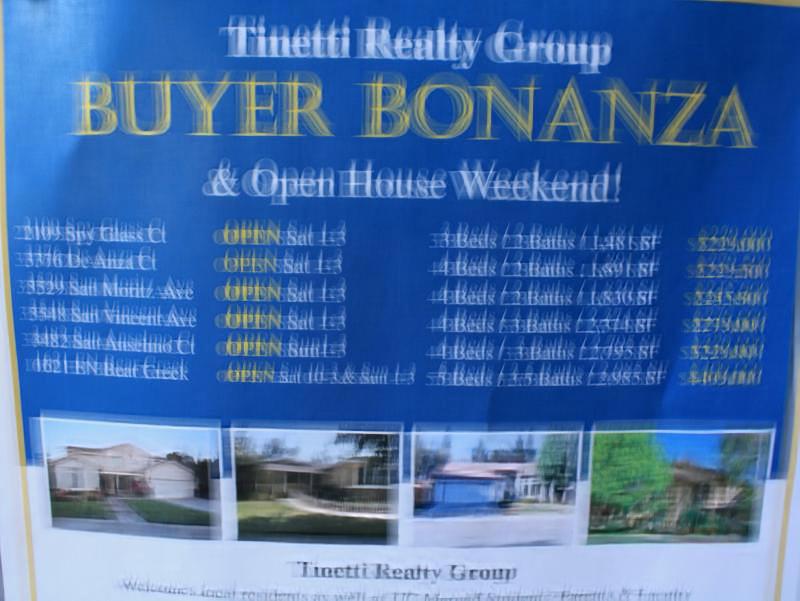}  &
\includegraphics[width=0.21\textwidth]{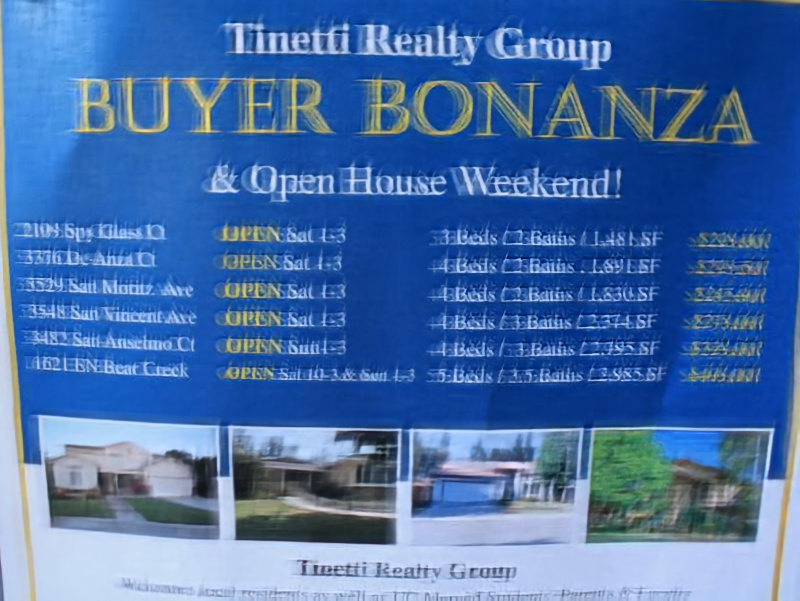}  &
\includegraphics[width=0.21\textwidth]{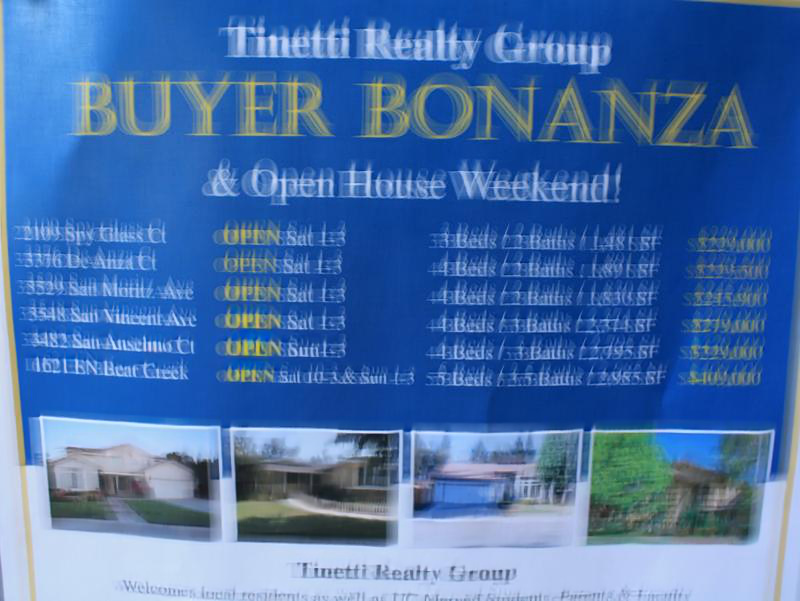}  &
\includegraphics[width=0.21\textwidth]{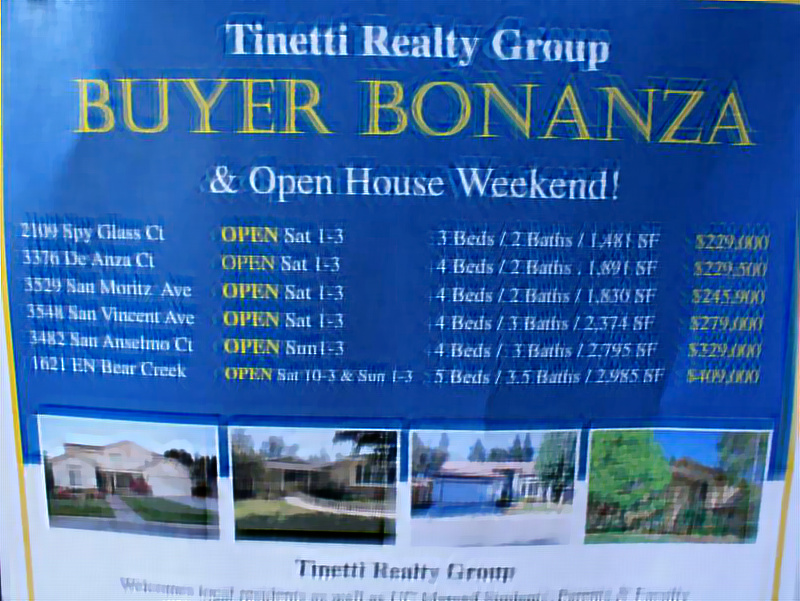}\\ % trim = left bottom right top
%%%%%%
%%%%%
%%%%%%%%%
% new one here
\includegraphics[trim=100 220 400 50,                    clip,width=0.105\textwidth]{latex/Lai/Blurry/text5.jpg}  
\includegraphics[trim=400 200 250 235, clip,width=0.105\textwidth]{latex/Lai/Blurry/text5.jpg}  &
\includegraphics[trim=100 220 400 50, clip,width=0.105\textwidth]{latex/Lai/DMPHN/text5.jpg}  
\includegraphics[trim=400 200 250 235, clip,width=0.105\textwidth]{latex/Lai/DMPHN/text5.jpg}  &
\includegraphics[trim=100 220 400 50, clip,width=0.105\textwidth]{latex/Lai/SRN/text5.jpg}  
\includegraphics[trim=400 200 250 235, clip,width=0.105\textwidth]{latex/Lai/SRN/text5.jpg}  &
\includegraphics[trim=100 220 400 50, clip,width=0.105\textwidth]{latex/Lai/RealBlur/text5.jpg} 
\includegraphics[trim=400 200 250 235, clip,width=0.105\textwidth]{latex/Lai/RealBlur/text5.jpg} &
\includegraphics[trim=100 220 400 50, clip,width=0.105\textwidth]{latex/Lai/MPRNet/text5.jpg} 
\includegraphics[trim=400 200 250 235, clip,width=0.105\textwidth]{latex/Lai/MPRNet/text5.jpg} &
\includegraphics[trim=100 220 400 50, clip,width=0.105\textwidth]{latex/Lai/ours_iccv/text5_restored.jpg}
\includegraphics[trim=400 200 250 235, clip,width=0.105\textwidth]{latex/Lai/ours_iccv/text5_restored.jpg}\\
%%%%%%%%%%%
\includegraphics[width=0.21\textwidth]{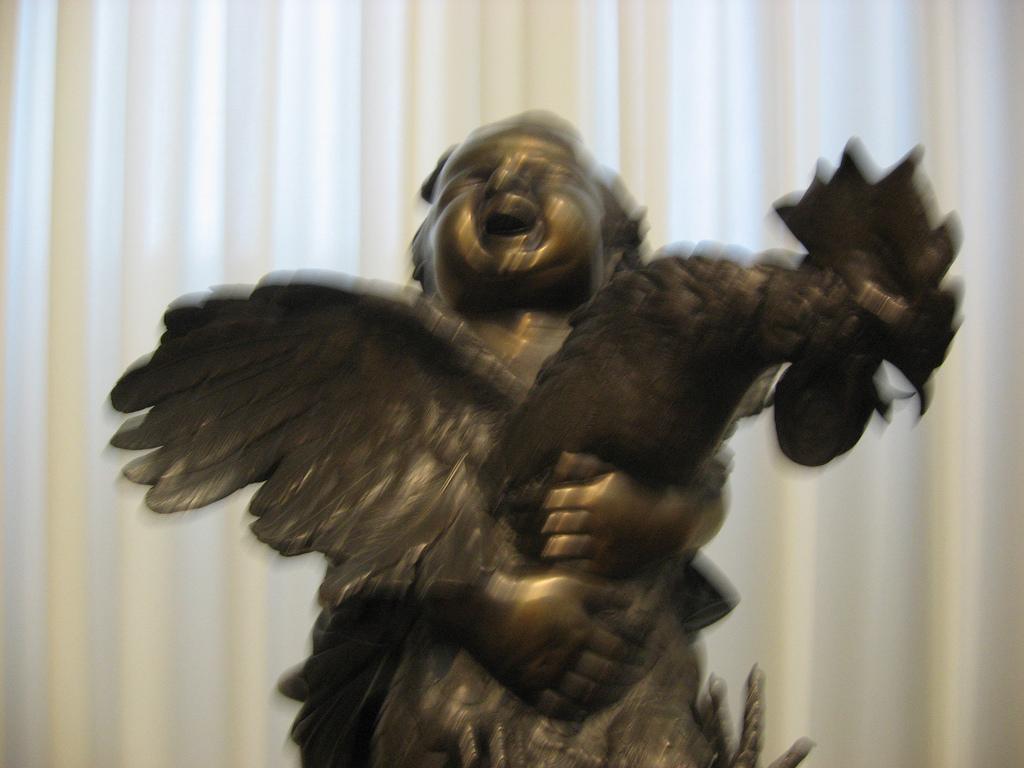}   &
\includegraphics[width=0.21\textwidth]{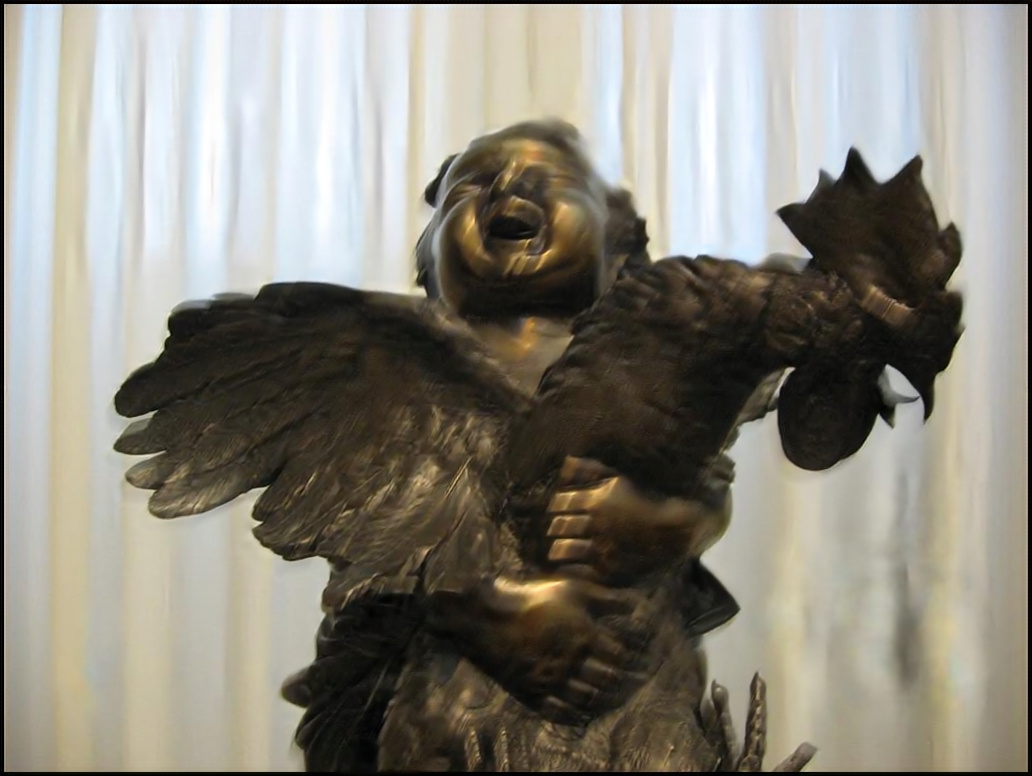} & 
\includegraphics[width=0.21\textwidth]{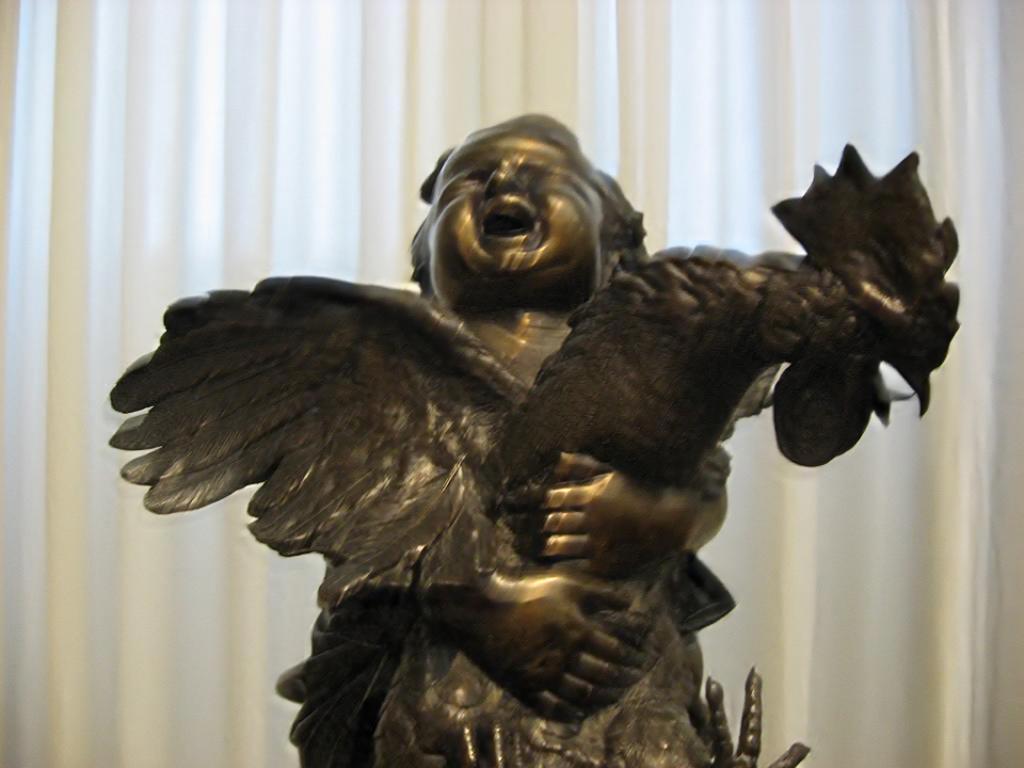}  &
\includegraphics[width=0.21\textwidth]{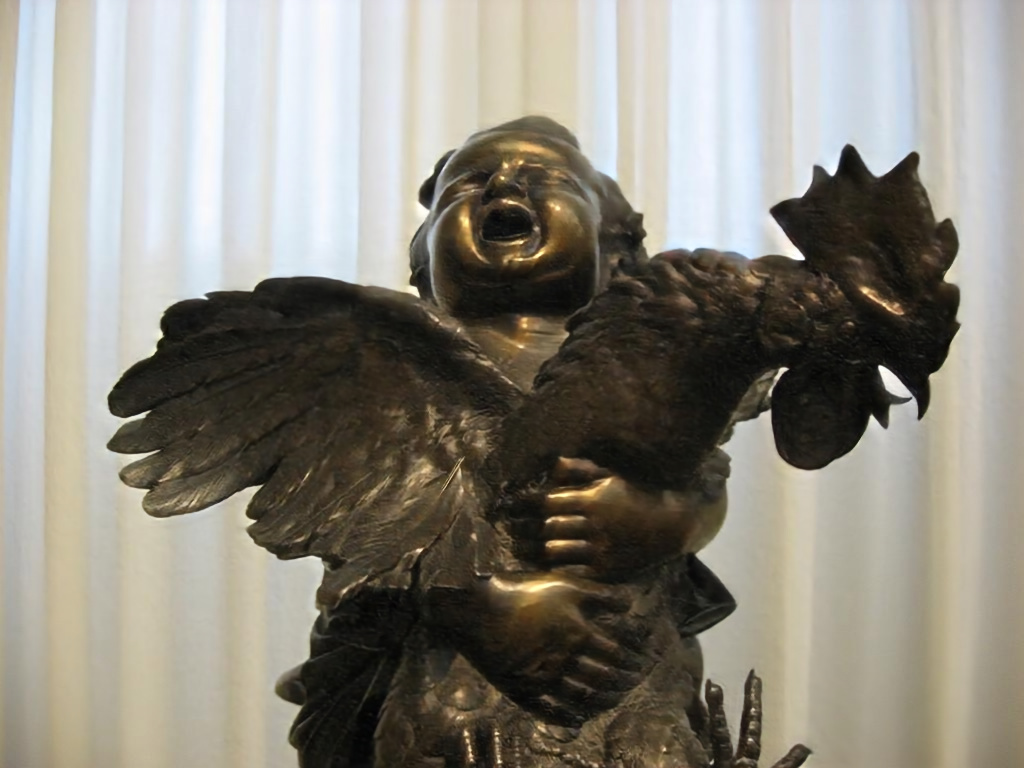}  &
\includegraphics[width=0.21\textwidth]{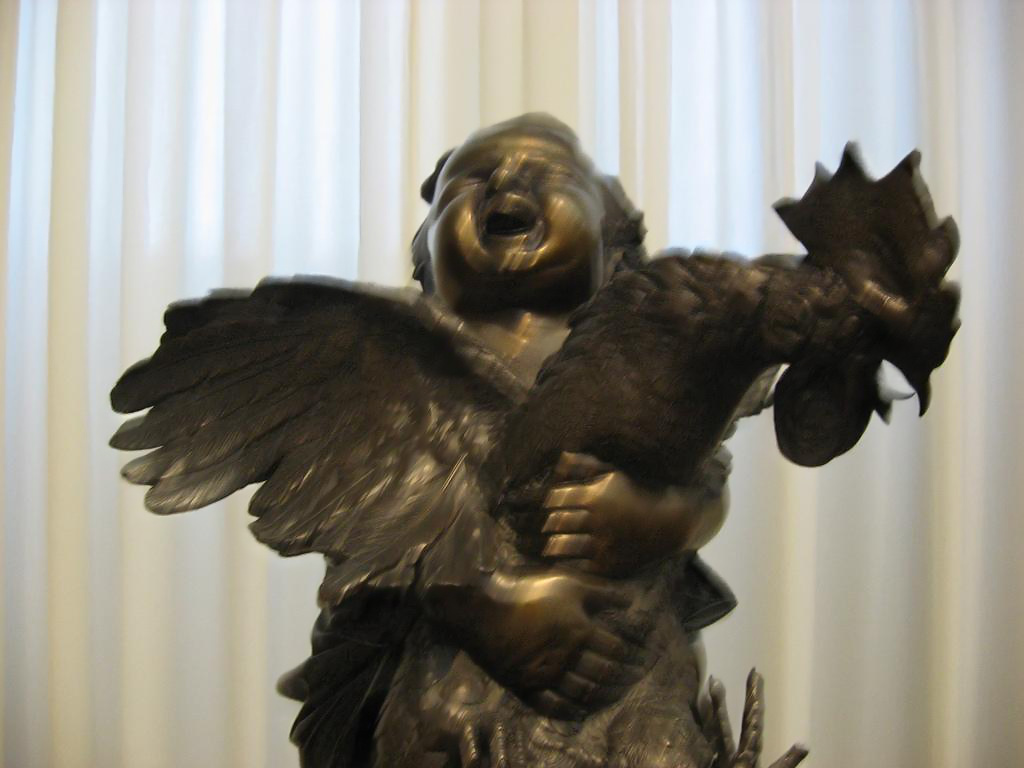}  &
\includegraphics[width=0.21\textwidth]{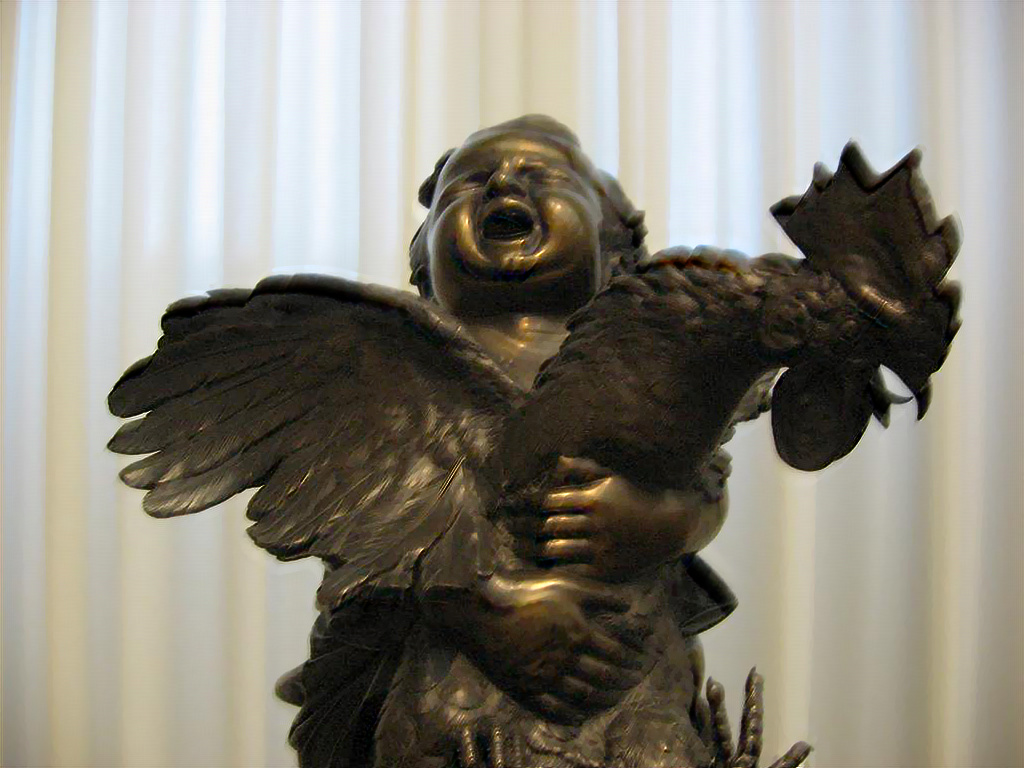}\\
\includegraphics[trim=550 270 50 130, clip,width=0.105\textwidth]{latex/Lai/Blurry/boy_statue.jpg}  
\includegraphics[trim=350 350 250 50, clip,width=0.105\textwidth]{latex/Lai/Blurry/boy_statue.jpg}  &
\includegraphics[trim=550 270 50 130, clip,width=0.105\textwidth]{latex/Lai/DMPHN/boy_statue.jpg}  
\includegraphics[trim=350 350 250 50, clip,width=0.105\textwidth]{latex/Lai/DMPHN/boy_statue.jpg}  &
\includegraphics[trim=550 270 50 130, clip,width=0.105\textwidth]{latex/Lai/SRN/boy_statue.jpg}  
\includegraphics[trim=350 350 250 50, clip,width=0.105\textwidth]{latex/Lai/SRN/boy_statue.jpg}  &
\includegraphics[trim=550 270 50 130, clip,width=0.105\textwidth]{latex/Lai/RealBlur/boy_statue.jpg} 
\includegraphics[trim=350 350 250 50, clip,width=0.105\textwidth]{latex/Lai/RealBlur/boy_statue.jpg} &
\includegraphics[trim=550 270 50 130, clip,width=0.105\textwidth]{latex/Lai/MPRNet/boy_statue.jpg} 
\includegraphics[trim=350 350 250 50, clip,width=0.105\textwidth]{latex/Lai/MPRNet/boy_statue.jpg} &
\includegraphics[trim=550 270 50 130, clip,width=0.105\textwidth]{latex/Lai/ours_iccv/boy_statue_restored.jpg}
\includegraphics[trim=350 350 250 50, clip,width=0.105\textwidth]{latex/Lai/ours_iccv/boy_statue_restored.jpg}\\
%%%%%%%%%%%
\includegraphics[width=0.21\textwidth]{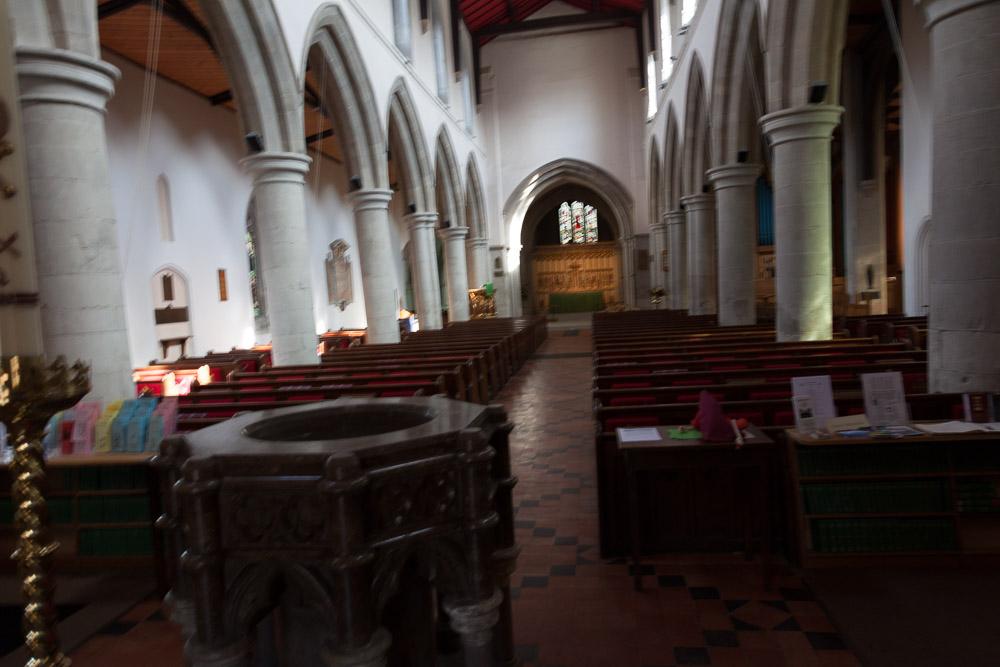}   &
\includegraphics[width=0.21\textwidth]{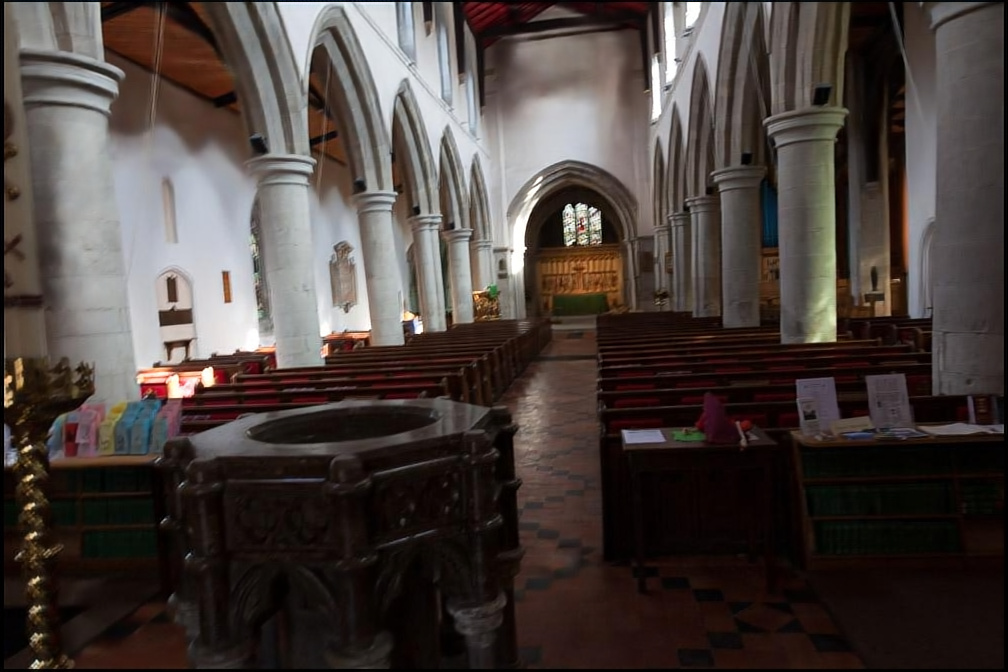} & 
\includegraphics[width=0.21\textwidth]{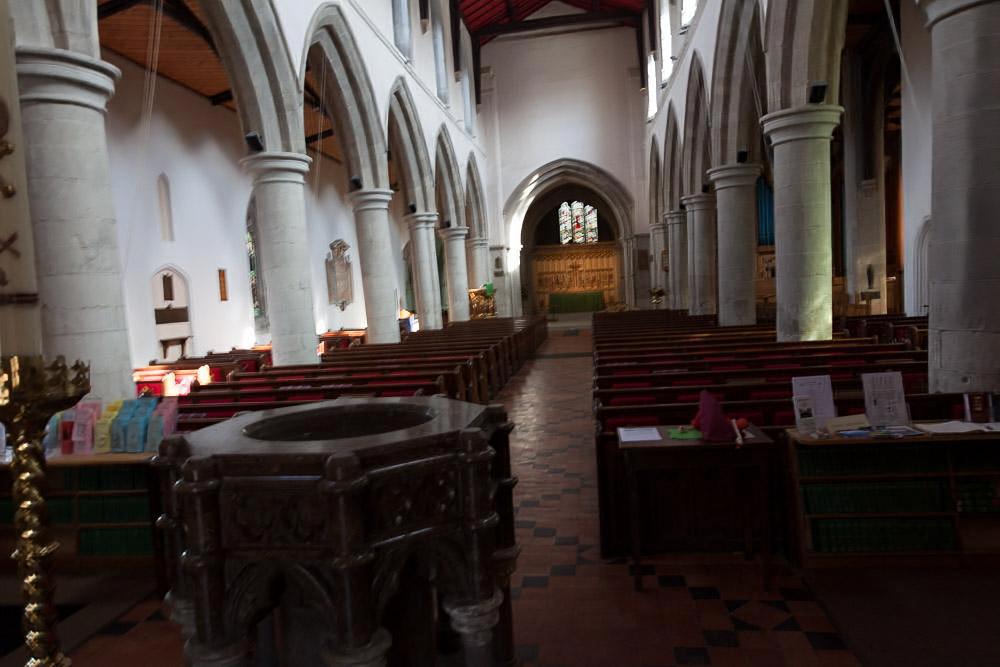}  &
\includegraphics[width=0.21\textwidth]{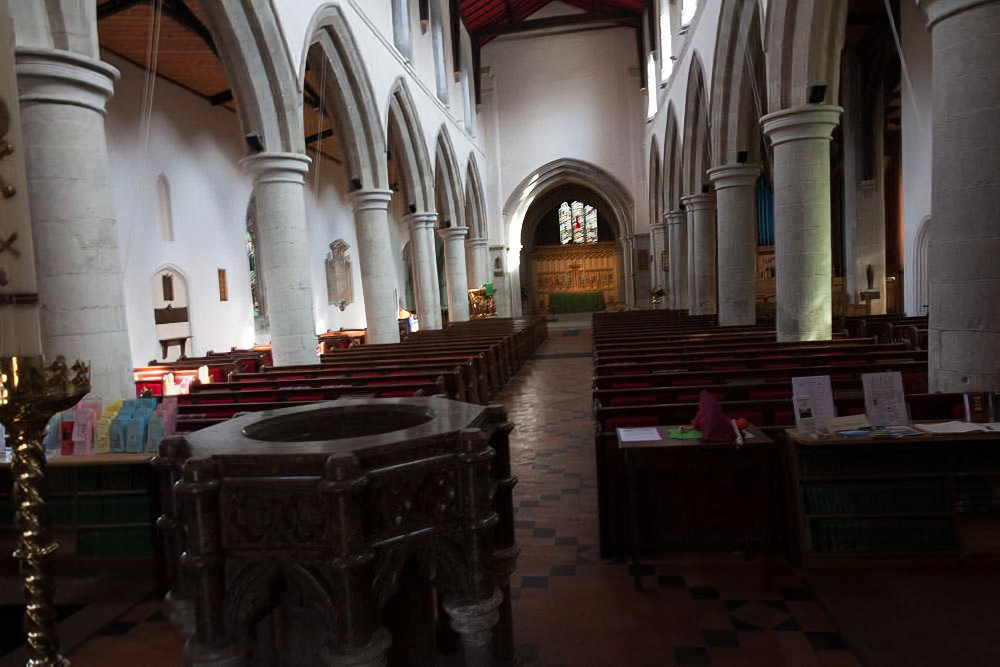}  &
\includegraphics[width=0.21\textwidth]{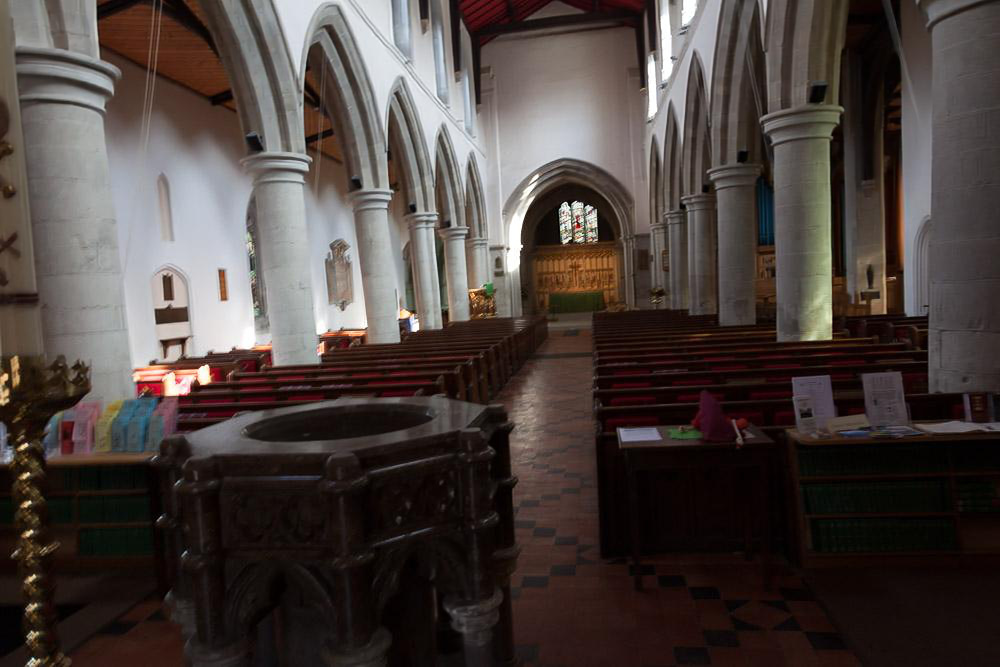}  &
\includegraphics[width=0.21\textwidth]{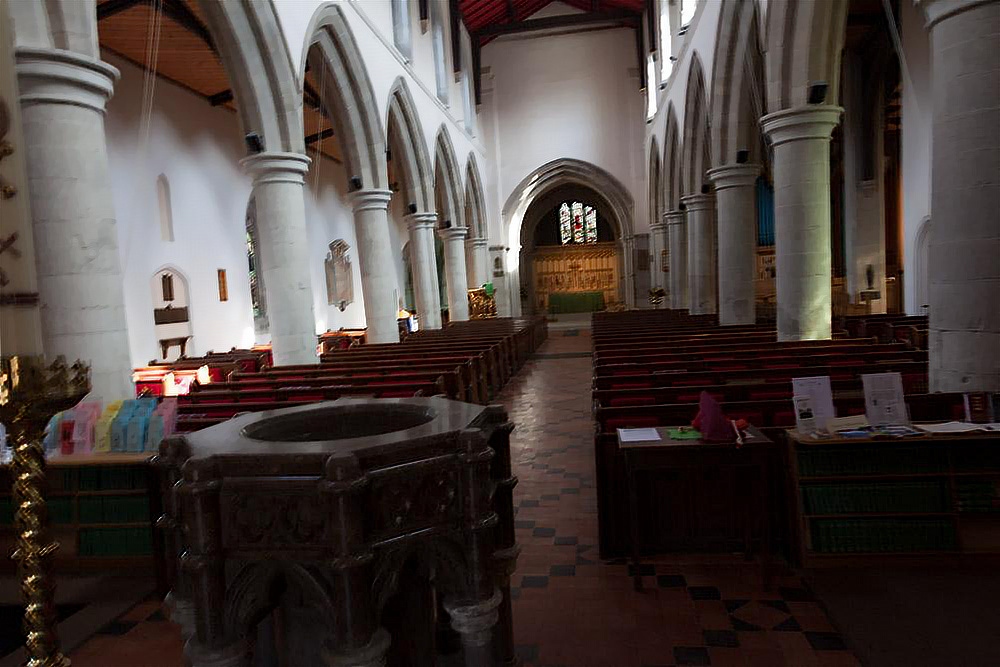}\\ % trim = left bottom right top
%%%%%%
%%%%%
%%%%%%%%%
% new one here
\includegraphics[trim=500 400 350 175,                    clip,width=0.21\textwidth]{latex/Lai/Blurry/church.jpg}  
&
\includegraphics[trim=500 400 350 175,                    clip,width=0.21\textwidth]{latex/Lai/DMPHN/church.jpg}  
&
\includegraphics[trim=500 400 350 175,                    clip,width=0.21\textwidth]{latex/Lai/SRN/church.jpg}  
&
\includegraphics[trim=500 400 350 175,                    clip,width=0.21\textwidth]{latex/Lai/RealBlur/church.jpg} 
&
\includegraphics[trim=500 400 350 175,                    clip,width=0.21\textwidth]{latex/Lai/MPRNet/church.jpg}
&
\includegraphics[trim=500 400 350 175,                    clip,width=0.21\textwidth]{latex/Lai/ours_iccv/church_restored.jpg}
%\includegraphics[trim=600 0 0 301, clip,width=0.105\textwidth]{latex/Lai/ours_iccv/church_restored.jpg}\\
%%%%%%%%%%%
% trim = left bottom right top
\end{tabular} 
    \caption{{Deblurring examples on real blurry images from Lai's dataset~\cite{lai2016comparative}.} 
   % Images from our algorithm were generated with $\alpha=0.9$.
    }
    \label{fig:ComparisonLaiDeepLearning}
\end{figure}
\end{landscape}

Figure~\ref{fig:ComparisonLaiDeepLearning} compares our results to those obtained with state-of-the-art deep learning-based methods on Lai's dataset. A comparison of our results on the same database with existing non-uniform motion blur estimation methods is presented in Figures~\ref{fig:comparisonLaiHalfSize} and  \ref{fig:comparisonLaiHalfSize2}. Figure~\ref{fig:comparisonKohler} shows additional deblurring examples from K\"{o}hler's dataset.

Additional examples of generated kernels together with the estimated deblurred images on RealBlur~\cite{rim_2020_ECCV} are shown in Figure~\ref{fig:comparisonRealBlur}. On the RealBlur examples, results are compared to other kernel estimation methods (Gong \etal~\cite{gong2017motion} and Sun \etal~\cite{sun2015learning}). Notice that, contrarily to these approaches, the kernels estimated by our method show almost no correlation with the image structure. Moreover, our approach is capable of retrieving better estimates in regions exhibiting low contrast, which are often present in motion blurred images suffering from limited exposure time.   

\begin{landscape}
\begin{figure}
\centering
\begin{tabular}{cc|cc|cc|cc}
\multicolumn{8}{c}{Input blurry images} \\
\includegraphics[width=0.15\textwidth]{latex/Lai/Blurry_HS/car2.jpg} &  &
\includegraphics[width=0.15\textwidth]{latex/Lai/Blurry_HS/car5.jpg}  & &
\includegraphics[width=0.15\textwidth]{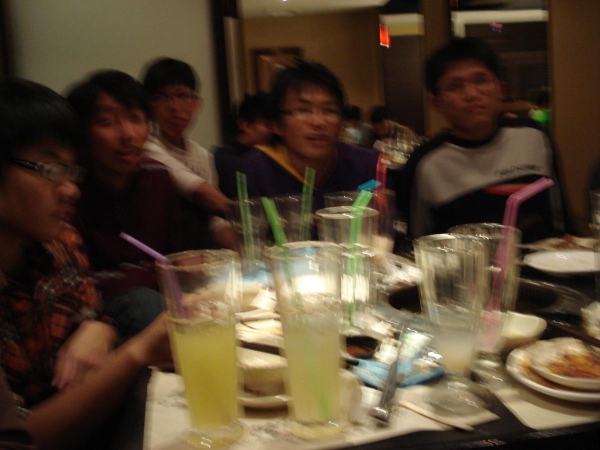} & &
\includegraphics[width=0.15\textwidth]{latex/Lai/Blurry_HS/night2.jpg} & \\
\multicolumn{8}{c}{Our reconstructions and kernels} \\
% Ours kernels
\includegraphics[width=0.15\textwidth]{latex/Lai/ours_iccv_HS_combined/car2_restored.jpg} &
\includegraphics[width=0.15\textwidth]{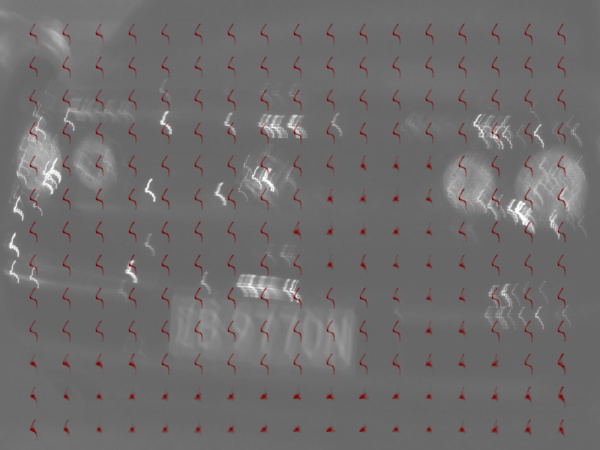} &

\includegraphics[width=0.15\textwidth]{latex/Lai/ours_iccv_HS_combined/car5_restored.jpg} &
\includegraphics[width=0.15\textwidth]{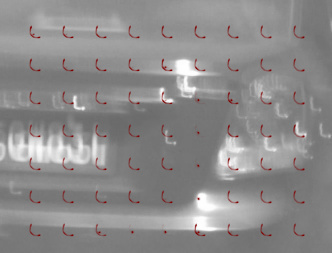} &
\includegraphics[width=0.15\textwidth]{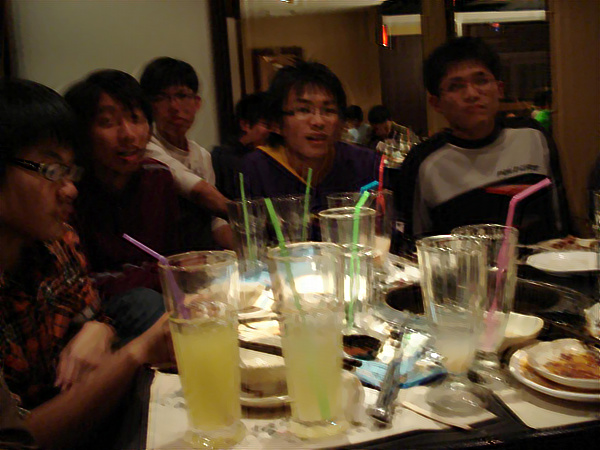}  & 
\includegraphics[width=0.15\textwidth]{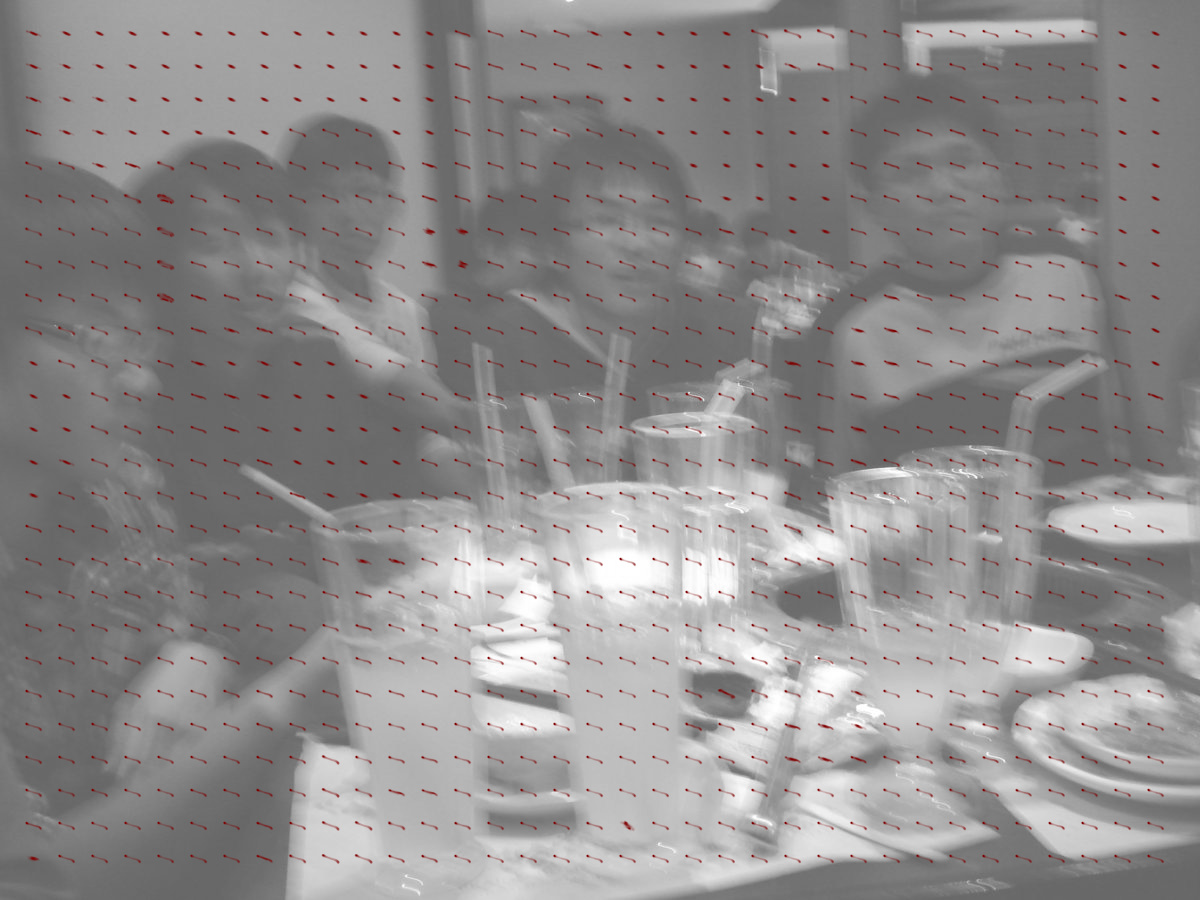} & 
\includegraphics[width=0.15\textwidth]{latex/Lai/ours_iccv_HS_combined/night2_restored.jpg} &
\includegraphics[width=0.15\textwidth]{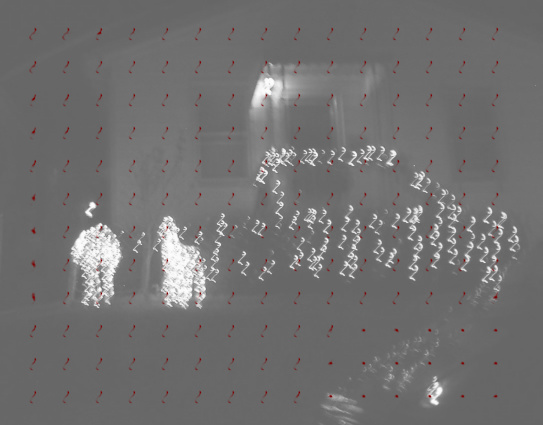} \\

\multicolumn{8}{c}{Gong \etal~\cite{gong2017motion} reconstructions and kernels} \\
\includegraphics[width=0.15\textwidth]{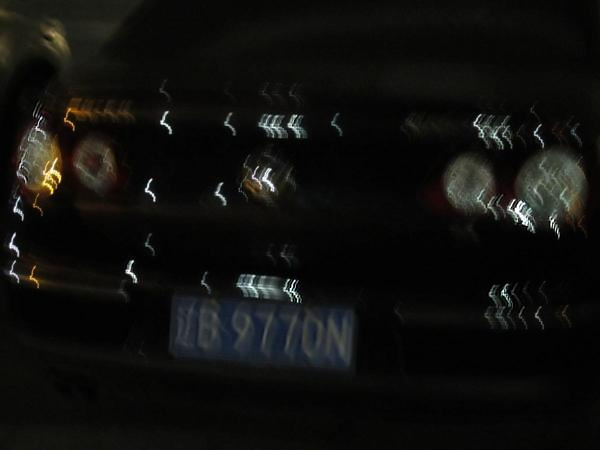} &
\includegraphics[width=0.15\textwidth]{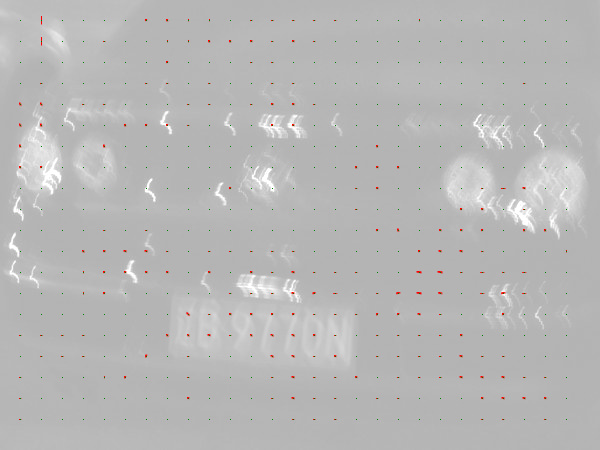}&
\includegraphics[width=0.15\textwidth]{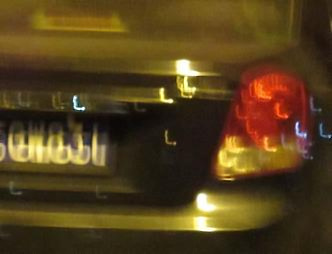} &
\includegraphics[width=0.15\textwidth]{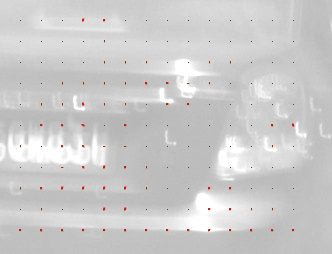} &
\includegraphics[width=0.15\textwidth]{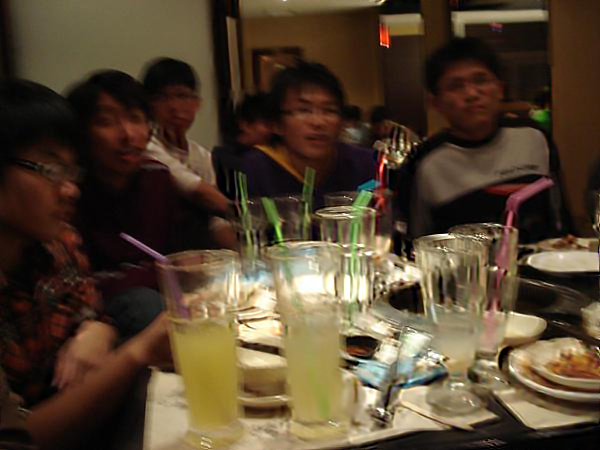} &
\includegraphics[width=0.15\textwidth]{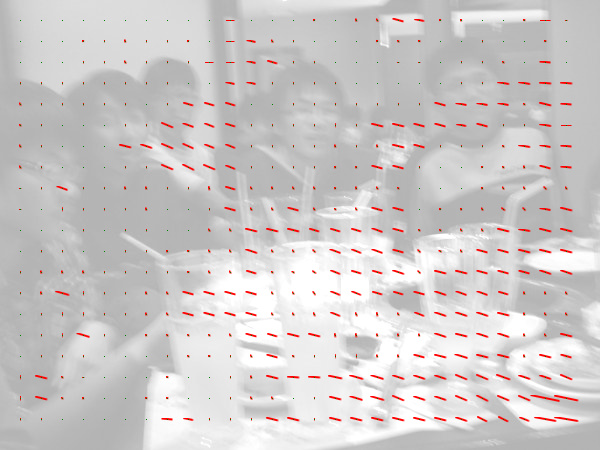} & 
\includegraphics[width=0.15\textwidth]{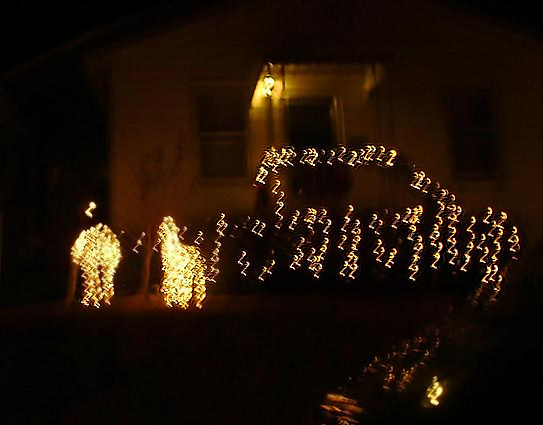} &
\includegraphics[width=0.15\textwidth]{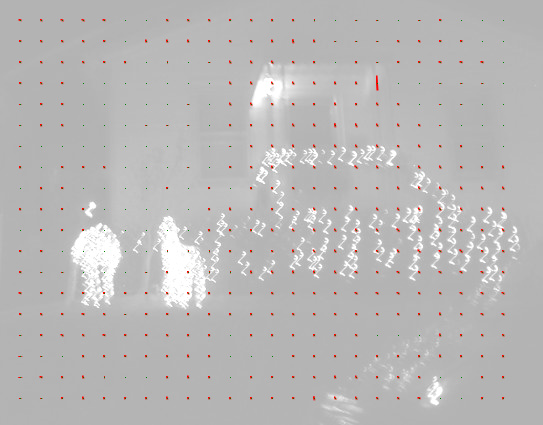}  \\  % Sun kernels
\multicolumn{8}{c}{Sun \etal~\cite{sun2015learning} reconstructions and kernels} \\

\includegraphics[width=0.15\textwidth]{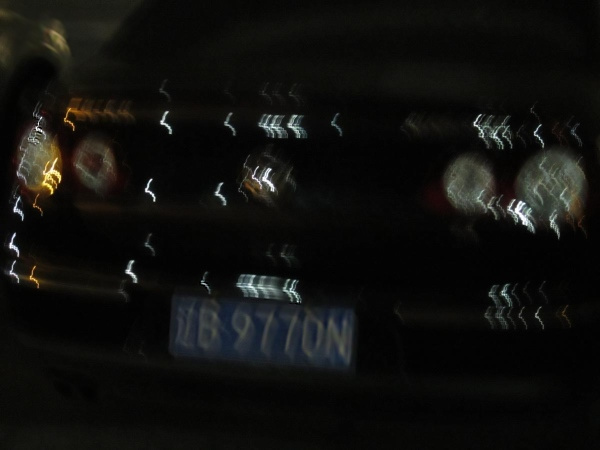} &
\includegraphics[width=0.15\textwidth]{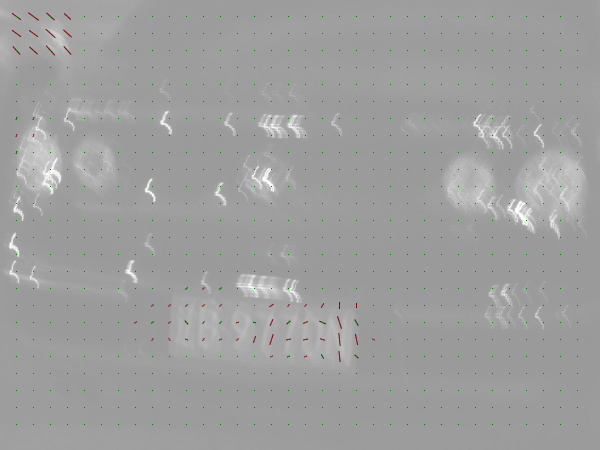} &

\includegraphics[width=0.15\textwidth]{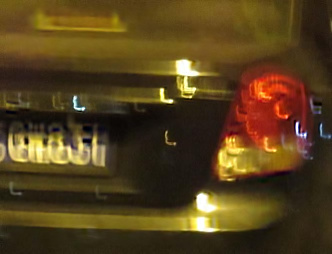} &
\includegraphics[width=0.15\textwidth]{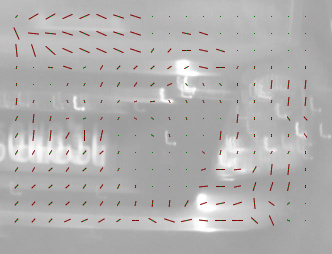} &
\includegraphics[width=0.15\textwidth]{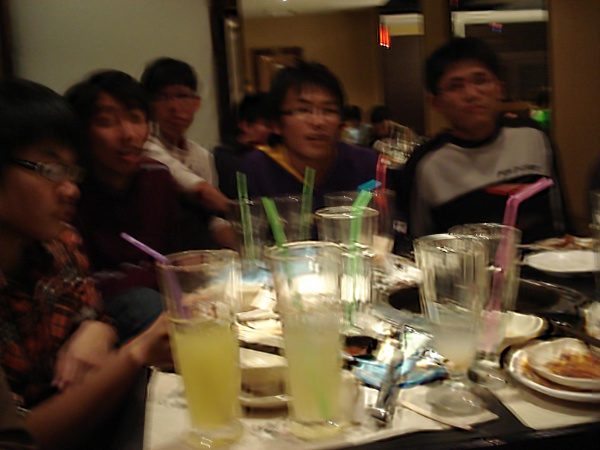} &
\includegraphics[width=0.15\textwidth]{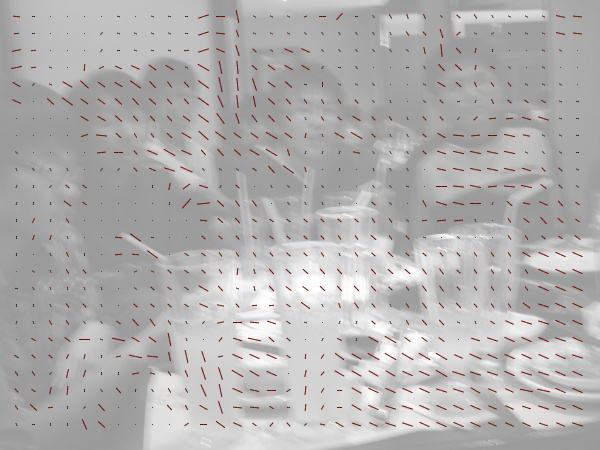} & 
\includegraphics[width=0.15\textwidth]{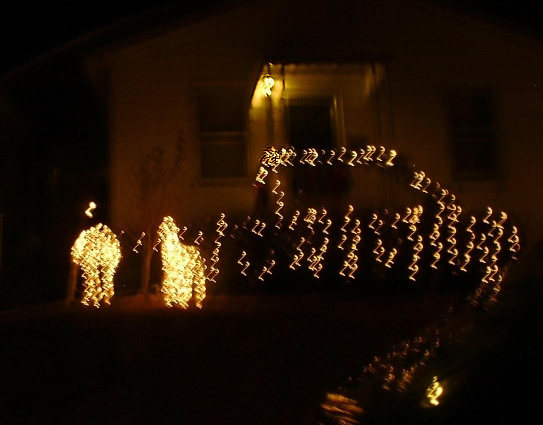}  &
\includegraphics[width=0.15\textwidth]{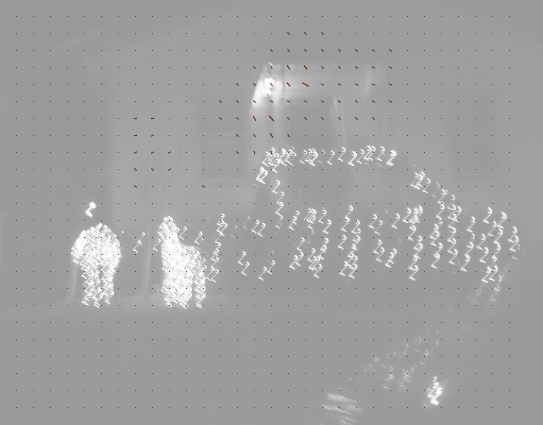} 
\\

%Blur & our kernels & our reconstruction & gong kernels & gong reconstruction & sun kernels & sun reconstruction
\end{tabular}
\caption{Examples of kernels predicted by our method and corresponding deblurred images on the Lai dataset~\cite{lai2016comparative} at half resolution. Comparison with~\cite{gong2017motion} and~\cite{sun2015learning}. Note that these approaches show a significant correlation with the image structure, and are more prone to fail at capturing the motion structure in low contrasted regions.
}
 \label{fig:app:lai}
\end{figure}
\end{landscape}

\begin{landscape}

\begin{figure}
\begin{tabular}{cc|cc|cc|cc}
\multicolumn{8}{c}{Input blurry images} \\
\includegraphics[height=0.115\textheight]{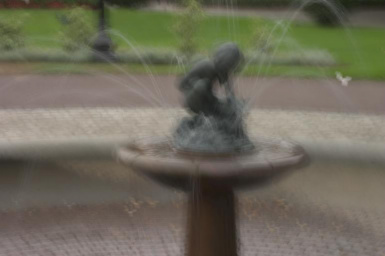} & 
&
\includegraphics[height=0.115\textheight]{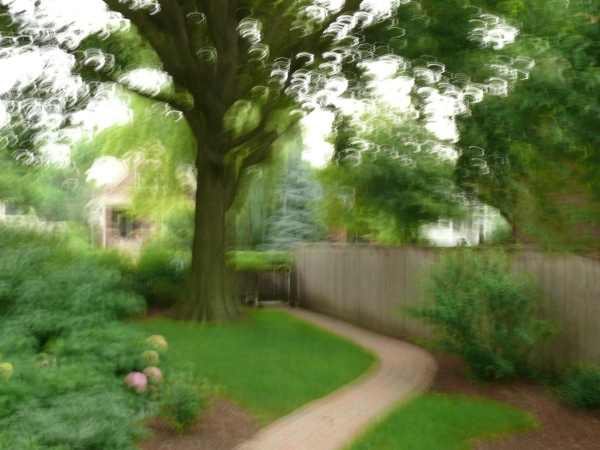}  &
&
\includegraphics[height=0.115\textheight]{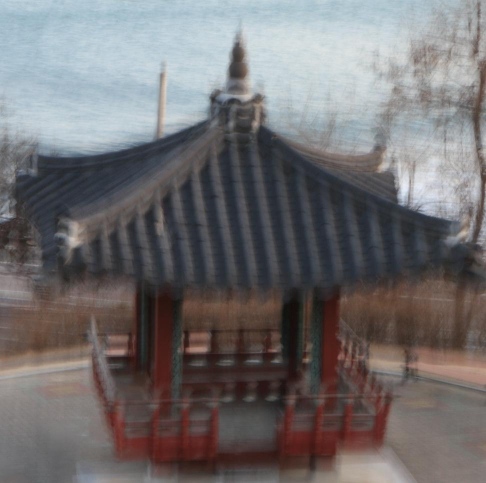} & &
\includegraphics[height=0.115\textheight]{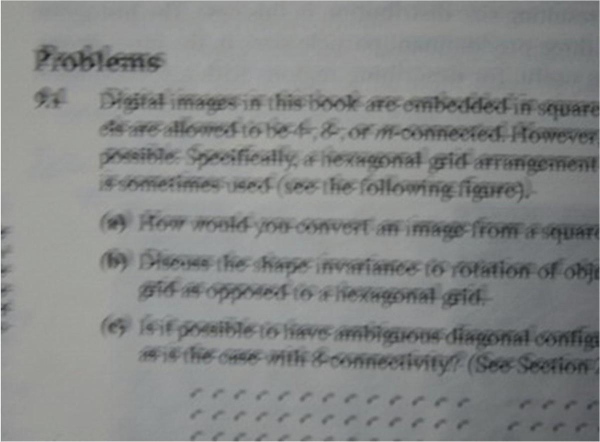} &\\
\multicolumn{8}{c}{Our reconstructions and kernels}  \\
\includegraphics[height=0.115\textheight]{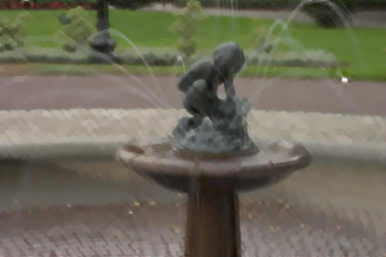} & 
\includegraphics[height=0.115\textheight]{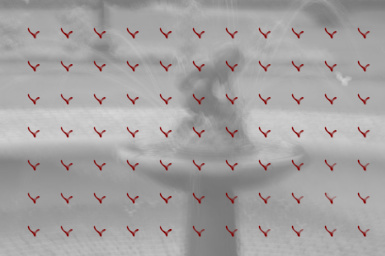} & 
\includegraphics[height=0.115\textheight]{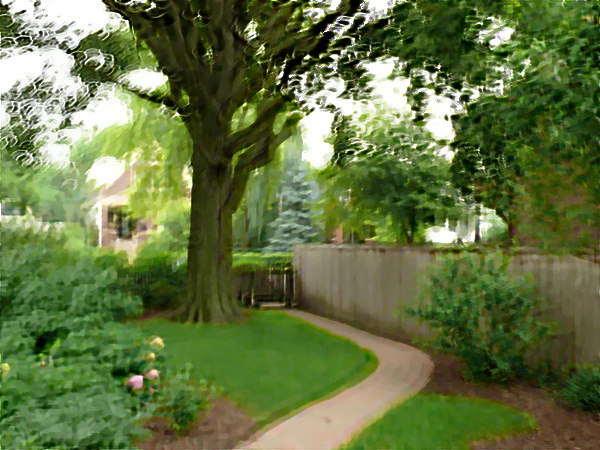}&
\includegraphics[height=0.115\textheight]{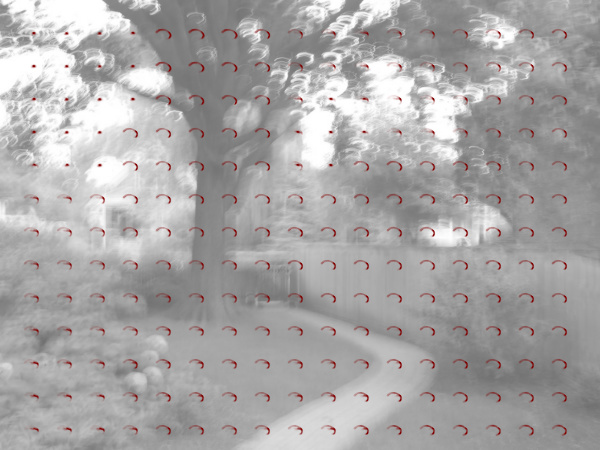} &
\includegraphics[height=0.115\textheight]{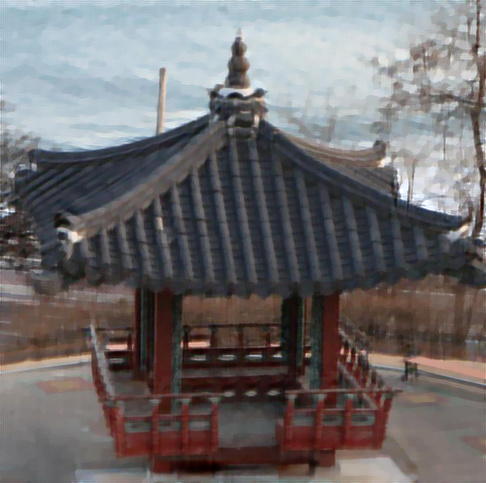} & 
\includegraphics[height=0.115\textheight]{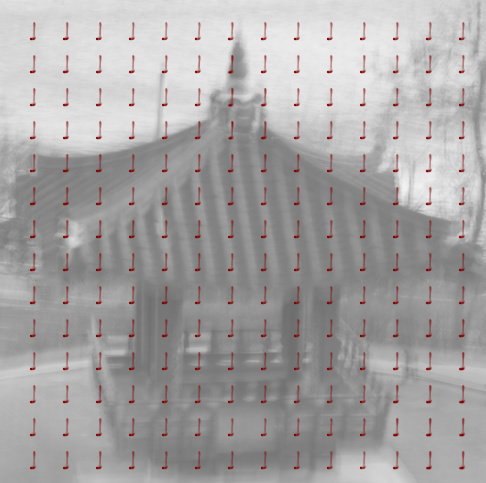} &
\includegraphics[height=0.115\textheight]{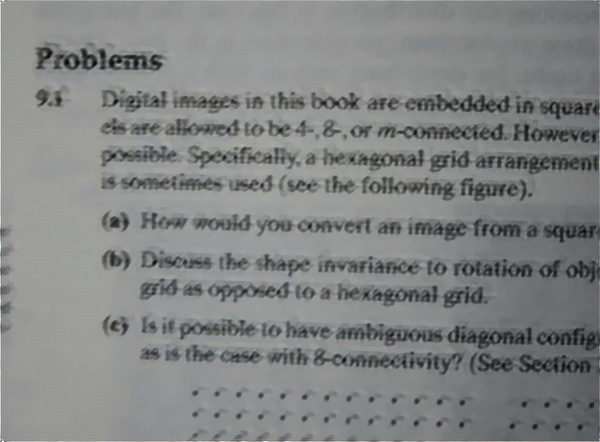} &
\includegraphics[height=0.115\textheight]{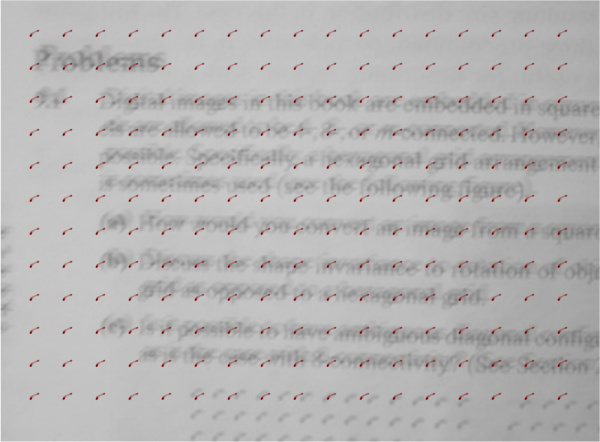} 
 \\
\multicolumn{8}{c}{Gong \etal~\cite{gong2017motion} reconstruction and kernels} \\
% Gong Results
\includegraphics[height=0.115\textheight]{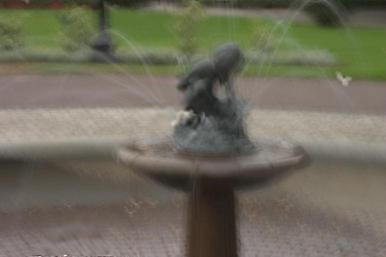} &
\includegraphics[height=0.115\textheight]{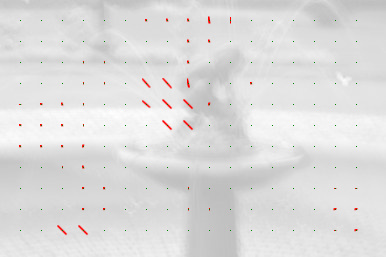} & 
\includegraphics[height=0.115\textheight]{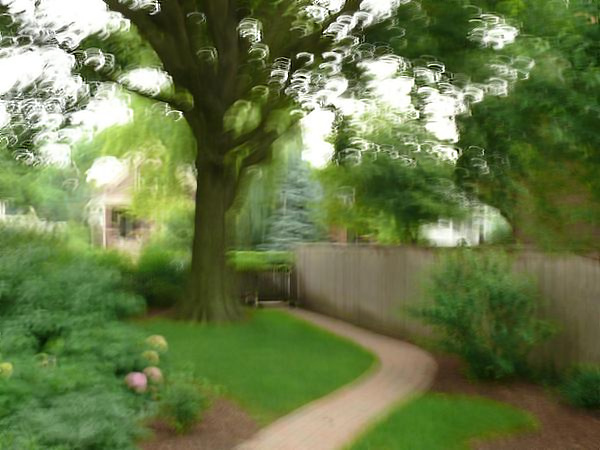}&
\includegraphics[height=0.115\textheight]{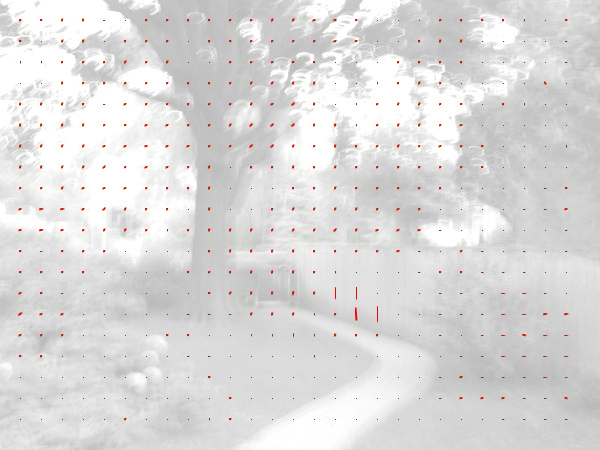}&
\includegraphics[height=0.115\textheight]{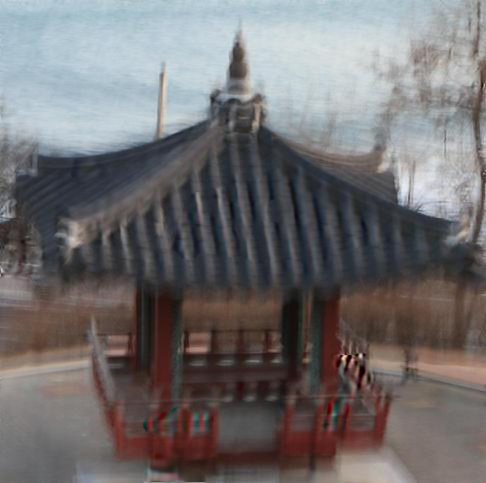}& 
\includegraphics[height=0.115\textheight]{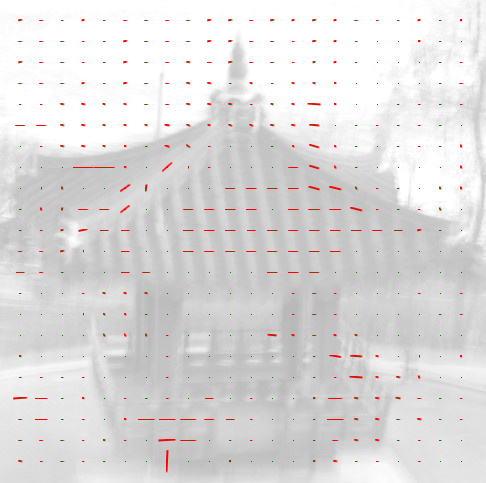}& 
\includegraphics[height=0.115\textheight]{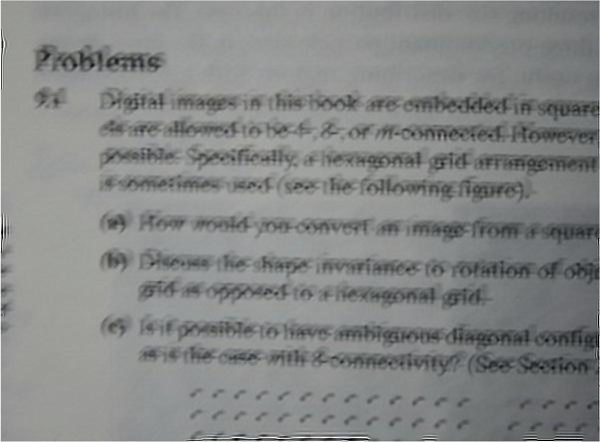} &
\includegraphics[height=0.115\textheight]{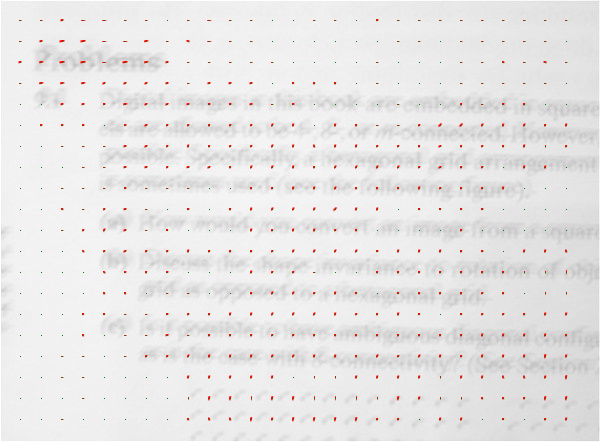} 
\\
\multicolumn{8}{c}{Sun \etal~\cite{sun2015learning} reconstruction and kernels} \\ % Sun reconstruction
% Sun Results
\includegraphics[height=0.115\textheight]{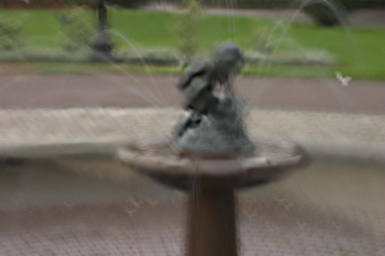} & 
\includegraphics[height=0.115\textheight]{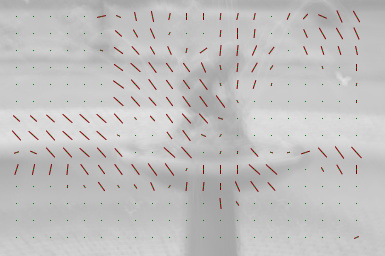} & 
\includegraphics[height=0.115\textheight]{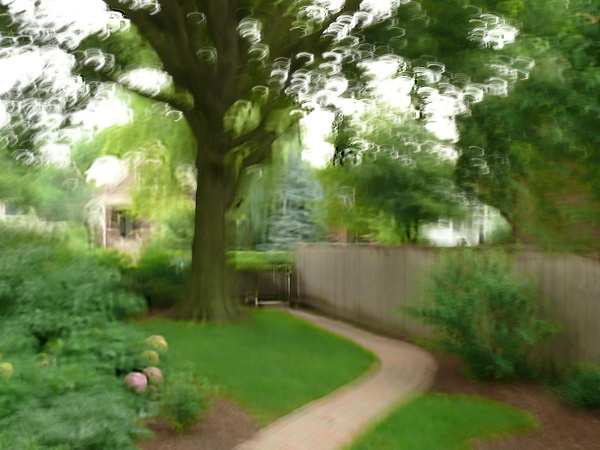}&
\includegraphics[height=0.115\textheight]{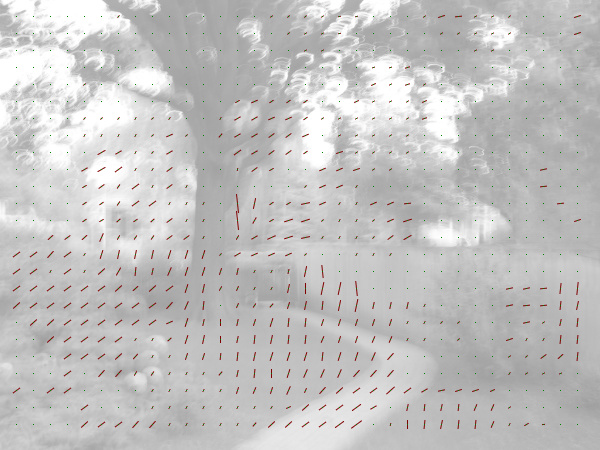}&
\includegraphics[height=0.115\textheight]{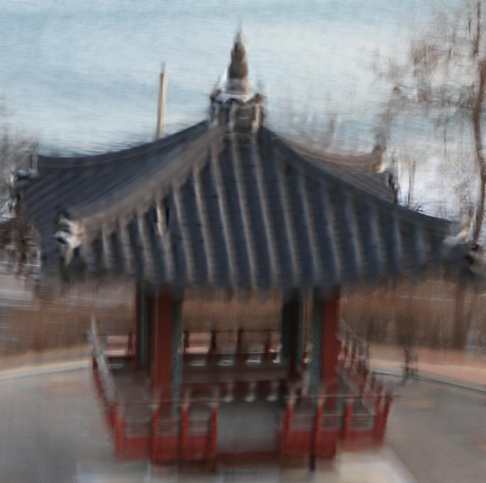}& 
\includegraphics[height=0.115\textheight]{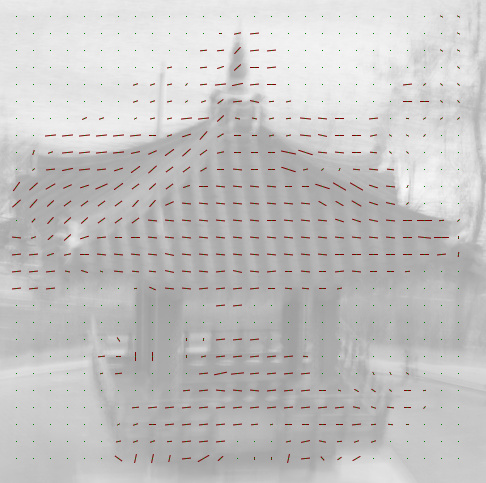}& 
\includegraphics[height=0.115\textheight]{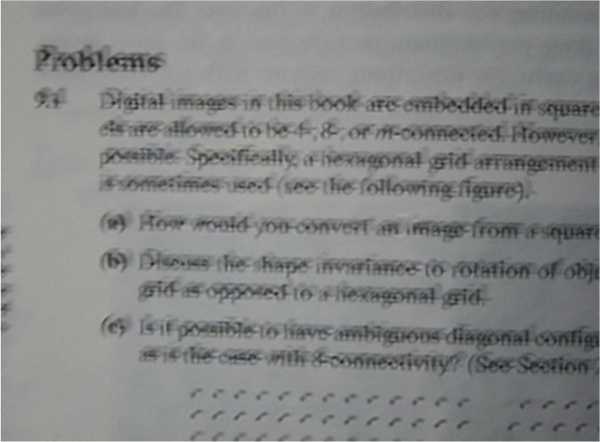} &
\includegraphics[height=0.115\textheight]{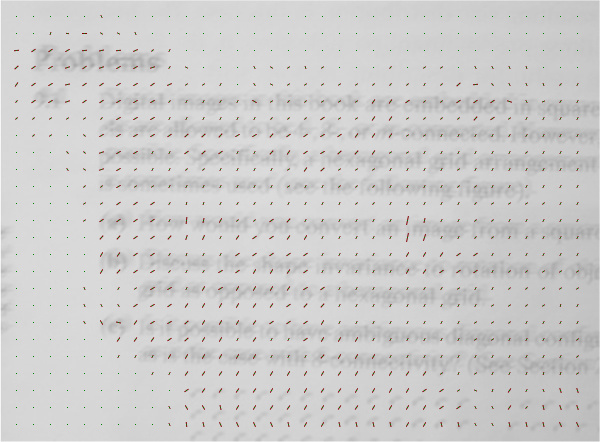} \\
\end{tabular}
\caption{Non-uniform kernel estimation and reconstruction examples for images of Lai's  dataset~\cite{lai2016comparative} ran at half resolution. 
%Blurry images (first row), kernels and reconstruction by our method (second and third row), by Gong \etal~\cite{gong2017motion} (fourth and fifth row) and by Sun \etal~\cite{sun2015learning} (sixth and seventh row)
}
\label{fig:comparisonLaiHalfSize2}
\end{figure}
\end{landscape}

\begin{landscape}
\begin{figure}[h]
\centering
\setlength{\tabcolsep}{2pt}
        \begin{tabular}{c|cccc|cccc}
        & \multicolumn{4}{c|}{  End-to-end } & \multicolumn{4}{c}{  Model-based} \\[.25em]
Original& D-GAN2~\cite{kupyn2019deblurgan} & SRN~\cite{tao2018scale} &  RealBlur~\cite{rim_2020_ECCV} & MPRNet~\cite{Zamir2021MPRNet} & Whyte ~\cite{whyte2010nonuniform} & Sun~\cite{sun2015learning}   & Gong~\cite{gong2017motion}  &  Ours  \\
 \includegraphics[height=0.13\textheight]{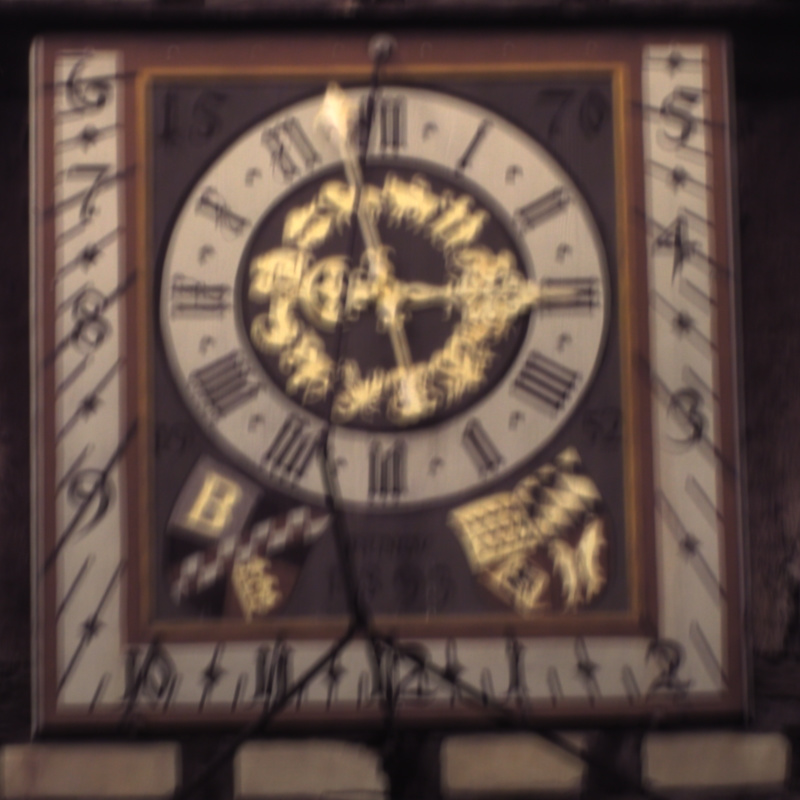} &
  \includegraphics[height=0.13\textheight]{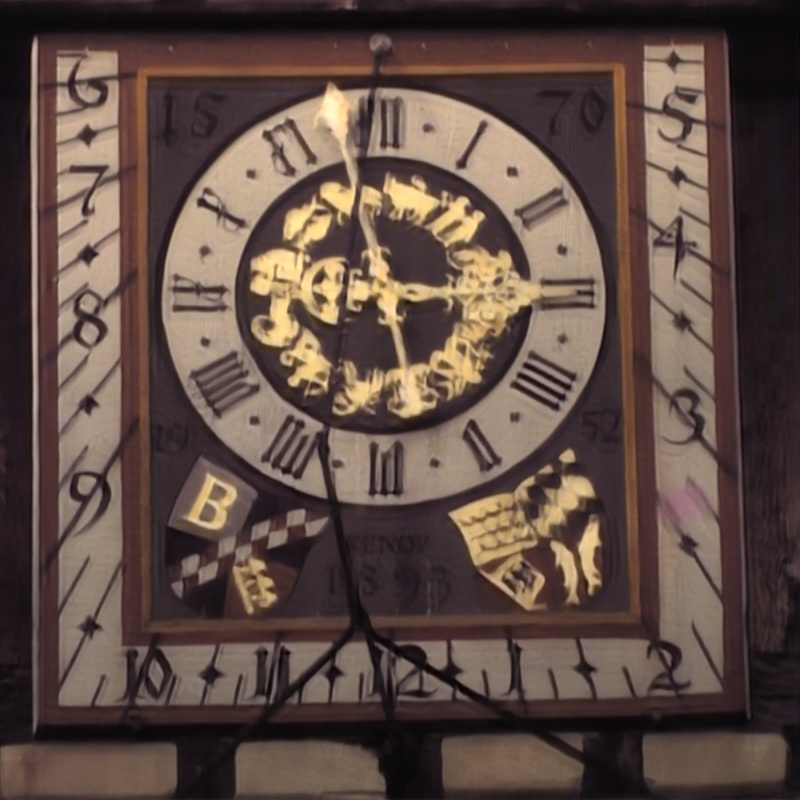} &   
 \includegraphics[height=0.13\textheight]{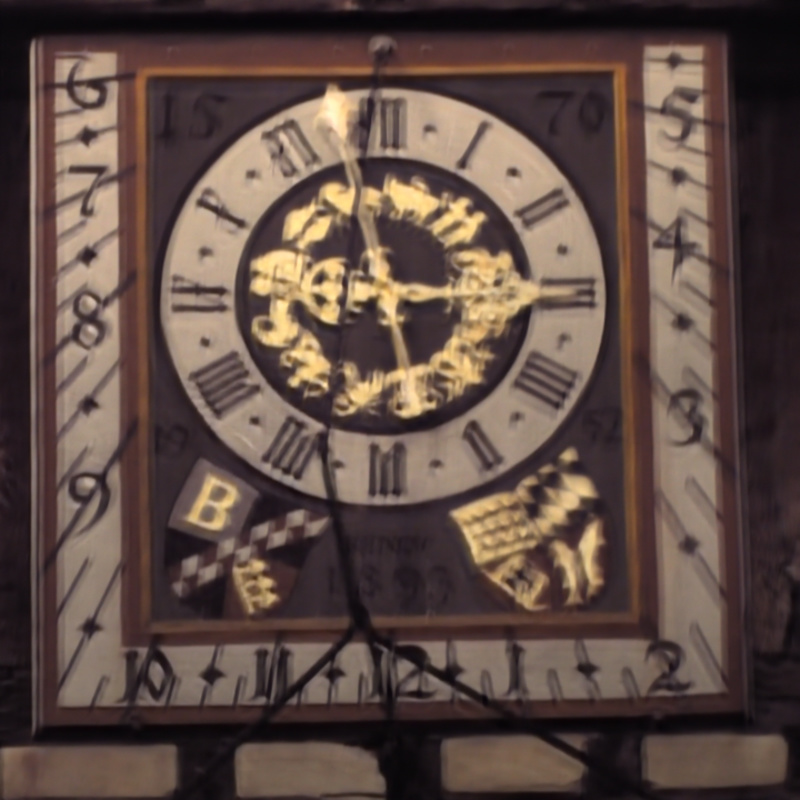} &
  \includegraphics[height=0.13\textheight]{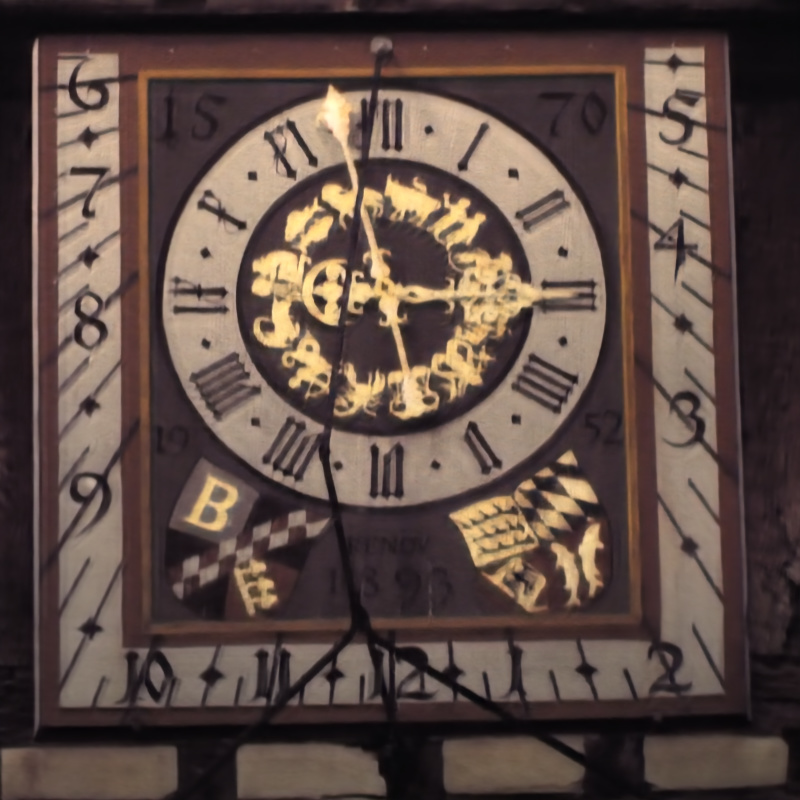} &
\includegraphics[height=0.13\textheight]{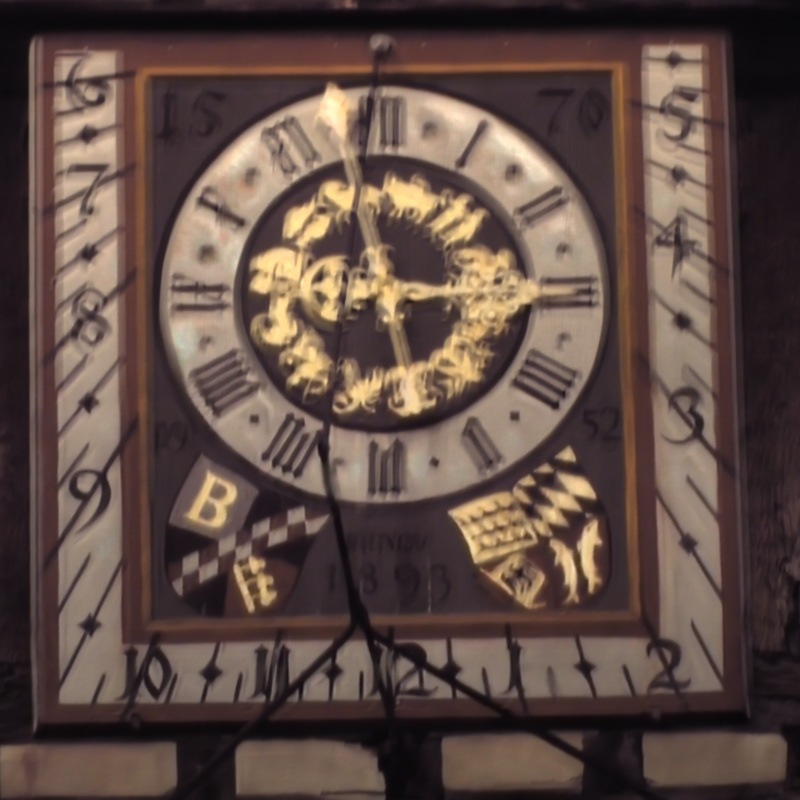}
  &
 
 \includegraphics[height=0.13\textheight]{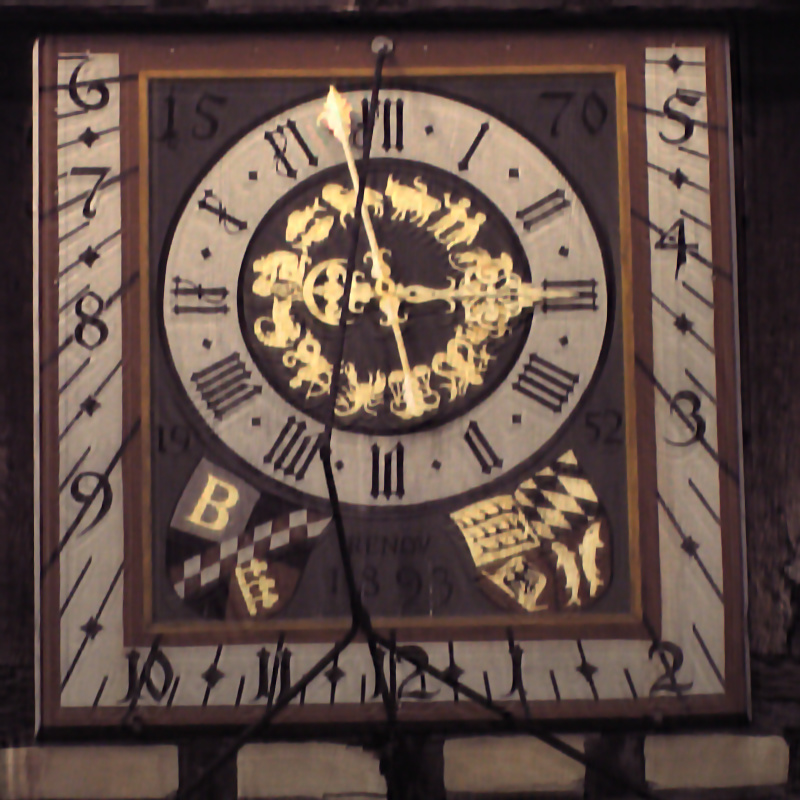} &
  \includegraphics[height=0.13\textheight]{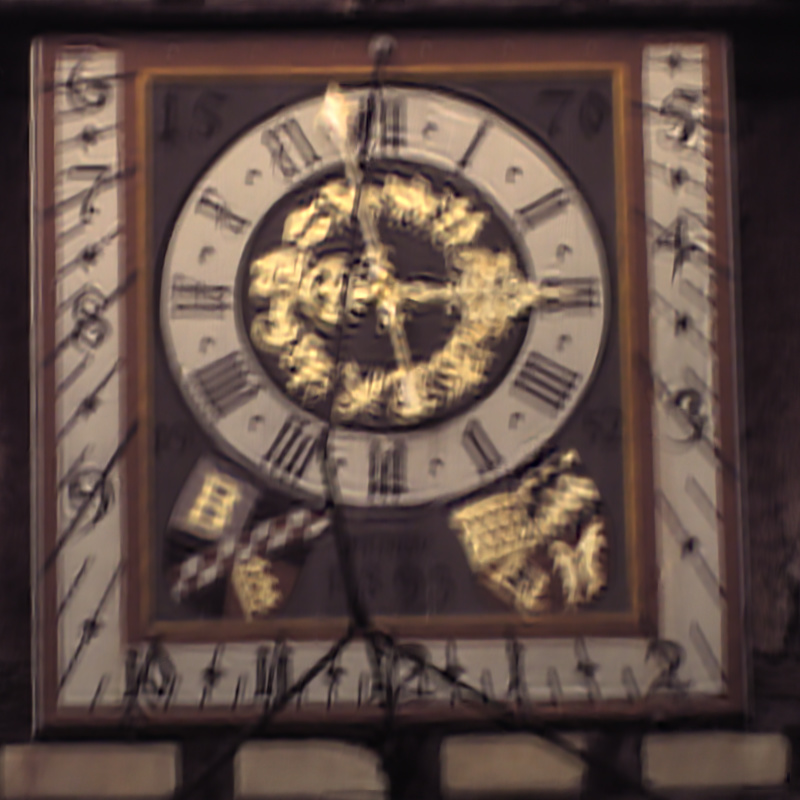} &
 \includegraphics[height=0.13\textheight]{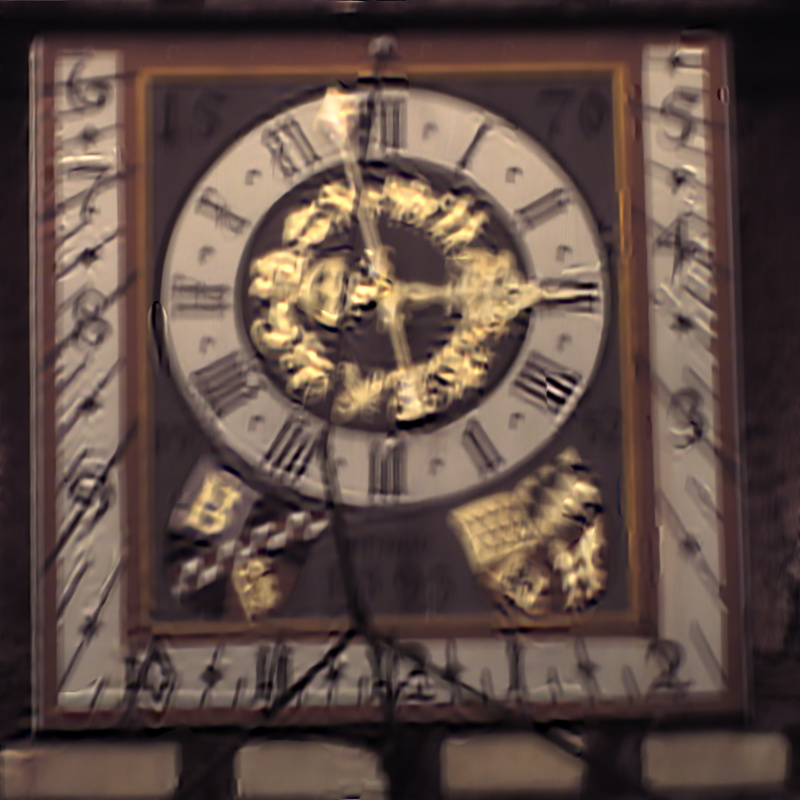} &
 \includegraphics[height=0.13\textheight]{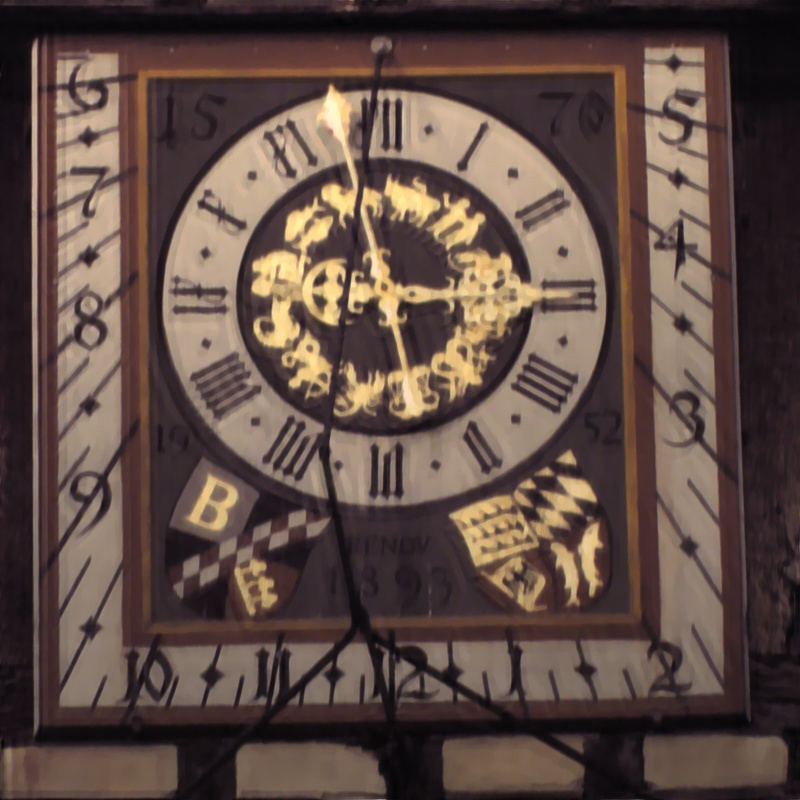} \\ 
 \includegraphics[trim=300 250 200 250, clip, height=0.13\textheight]{latex/Kohler/Blurry/Blurry2_6.jpg} &
  \includegraphics[trim=300 250 200 250, clip,height=0.13\textheight]{latex/Kohler/DeblurGANv2Inception/Blurry2_6.jpg} &  
   \includegraphics[trim=300 250 200 250, clip,height=0.13\textheight]{latex/Kohler/SRN/Blurry2_6.jpg} &
    \includegraphics[trim=300 250 200 250, clip,height=0.13\textheight]{latex/Kohler/RealBlurJ_pre_trained+GOPRO+BSD500/Blurry2_6.jpg} &
 \includegraphics[trim=300 250 200 250, clip,height=0.13\textheight]{latex/Kohler/MPRNet/Blurry2_6.jpg} &
 \includegraphics[trim=300 250 200 250, clip,height=0.13\textheight]{latex/Kohler/Whyte/Blurry2_6_result_whyteMAP-krishnan.jpg}&
 \includegraphics[trim=300 250 200 250, clip,height=0.13\textheight]{latex/Kohler/Sun/Blurry2_6.jpg} &
 \includegraphics[trim=300 250 200 250, clip,height=0.13\textheight]{latex/Kohler/gong/Blurry2_6_deblurred.jpg} &
 \includegraphics[trim=300 250 200 250, clip,height=0.13\textheight]{latex/Kohler/our_no_FC/Blurry2_6_restored.jpg} \\ 
 21.89 dB & 26.54  dB& 24.73 dB &  28.11 dB  & 24.08 dB & 27.06 dB & 22.31 dB & 21.87 dB   & 28.20 dB \\
 \includegraphics[width=0.13\textwidth]{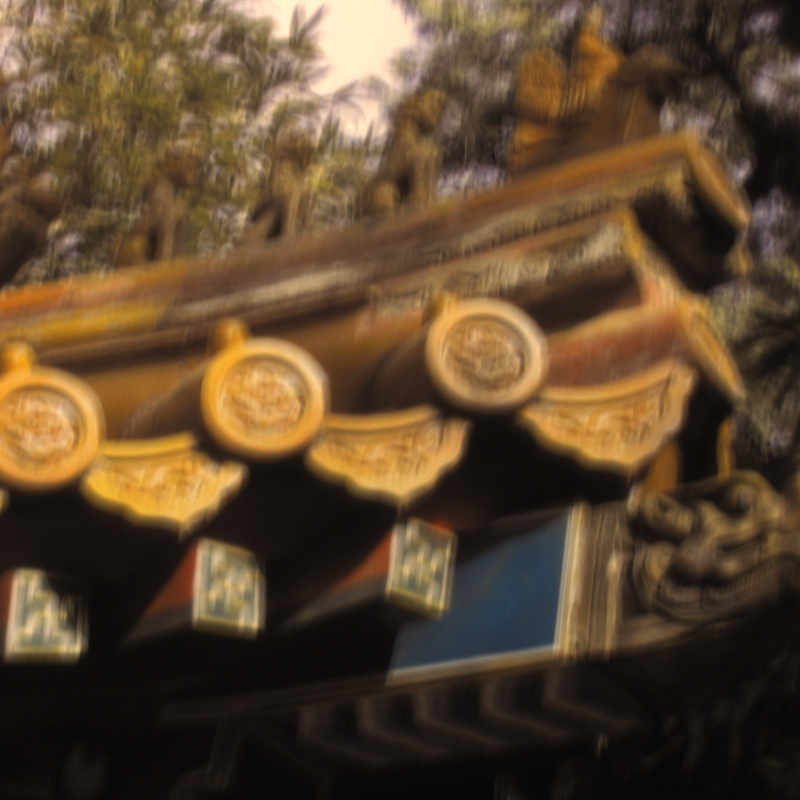} &
 \includegraphics[width=0.13\textwidth]{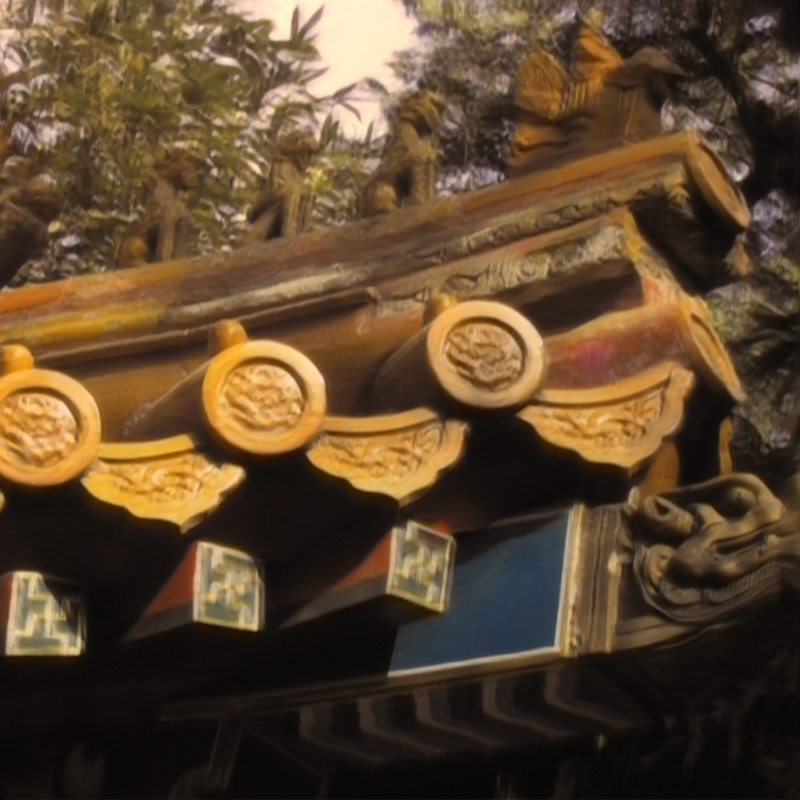} & 
 \includegraphics[width=0.13\textwidth]{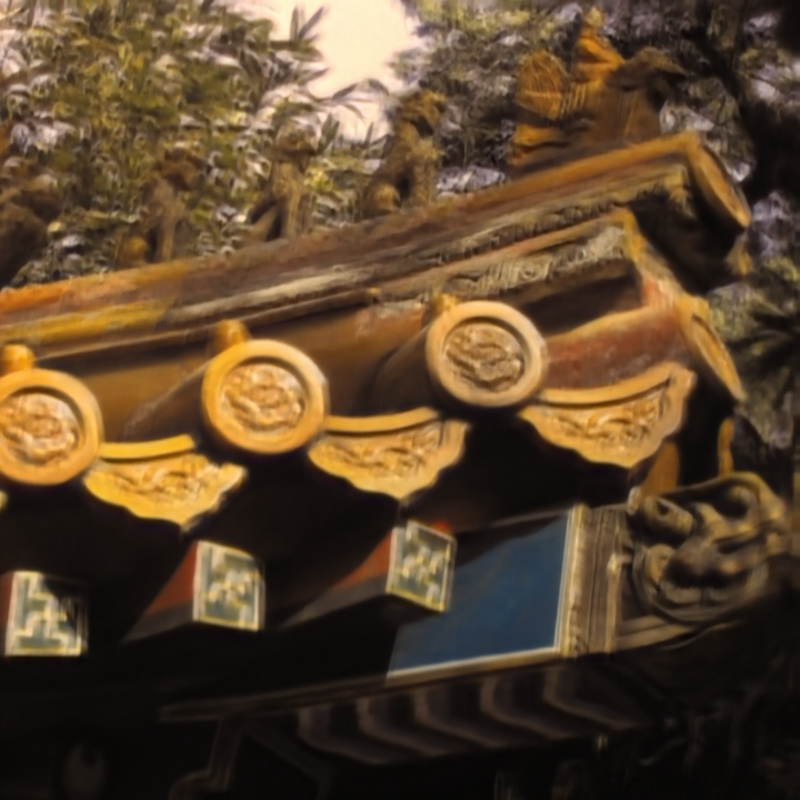} &
 \includegraphics[width=0.13\textwidth]{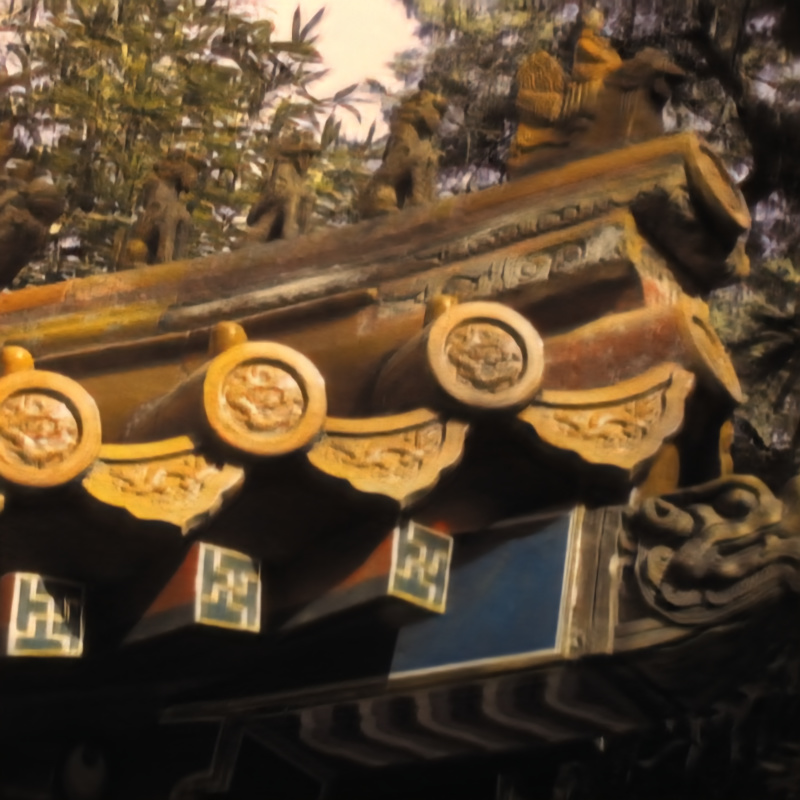} &
  \includegraphics[height=0.13\textheight]{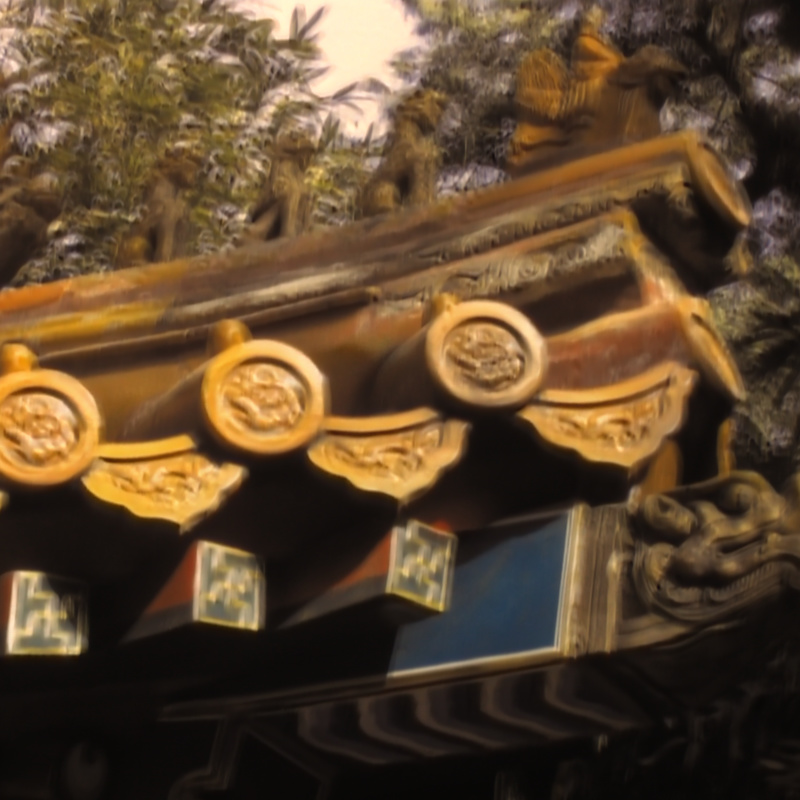} &
\includegraphics[width=0.13\textwidth]{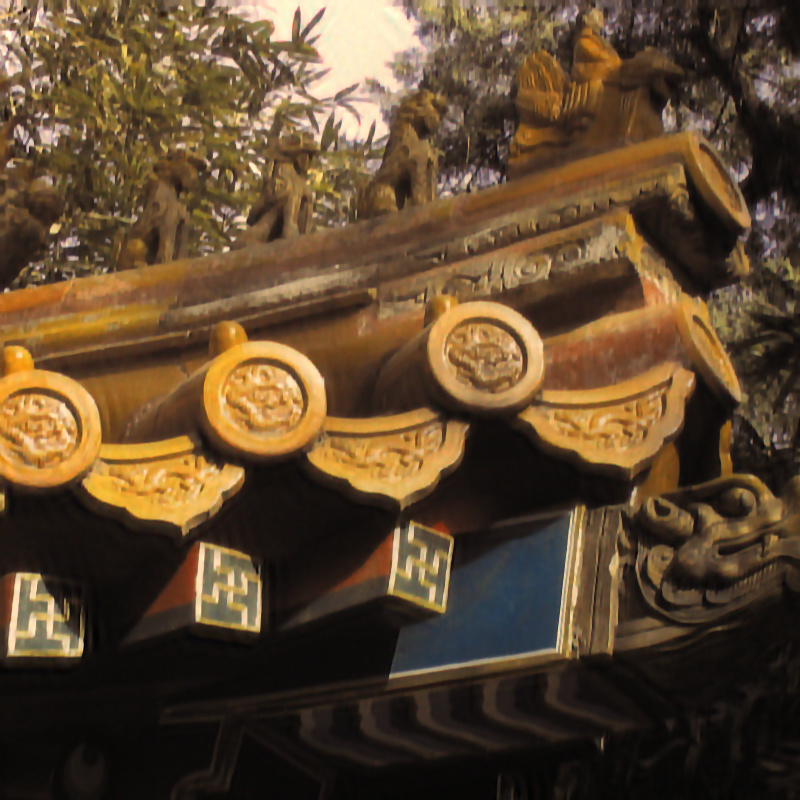} &
\includegraphics[width=0.13\textwidth]{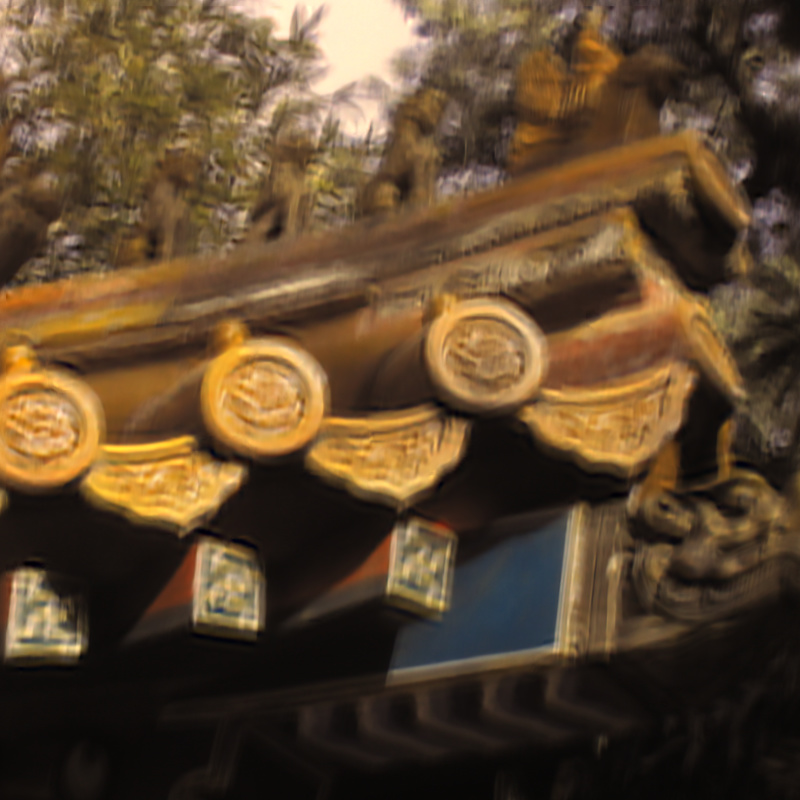}
&
\includegraphics[width=0.13\textwidth]{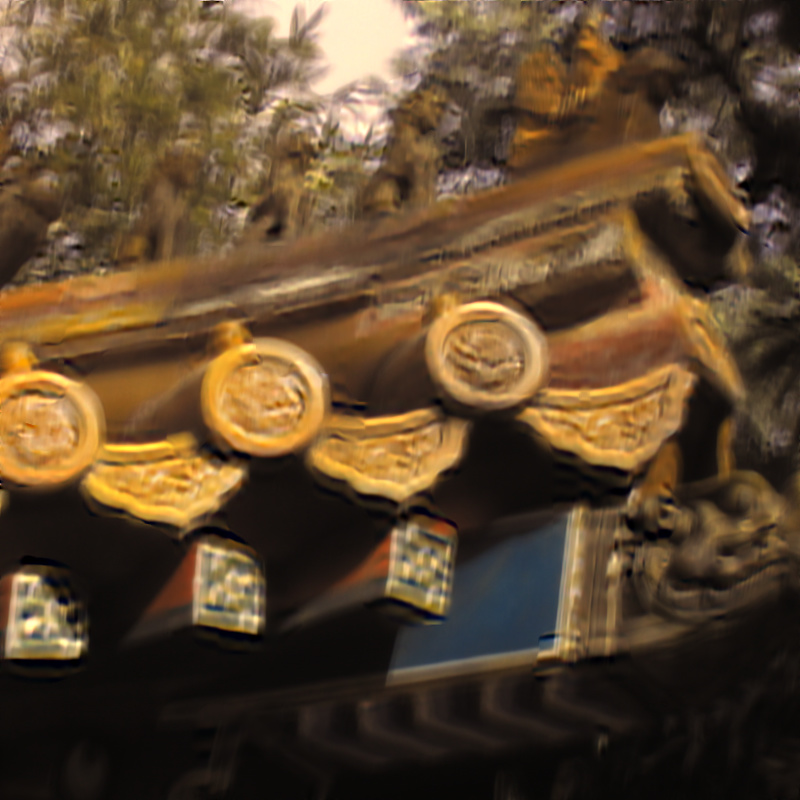} &
\includegraphics[width=0.13\textwidth]{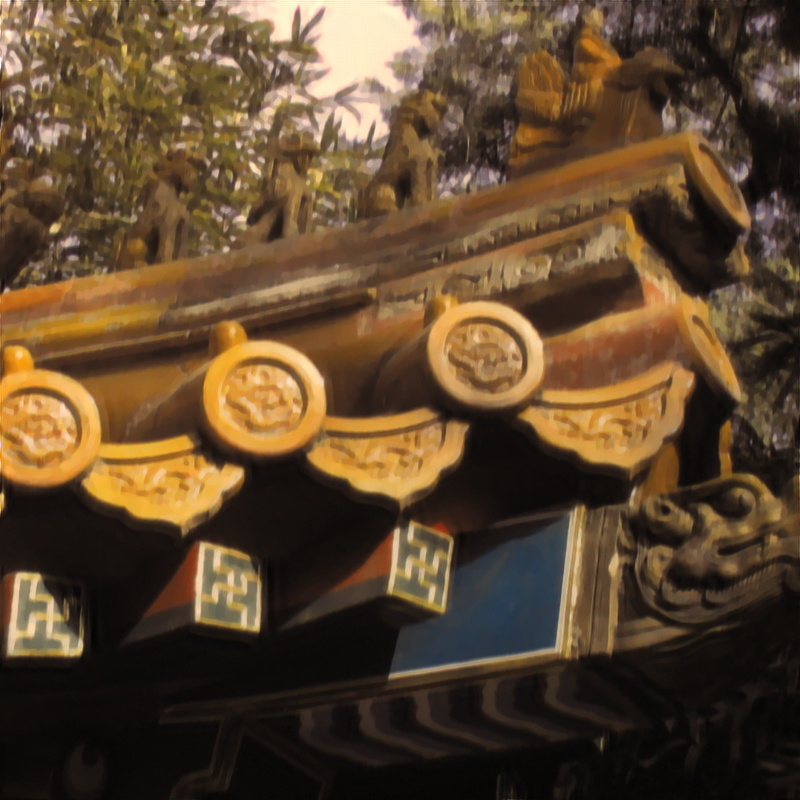}
\\ 
\includegraphics[trim=300 200 200 250, clip, width=0.13\textwidth]{latex/Kohler/Blurry/Blurry4_6.jpg}  &
\includegraphics[trim=300 200 200 250, clip,width=0.13\textwidth]{latex/Kohler/DeblurGANv2Inception/Blurry4_6.jpg} &  
\includegraphics[trim=300 200 200 250, clip,width=0.13\textwidth]{latex/Kohler/SRN/Blurry4_6.jpg} &
\includegraphics[trim=300 200 200 250, clip,width=0.13\textwidth]{latex/Kohler/RealBlurJ_pre_trained+GOPRO+BSD500/Blurry4_6.jpg} &
\includegraphics[trim=300 200 200 250, clip,width=0.13\textheight]{latex/Kohler/MPRNet/Blurry4_6.jpg}
& 
\includegraphics[trim=300 200 200 250, clip,width=0.13\textwidth]{latex/Kohler/Whyte/Blurry4_6_result_whyteMAP-krishnan.jpg}&
\includegraphics[trim=300 200 200 250, clip,width=0.13\textwidth]{latex/Kohler/Sun/Blurry4_6.jpg}&
\includegraphics[trim=300 200 200 250, clip,width=0.13\textwidth]{latex/Kohler/gong/Blurry4_6_deblurred.jpg} &
\includegraphics[trim=300 200 200 250, clip,width=0.13\textwidth]{latex/Kohler/our_no_FC/Blurry4_6_restored.jpg}
\\ 
23.08 dB & 25.94 dB   & 25.14 dB & 28.04 dB & 24.78 dB &   29.57 dB & 23.41 dB  &  23.08 dB & 29.00 dB \\

    \end{tabular}
    \caption{Qualitative comparison of different deblurring methods on K\"{o}hler's Dataset~\cite{kohler2012recording}.}
    \label{fig:comparisonKohler}
\end{figure}
\end{landscape}

%%%%%%%%%%%%%%%%%%%%%%%%%%%%%%%%%%%%%%%%%%%%%%%%%%%%%%%%%%%%%%%%%%%%%%%
\begin{landscape}

\begin{figure}
\centering
\begin{tabular}{cc|cc|cc|cc|cc}
\multicolumn{10}{c}{Input blurry images} \\
\includegraphics[width=0.12\textwidth]{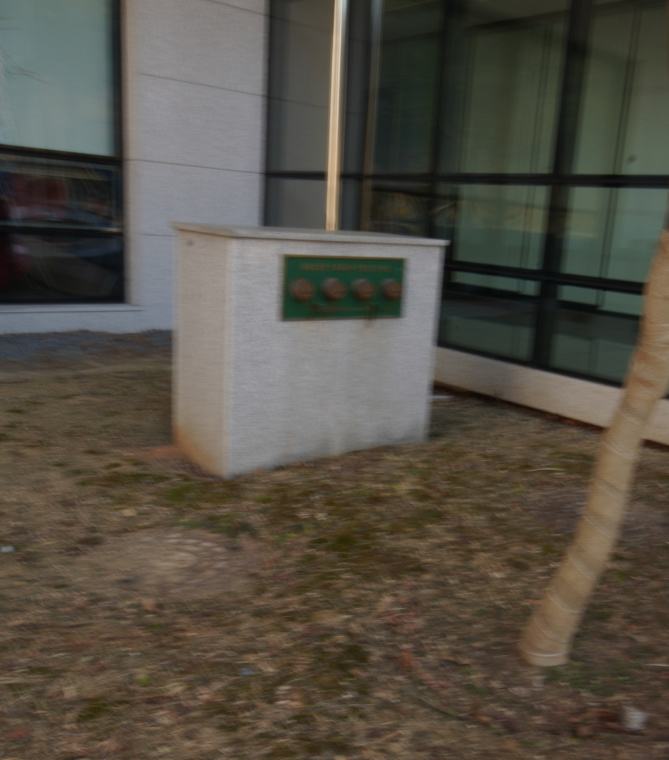} & &
\includegraphics[width=0.12\textwidth]{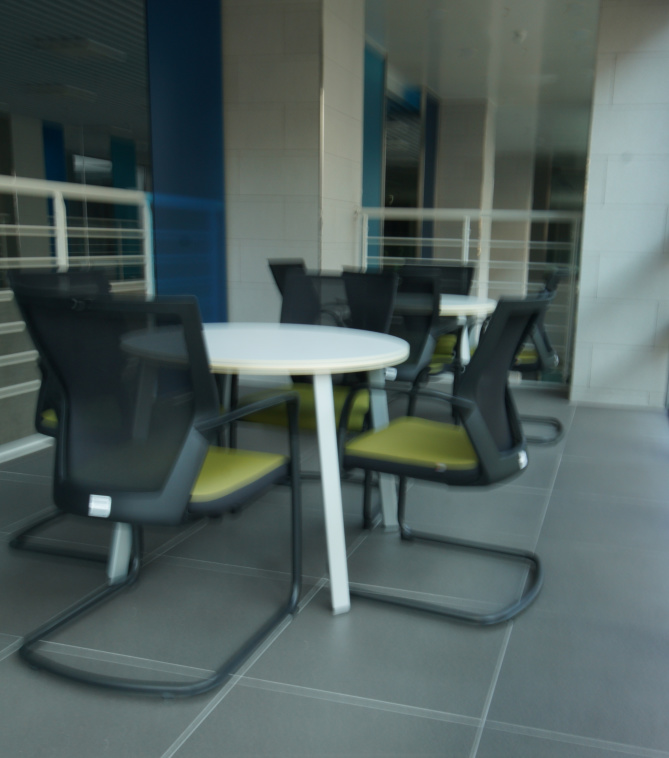}  & &
\includegraphics[width=0.12\textwidth]{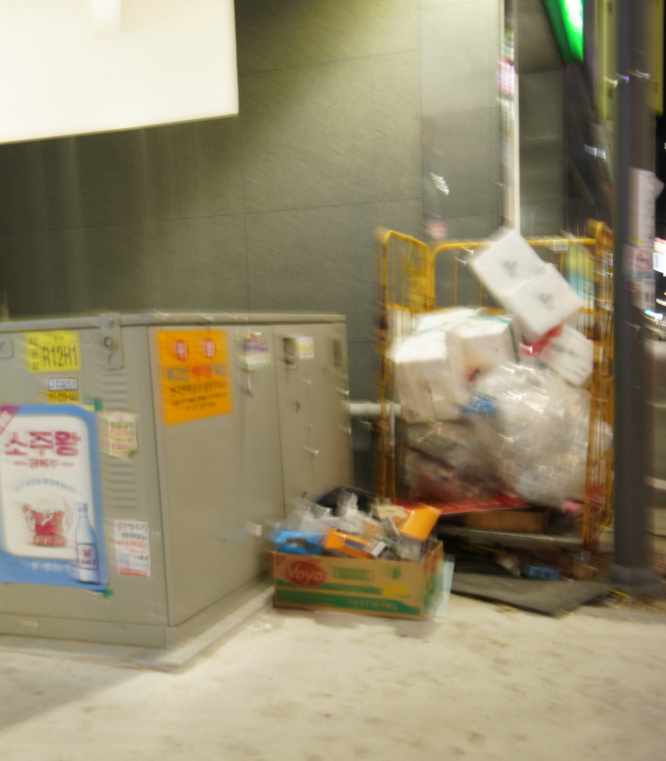} &  &
\includegraphics[width=0.12\textwidth]{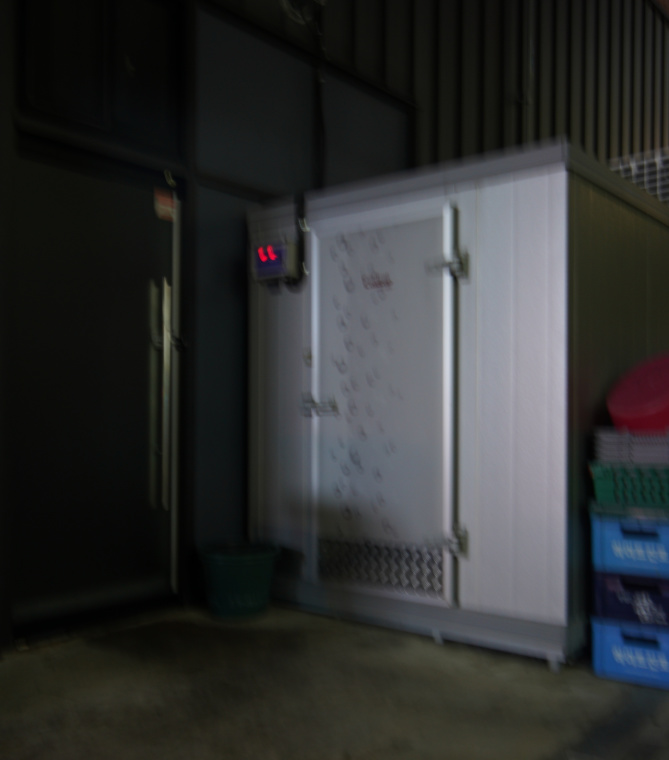} & &
\includegraphics[width=0.12\textwidth]{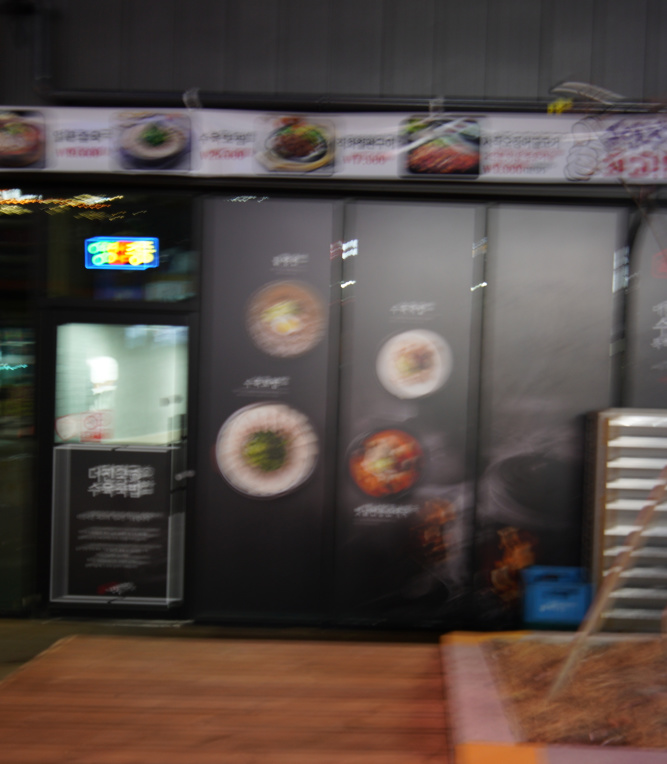} & \\
\multicolumn{10}{c}{Our reconstructions and kernels} \\
% Ours kernels
\includegraphics[width=0.12\textwidth]{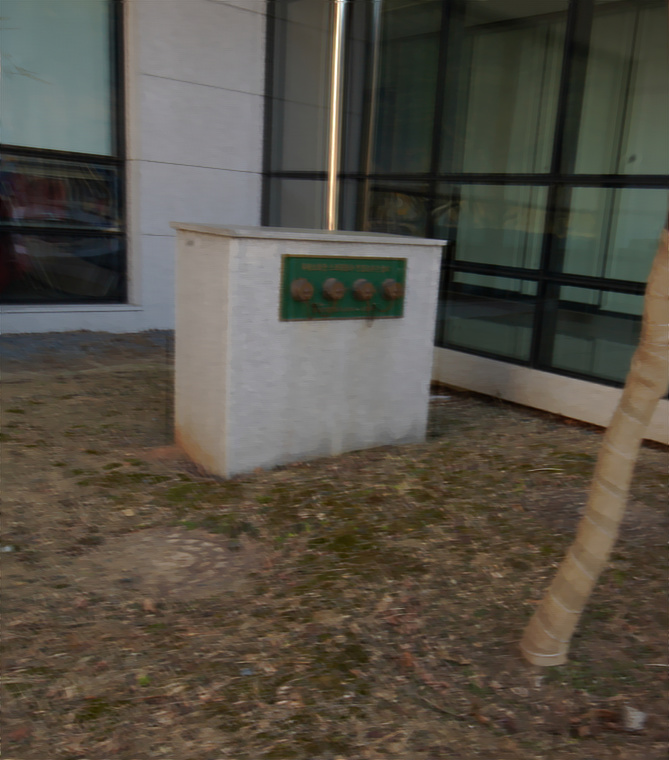} &
\includegraphics[width=0.12\textwidth]{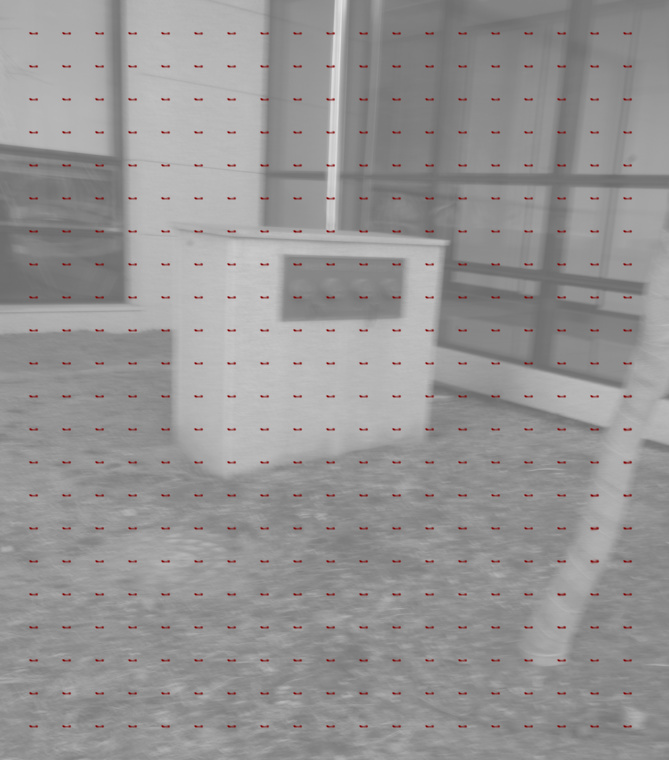} &
\includegraphics[width=0.12\textwidth]{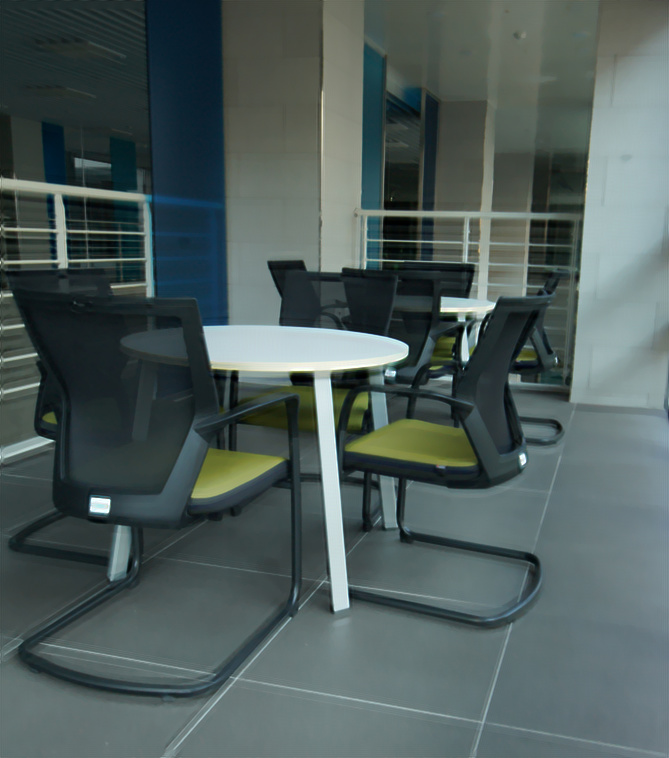} &
\includegraphics[width=0.12\textwidth]{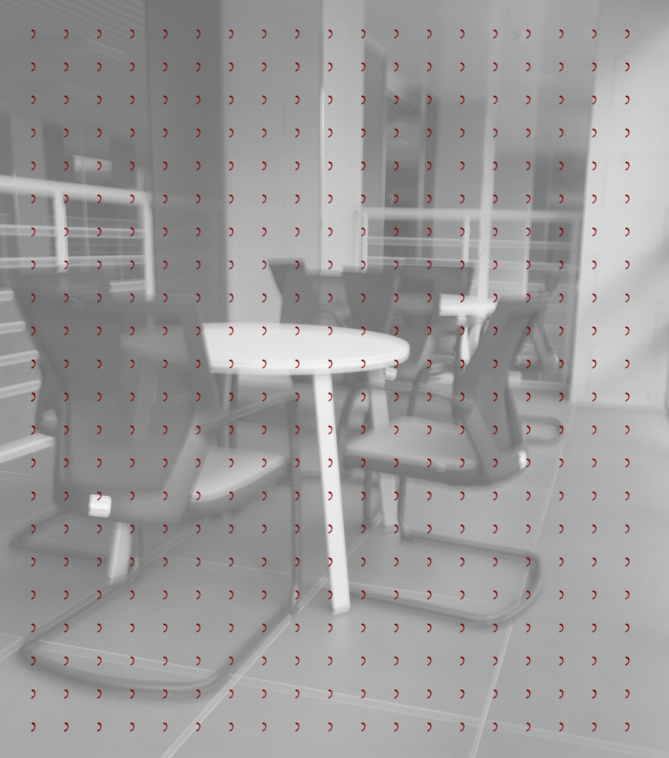} &
\includegraphics[width=0.12\textwidth]{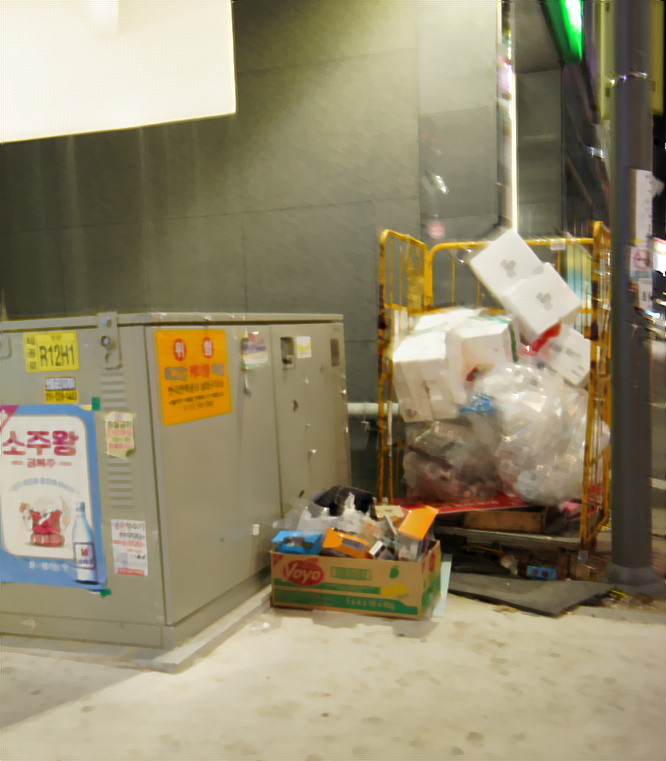}  & 
\includegraphics[width=0.12\textwidth]{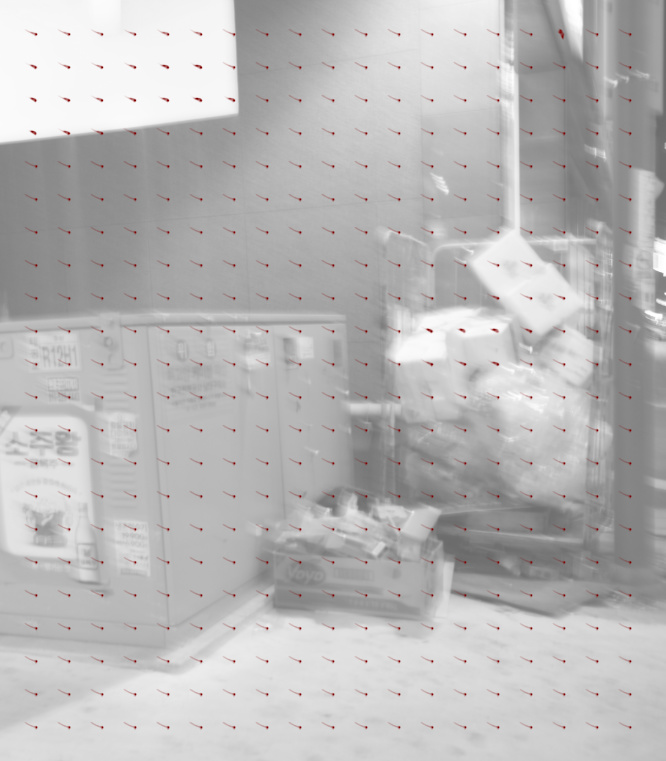} & 
\includegraphics[width=0.12\textwidth]{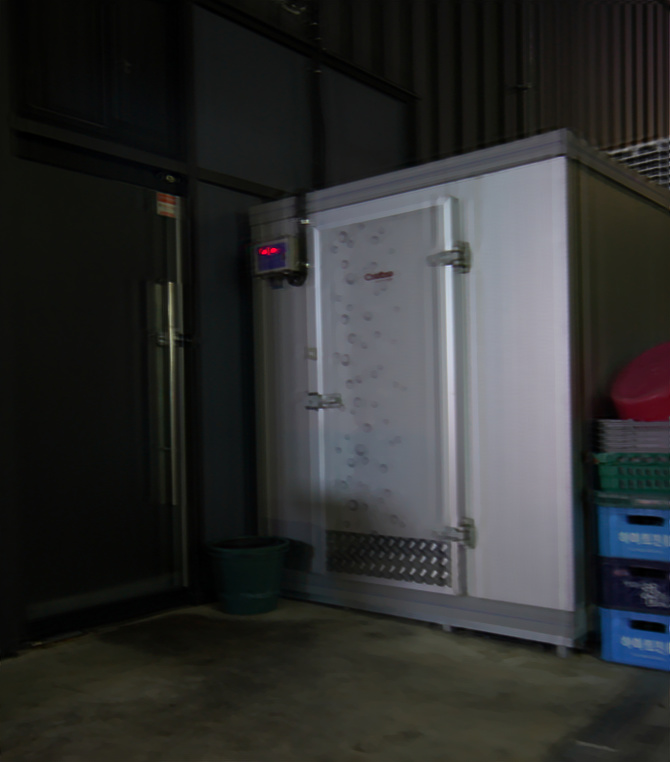}  & 
\includegraphics[width=0.12\textwidth]{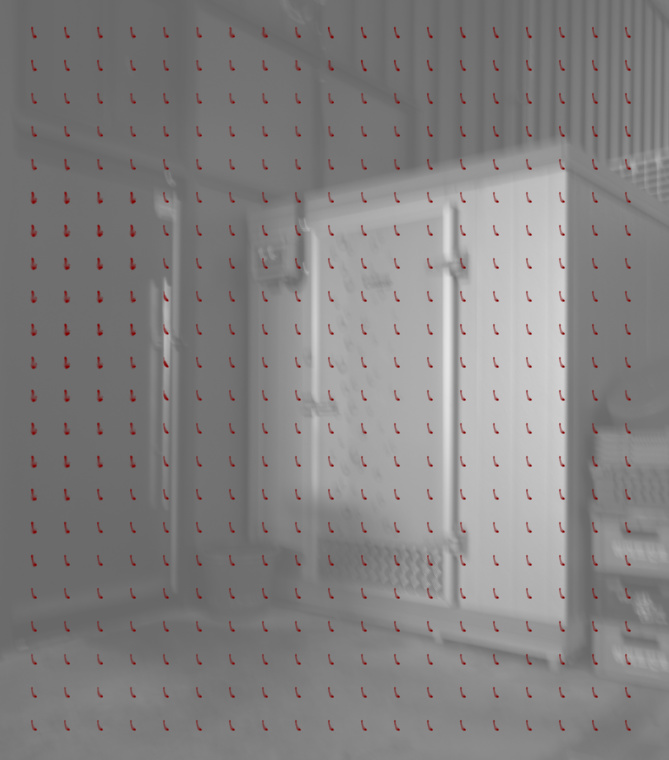} & 
\includegraphics[width=0.12\textwidth]{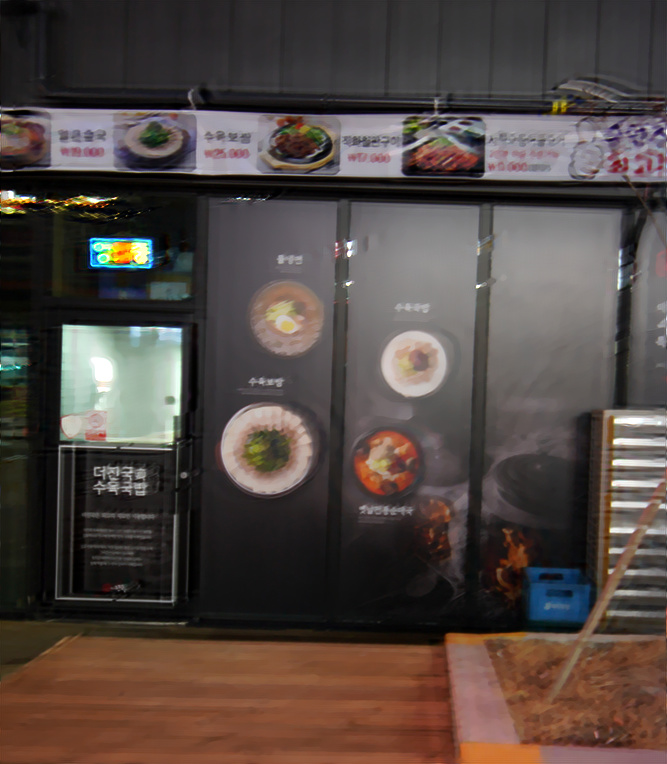} &
\includegraphics[width=0.12\textwidth]{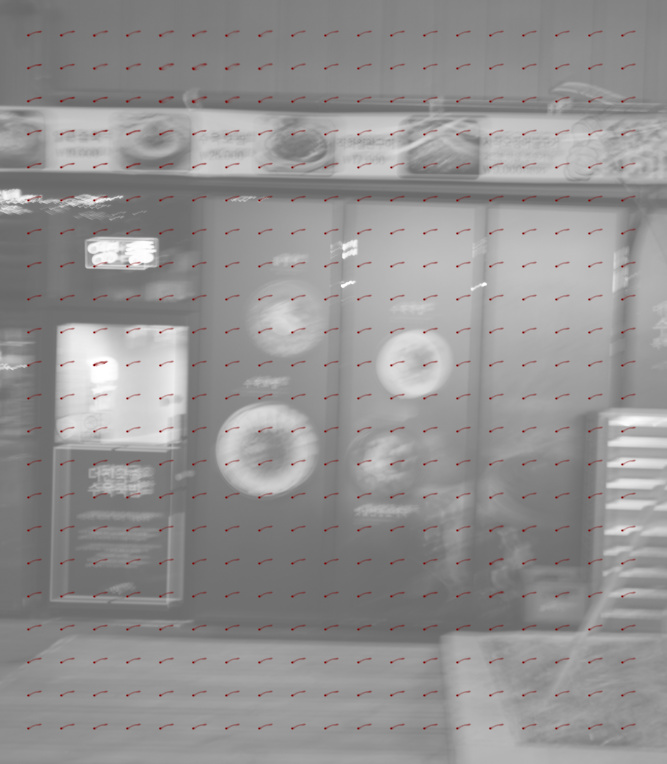} \\
\multicolumn{10}{c}{Gong \etal~\cite{gong2017motion} and reconstructions kernels} \\
\includegraphics[width=0.12\textwidth]{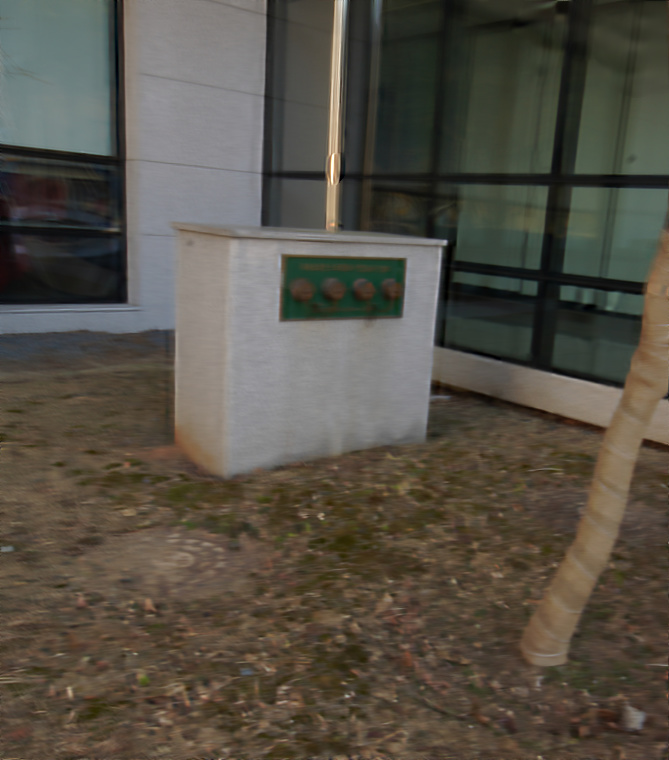} &
\includegraphics[width=0.12\textwidth]{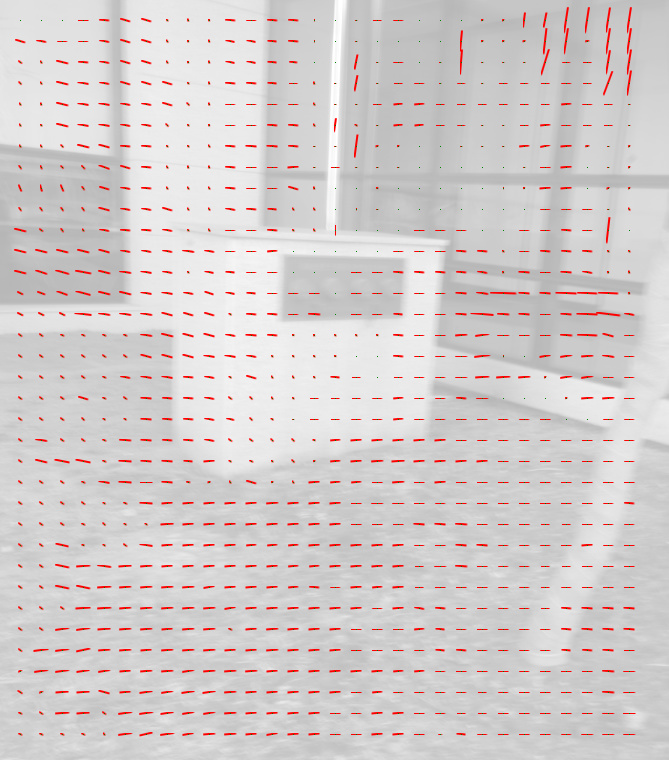}&
\includegraphics[width=0.12\textwidth]{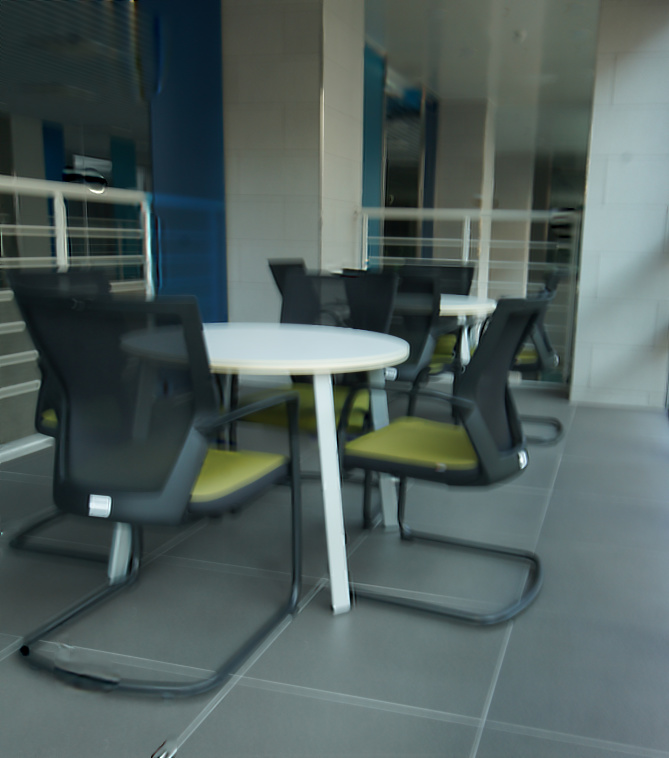} &
\includegraphics[width=0.12\textwidth]{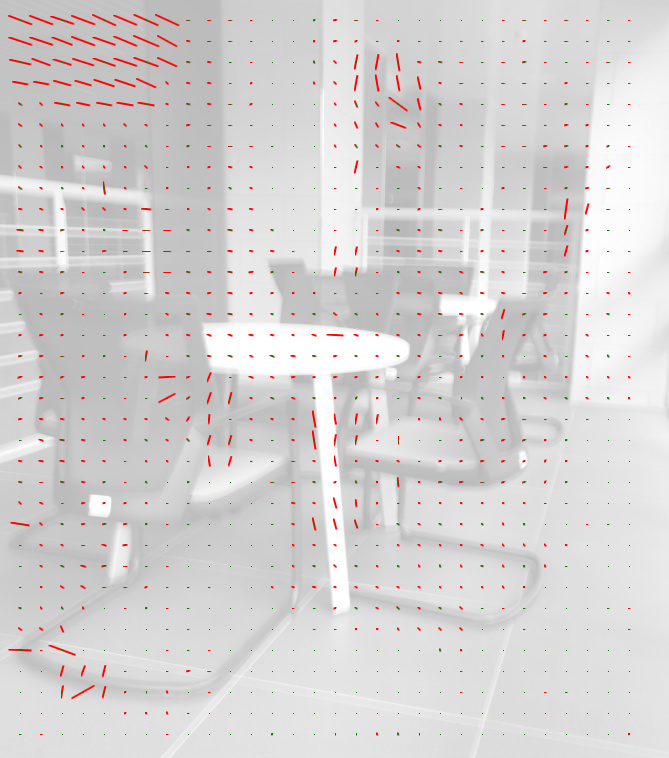} &
\includegraphics[width=0.12\textwidth]{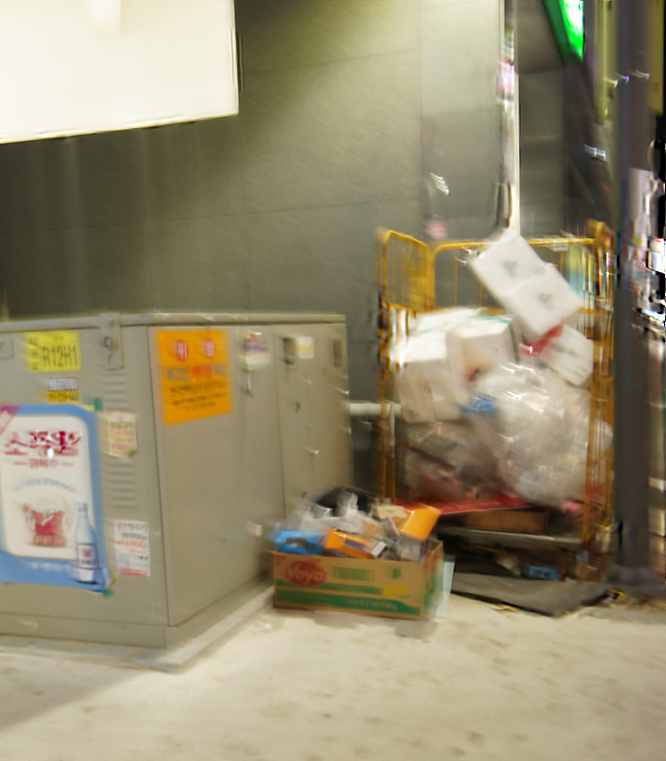} &
\includegraphics[width=0.12\textwidth]{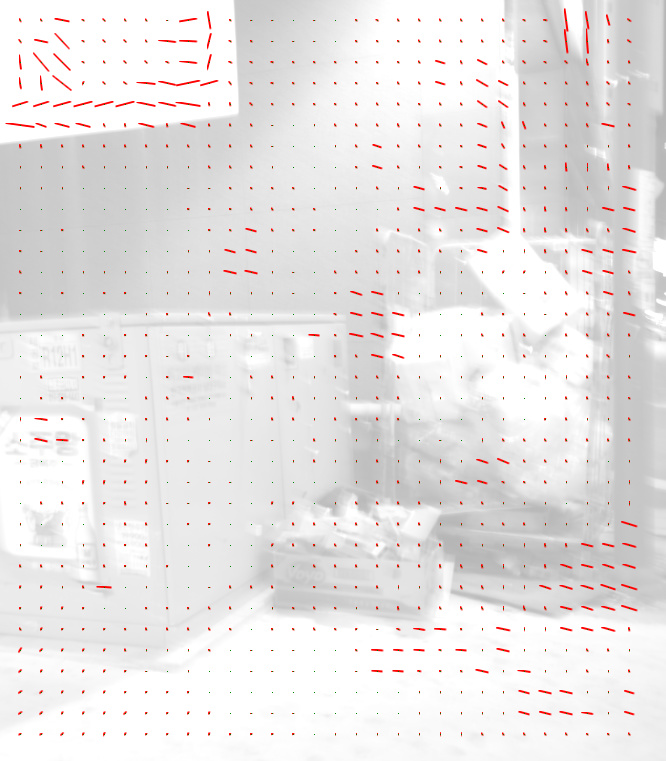} & 
\includegraphics[width=0.12\textwidth]{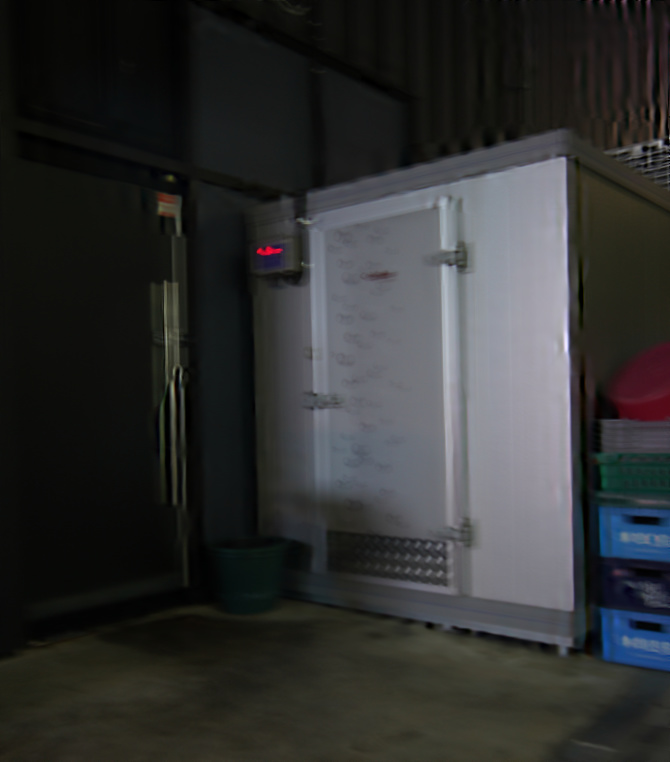} &
\includegraphics[width=0.12\textwidth]{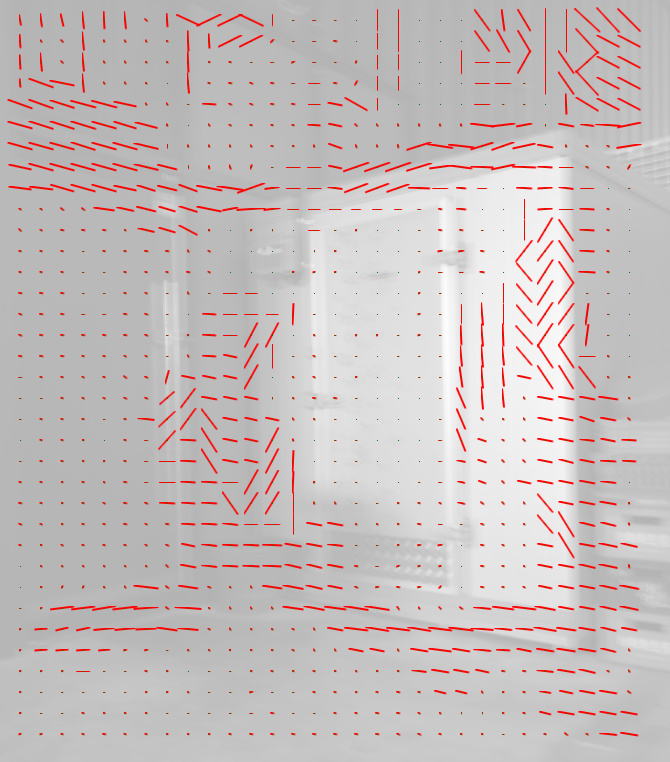} & 

\includegraphics[width=0.12\textwidth]{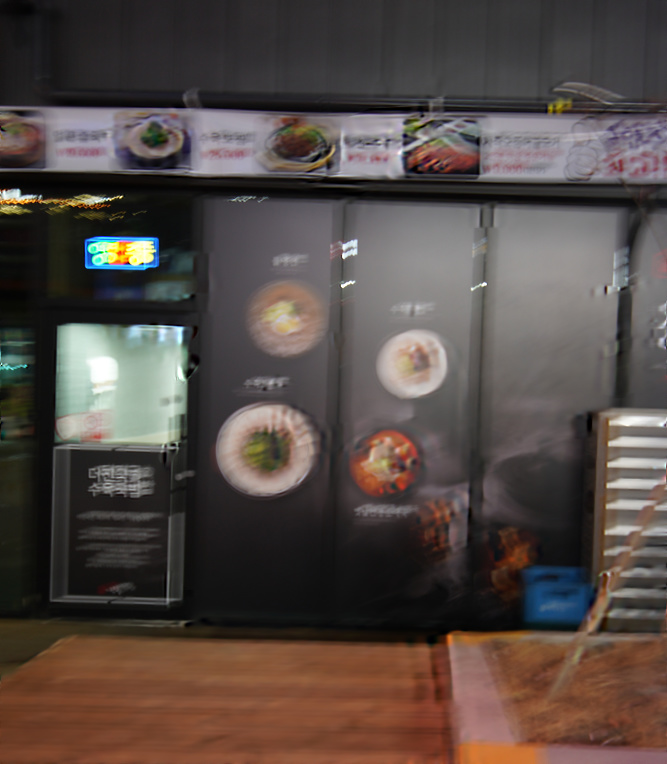} &
\includegraphics[width=0.12\textwidth]{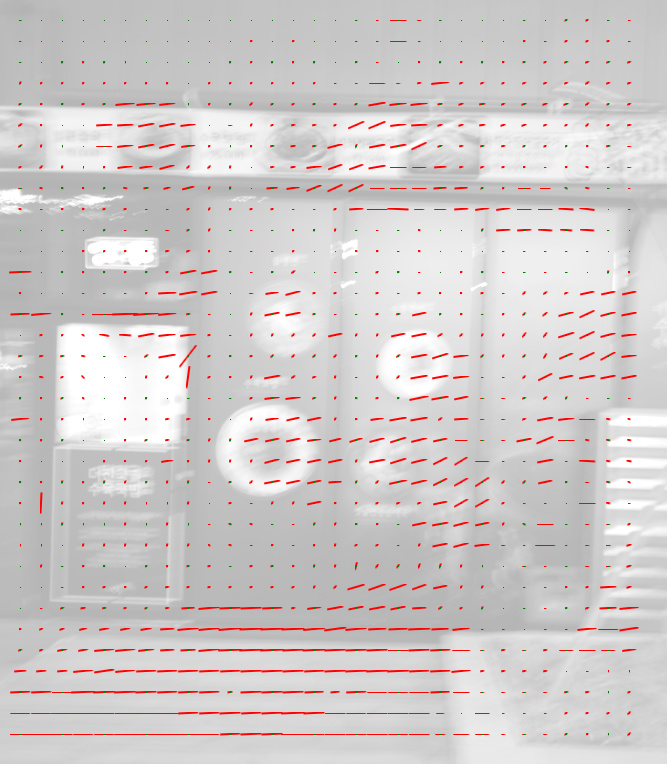}  
 \\  % Sun kernels
\multicolumn{10}{c}{Sun \etal~\cite{sun2015learning} reconstructions and Kernels} \\
\includegraphics[width=0.12\textwidth]{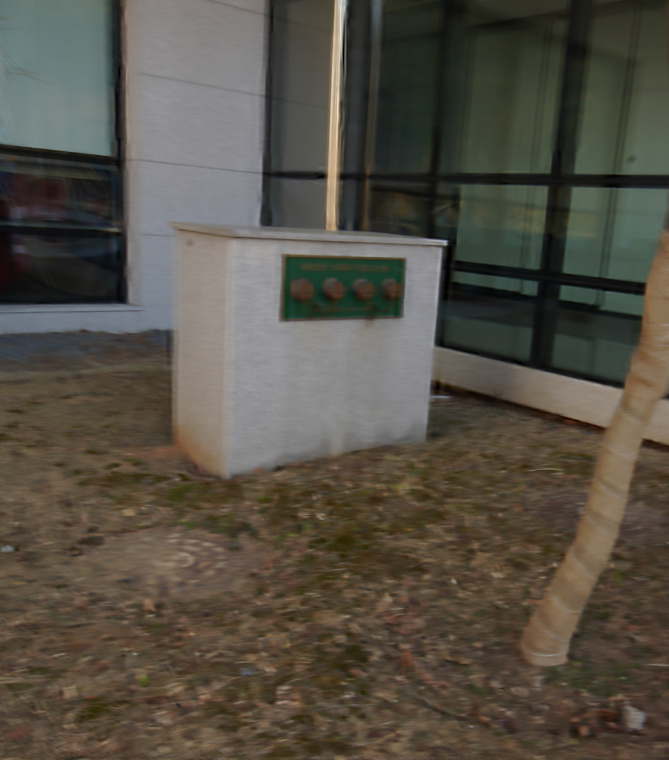} &
\includegraphics[width=0.12\textwidth]{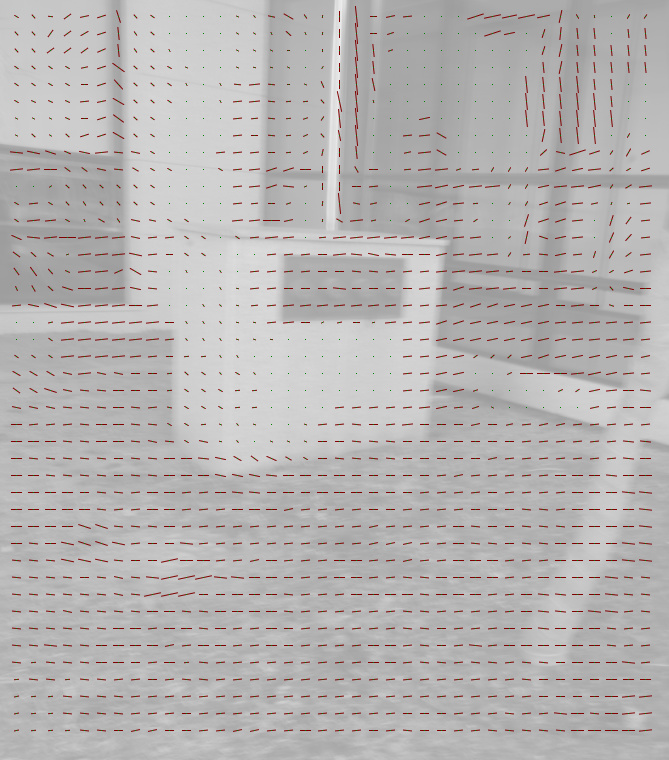} &
\includegraphics[width=0.12\textwidth]{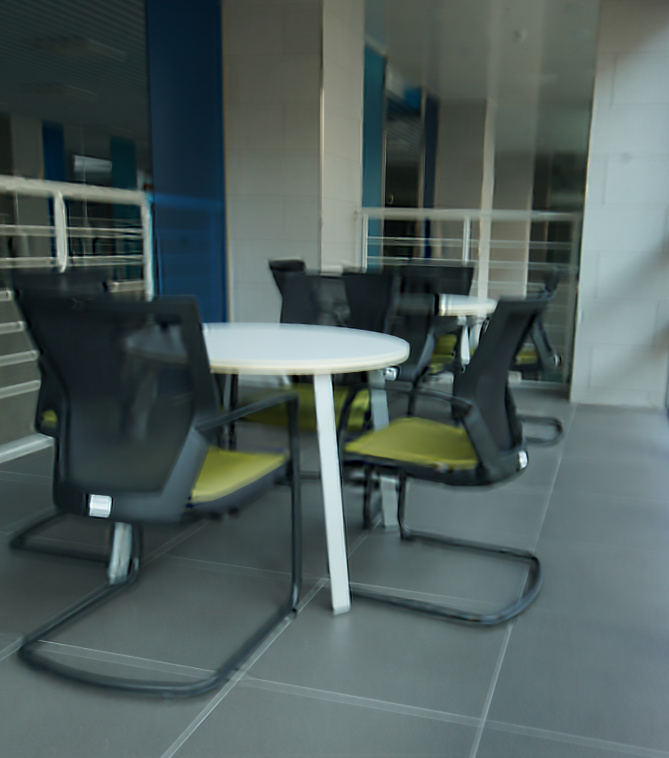} &
\includegraphics[width=0.12\textwidth]{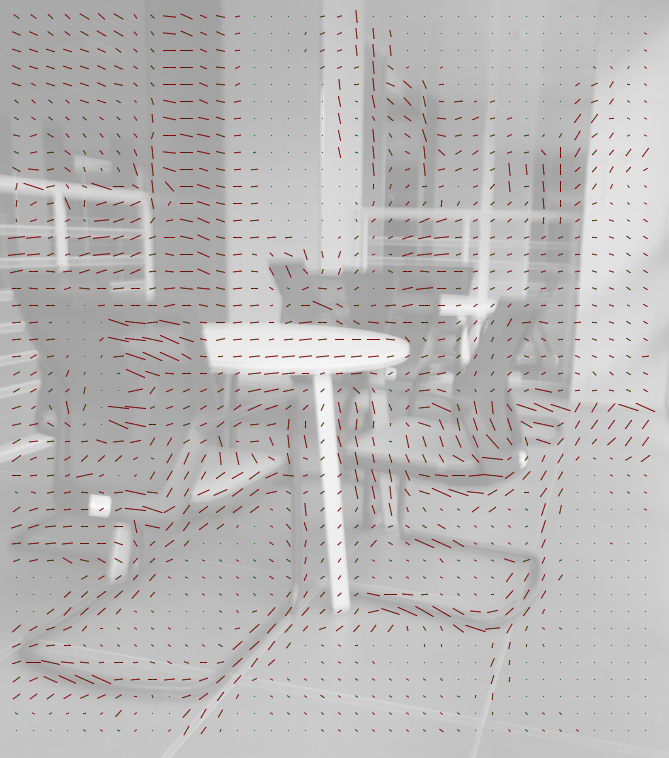} &
\includegraphics[width=0.12\textwidth]{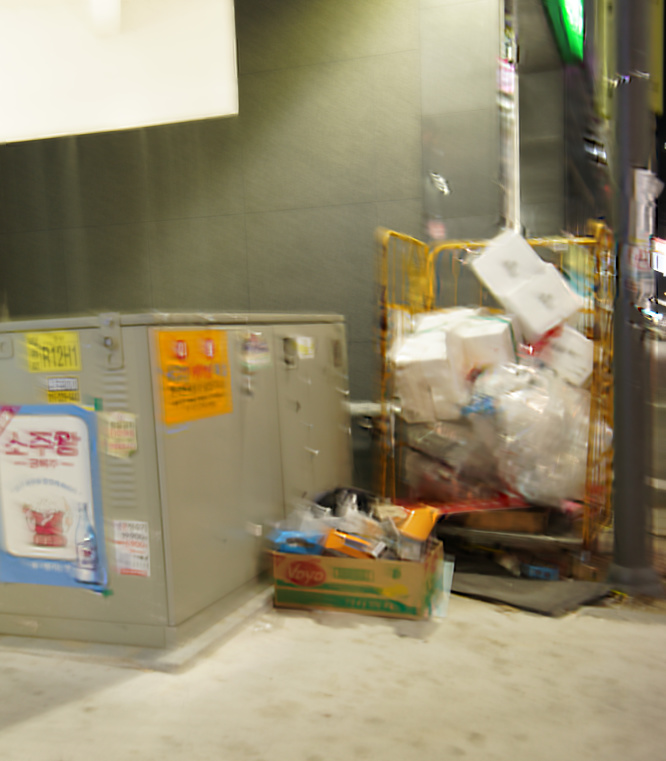} & 
\includegraphics[width=0.12\textwidth]{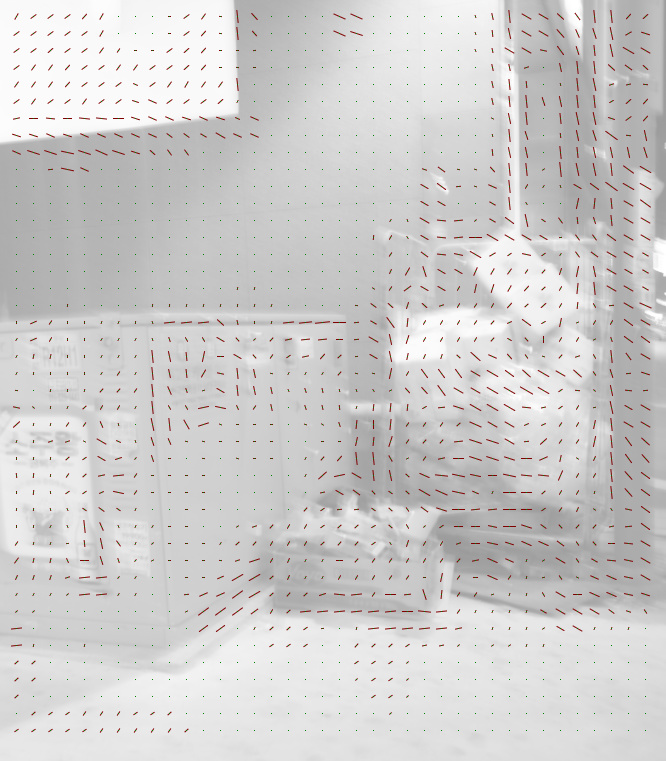} & 
\includegraphics[width=0.12\textwidth]{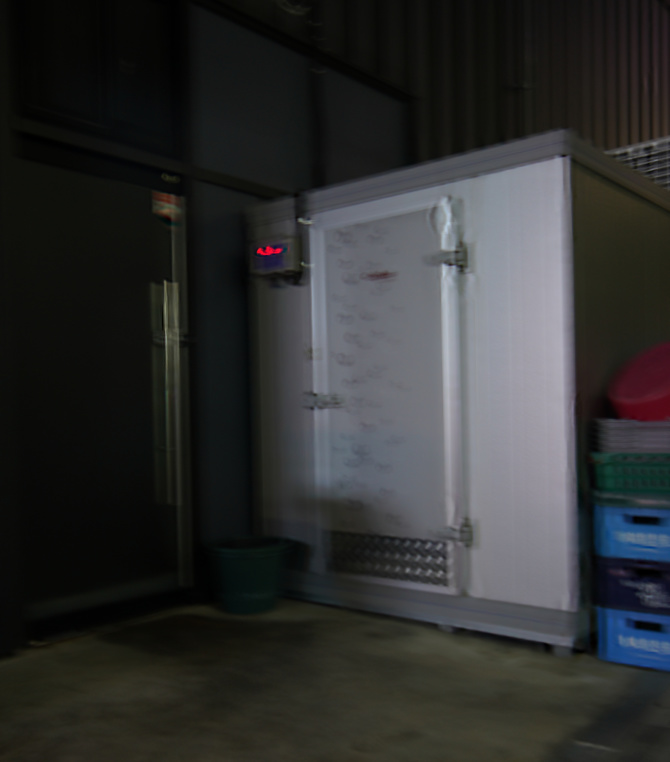} &
\includegraphics[width=0.12\textwidth]{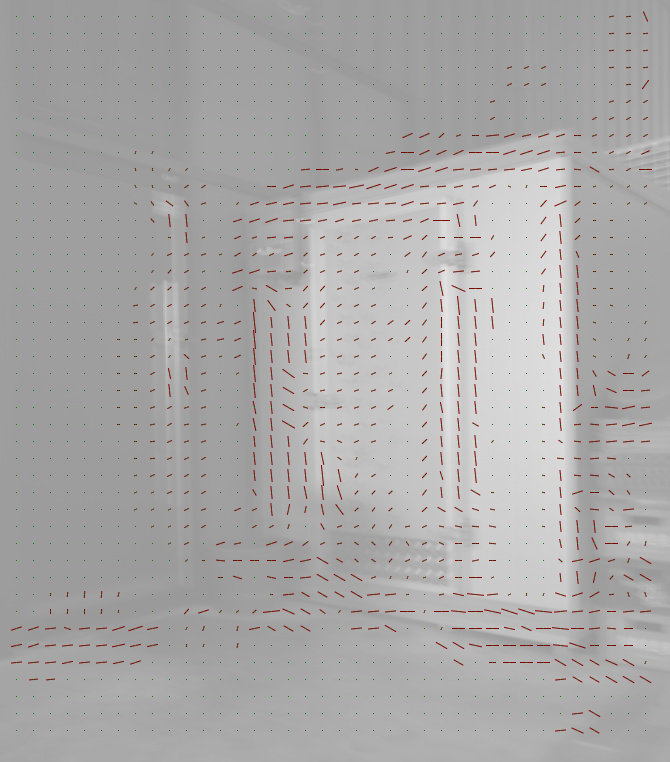} &
\includegraphics[width=0.12\textwidth]{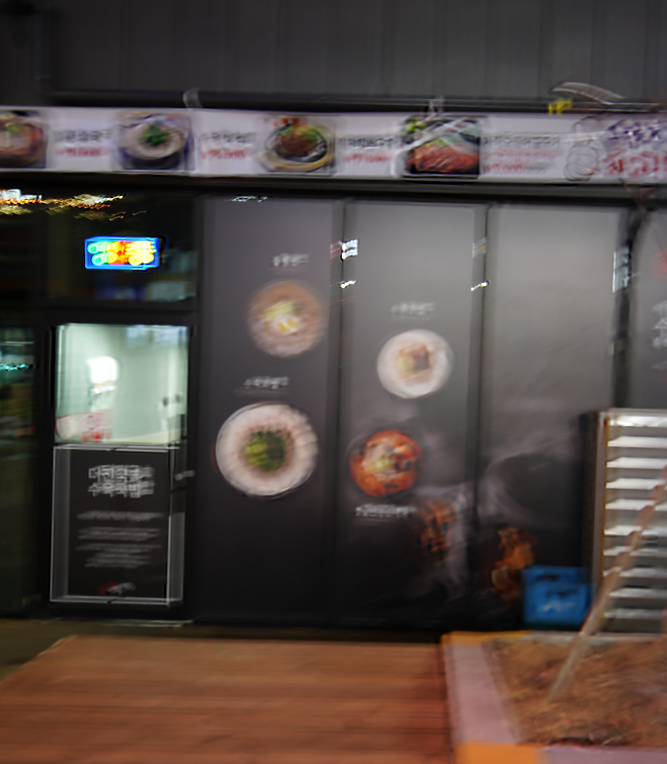} &
\includegraphics[width=0.12\textwidth]{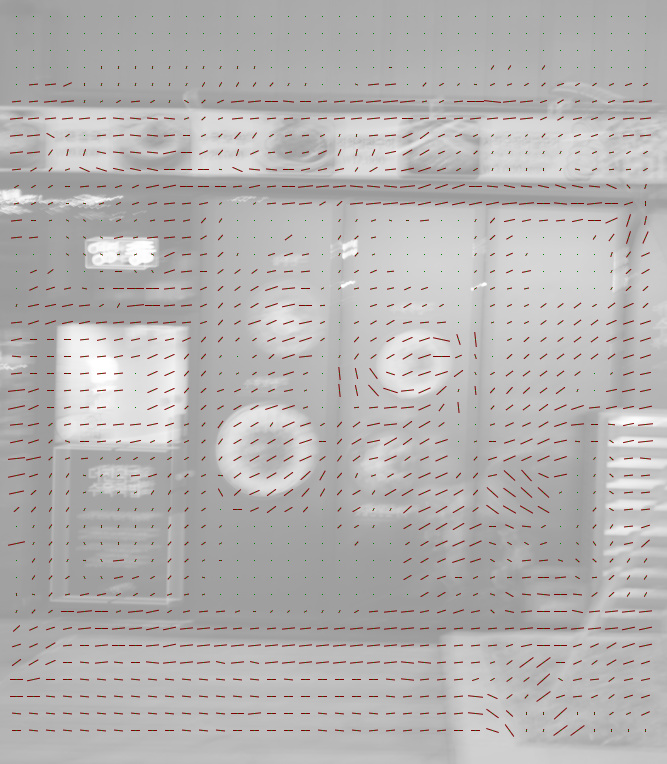} \\

%Blur & our kernels & our reconstruction & gong kernels & gong reconstruction & sun kernels & sun reconstruction
\end{tabular}
\caption{Examples of kernels predicted by our method and corresponding deblurred images on the RealBlur dataset~\cite{rim_2020_ECCV}, and comparison with~\cite{gong2017motion} and~\cite{sun2015learning}. Note that these approaches show a significant correlation with the image structure, and are more prone to fail at capturing the motion structure in low contrasted regions. 
%{\em First row:} blurry images. {\em Second and third rows:} kernels and corresponding reconstructions estimated by our method. {\em Fourth and fifth rows:} kernels and reconstructions by Gong \etal~\cite{gong2017motion}. {\em Sixth and seventh rows:} kernels and reconstructions by Sun \etal~\cite{sun2015learning}.
}
 \label{fig:comparisonRealBlur}
\end{figure}

\end{landscape}

\section{Richardson-Lucy derivation}\label{sec:app:rl}
% {\color{red}JL: hay que formalizar el problema primero. También hablamos de maximum likelihood cuando en el párrafo anterior dijimos que ibamos a hacer MAP. La gran mayoría de las variables no están definidas. La ecuacion (6) no debería aparecer u en el rhs? Para las condiciones KKT cuales son las restricciones del problema?}
% {\color{blue}GC: RL maximiza el likelihood. Si se agrega un término de prior se obtiene el MAP. Al final se obtiene el MAP porque se usa TV. En la ec. 6 el u está metido en $\hat{v_i}$. La restricción KKT que se usa es que $u_j>0$.}

For the sake of completeness, in this section we summarize the derivation of the Richardson-Lucy iteration \cite{whyte11a}.  Richardson-Lucy \cite{Richardson1972, Lucy1974}, algorithm recovers the latent sharp image as the \textit{maximum-likelihood} estimate under a \textit{Poisson noise} model. The likelihood of the blurry image $\mathbf{v}$ given the latent image $\mathbf{u}$ is given by
\begin{equation}
    p\left( \mathbf{u} \vert \mathbf{v} \right) = \prod_i \frac{\hat{v_i}^{v_i} \exp^{-\hat{v_i}}}{v_i!},
    \label{ec:poisson_likelihood}
\end{equation}
where
\begin{equation}
    \hat{v_i} = \sum_j  \langle \bar{\mathbf{u}}_{i,j}, \mathbf{k}_{i,j} \rangle = \sum_j H_{ij} u_j.
\end{equation}
Here, $\mathbf{H}$ denotes the  \textit{degradation} matrix of size $N_v \times N_v$, where $N_v$ is the number of pixels in the image. Each column of $\mathbf{H}$ contains the \textit{Point Spread Function} for the corresponding pixel in the latent image $\mathbf{u}$.
Maximizing \eqref{ec:poisson_likelihood} is equivalent to minimizing the negative log-likelihood
\begin{equation}
    \mathcal{L}_{pois} = \sum_i \hat{v_i} - v_i \log \hat{v_i} + \log \left( v_i! \right).
\end{equation}
The cost function is optimized over the latent image $\mathbf{u}$, under the constraint $u_j \geq 0$, for all pixel $j$. Therefore, the optimum must satisfy the Karush-Kuhn-Tucker conditions for every pixel:
\begin{numcases}{}
u_j\frac{\partial }{\partial u_j} \mathcal{L}_{pois} = 0 & $u_j > 0$, \label{eq:KKT_RL_2a} \\
\frac{\partial }{\partial u_j} \mathcal{L}_{pois} \geq 0  &  $u_j=0$. \label{eq:KKT_RL_2b}
\end{numcases}
The Richardson-Lucy algorithm builds up on~\eqref{eq:KKT_RL_2a}. From this, 
\begin{equation}
u_j \sum_i \frac{\partial }{\partial u_j} \left( \hat{v_i} - v_i \log \hat{v_i} + \log \left({v_i} ! \right) \right) = 0 
\end{equation}
\begin{equation}
u_j \sum_i  \left( \frac{ \partial \hat{v_i} }{\partial u_j} - \frac{v_i}{\hat{v_i}} \frac{\partial \hat{v_i} }{\partial u_j} \right) = 0 
\label{ec:RL_der1}
\end{equation}
\begin{equation}
u_j \sum_i H_{ij} = u_j \sum_i H_{ij} \frac{v_i}{\hat{v_i} }.  
\end{equation}
Since the blur is conservative, the sum of each column of $\mathbf{H}$ is equal to one. Hence, 
\begin{equation}
u_j = u_j \sum_i H_{ij} \frac{{v}_i}{\hat{v_i} }.  
\end{equation}
This equality corresponds to a fixed point equation, that can be solved {\em via} the following update rule:
\begin{equation}
\hat{u}_j^{t+1} = \hat{u}_j^{t} \sum_i H_{ij} \frac{v_i}{\left( \mathbf{H} \hat{\mathbf{u}} \right)_i}.  
\end{equation}
In matrix-vector form, this writes
\begin{equation}
    \hat{\mathbf{u}}^{t+1} = \hat{\mathbf{u}}^t \circ \mathbf{H}^T \left( \frac{\mathbf{v}}{\mathbf{H}\hat{\mathbf{u}}^t} \right), 
    \label{ec:basicRL}
\end{equation}
where the $\circ$ denotes the Hadamard product and the fraction represents the element-wise division of two vectors. Note that choosing an initial condition $\hat{\mathbf{u}}^0 \geq 0$ ensures that $\hat{\mathbf{u}}^t \geq 0$ for all $t > 0$.

\end{document}